\PassOptionsToPackage{table,dvipsnames}{xcolor}

\documentclass[sigconf]{acmart}
\AtBeginDocument{%
  }

\usepackage{cleveref}
\usepackage{subcaption}
\usepackage{graphicx}
\usepackage{array}

\usepackage{amssymb}
\usepackage{multirow}

\newcommand{\resolution}{2319 × 1553}  % 定义平均分辨率
\setcopyright{acmlicensed}
\copyrightyear{2025}
\acmYear{2025}
\acmDOI{10.1145/3591106.3592251} % 假设 DOIs 已知

\begin{document}

\title{Single Document Image Highlight Removal via A Large-Scale Real-World Dataset and A Location-Aware Network}

\author{Lu Pan$^{1*}$, Yu-Hsuan Huang$^{2*}$, Hongxia Xie$^{1\dag}$, Cheng Zhang$^{1}$, Hongwei Zhao$^{1\dag}$, Hong-Han Shuai$^{3}$, Wen-Huang Cheng$^{4}$}
\thanks{$^*$Equal contribution. $^\dag$Corresponding author.}

\affiliation{%
  % \institution{$^1$College of Computer Science, Jilin University}
    \institution{$^1$College of Computer Science, Jilin University}
  % \city{Changchun}
  % \country{China}
  \country{}
}

\affiliation{%
  \institution{$^2$Institute of Information Science, Academia Sinica}
  \country{}
}

\affiliation{%
  \institution{$^3$National Yang Ming Chiao Tung University}
  \country{}
}

\affiliation{%
  \institution{$^4$National Taiwan University}
  \country{}
}

\renewcommand{\shortauthors}{Lu Pan et al.}
\renewcommand\footnotetextcopyrightpermission[1]{}
\settopmatter{printacmref=false} %remove ACM reference format

\begin{abstract}

Reflective documents often suffer from specular highlights under ambient lighting, severely hindering text readability and degrading overall visual quality. Although recent deep learning methods show promise in highlight removal, they remain suboptimal for document images, primarily due to the lack of dedicated datasets and tailored architectural designs. To tackle these challenges, we present DocHR14K, a large-scale real-world dataset comprising 14,902 high-resolution image pairs across six document categories and various lighting conditions. To the best of our knowledge, this is the first high-resolution dataset for document highlight removal that captures a wide range of real-world lighting conditions. Additionally, motivated by the observation that the residual map between highlighted and clean images naturally reveals the spatial structure of highlight regions, we propose a simple yet effective Highlight Location Prior (HLP) to estimate highlight masks without human annotations.
Building on this prior, we present the Location-Aware Laplacian Pyramid Highlight Removal Network (L²HRNet), which effectively removes highlights by leveraging estimated priors and incorporates diffusion module to restore details.
Extensive experiments demonstrate that DocHR14K improves highlight removal under diverse lighting conditions. Our L\textsuperscript{2}HRNet achieves state-of-the-art performance across three benchmark datasets, including a 5.01\% increase in PSNR and a 13.17\% reduction in RMSE on DocHR14K. 

%especially achieving a significant improvement in terms of PSNR, improving from 24.430dB to 27.558dB over RD dataset. 
\vspace{-20pt}
\end{abstract}

% \begin{CCSXML}
% <ccs2012>
%    <concept>
%        <concept_id>10010147.10010178.10010224.10010226.10010236</concept_id>
%        <concept_desc>Computing methodologies~Computational photography</concept_desc>
%        <concept_significance>500</concept_significance>
%        </concept>
%    <concept>
%        <concept_id>10010147.10010178.10010224.10010245.10010254</concept_id>
%        <concept_desc>Computing methodologies~Reconstruction</concept_desc>
%        <concept_significance>500</concept_significance>
%        </concept>
%  </ccs2012>
% \end{CCSXML}

% \ccsdesc[500]{Computing methodologies~Reconstruction}
% \ccsdesc[500]{Computing methodologies~Computational photography}

% \keywords{document image highlight removal, highlight localization, diffusion-based enhancement}

\begin{teaserfigure}
  \includegraphics[width=\textwidth]{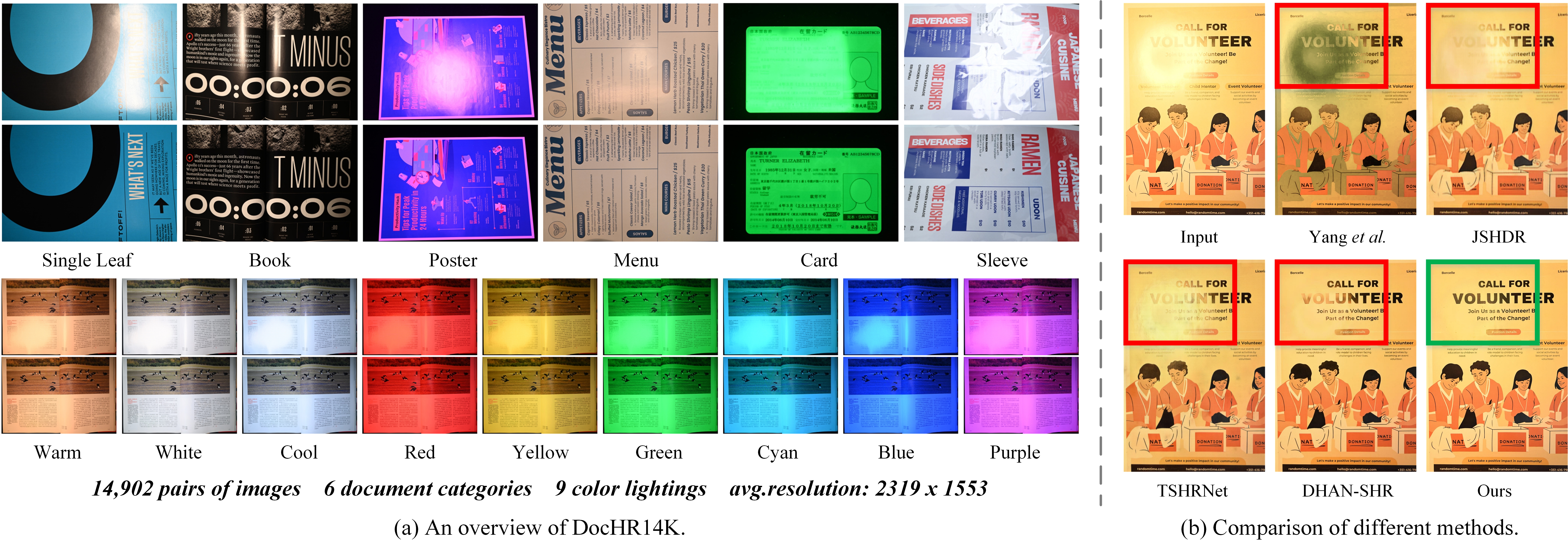}
  \vspace*{-20pt}
  \caption{(a) An overview of DocHR14K. Top: Example images of six document categories. Bottom: Example images of nine lighting colors. (b) Comparisons among the input image, Yang \textit{et al.} \cite{yang2010real}, JSHDR \cite{fu2021multi}, TSHRNet \cite{fu2023towards}, DHAN-SHR \cite{guo2024dual}, and our L\textsuperscript{2}HRNet result on real-world document image degraded by the specular highlight. Zoom in for the best view. All images are captured in real-world conditions.}
  \label{fig:image_samples}
\end{teaserfigure}
\maketitle

\section{Introduction}
% \vspace*{-6pt}
Imagine trying to scan or photograph a document only to find that glaring reflections and bright spots obscure critical information, making text hard to read and essential details easy to miss. This challenge is all too common in everyday scenarios where ambient lighting creates harsh reflections on glossy or smooth surfaces. These specular highlights not only degrade image quality, but also severely impact the performance of text-related vision tasks such as document enhancement \cite{kligler2018document}, optical character recognition (OCR) \cite{bissacco2013photoocr,shi2016end, qiao2022davarocr, li2022dit}, and layout analysis \cite{zhong2019publaynet, li2020docbank, xu2020layoutlm, li2022dit}.

Removing highlights from digital document images remains a challenging problem. Traditional physics-based illumination models~\cite{shafer1985using, bajcsy1996detection,tan2005separating,yoon2006fast,kim2013specular,akashi2014separation,yang2014efficient,guo2018single,yamamoto2019general} often struggle in real-world settings due to their strict assumptions.
While recent deep learning approaches~\cite{fu2021multi, wu2021single, fu2023towards, hu2024highlight, guo2014robust,zhang2024highlightremover} have shown promise on natural images, they are generally designed for low-resolution, general scenes. As a result, they are less effective on high-resolution document images, where preserving fine details like text and graphics is critical~\cite{li2023high}. This challenge is further exacerbated by the lack of large-scale document datasets that capture diverse lighting conditions, limiting the ability of models to generalize and effectively remove highlights across real-world scenarios.

%To solve the above problems, we first construct a large-scale real-world highlight dataset (\textbf{DocHR14K}) in the document domain, which consists of 14,902 highlight and highlight-free document image pairs. 
To address these limitations, we introduce \textbf{DocHR14K}, a large-scale dataset comprising 14,902 paired highlight and highlight-free document images. Typical images are given in \Cref{fig:image_samples}. %Unlike existing highlight removal datasets, DocHR14K includes abundant text-related information based on various document types, such as art albums, posters, and ID cards. 
Unlike existing highlight removal datasets, DocHR14K covers various document types with rich textual content, ranging from books and posters to more uncommon forms like menus and ID cards. 
Moreover, while prior datasets often \textit{assume that specular chromaticity is uniform (i.e., highlight is colorless)}, our dataset captures images under nine representative colored lighting conditions, enabling better generalization to real-world scenarios.
Furthermore, motivated by the observation that \textbf{the residual map of highlight-contaminated and clean images naturally reveals the spatial structure of highlight regions}, we introduce a simple yet effective \textbf{H}ighlight \textbf{L}ocation \textbf{P}rior (HLP) to estimate highlight masks without annotations.
Based on HLP, we propose \textbf{L}ocation-Aware \textbf{L}aplacian pyramid-based \textbf{H}ighlight \textbf{R}emoval \textbf{Net}work (L\textsuperscript{2}HRNet), which incorporates the estimated highlight prior to enhance highlight removal and uses a diffusion module \cite{sohl2015deep, ho2020denoising, song2020denoising, saharia2022image} to refine fine details.
Specifically, we first decompose the document image into low- and high-frequency components, where the former capture global highlight patterns and the latter preserve fine details. %For the low-frequency band, we propose a two-stage cascade network, which first predicts highlight location mask by a highlight detection network (HDNet) and then utilizes the predicted soft mask to achieve global highlight removal effectively by a highlight removal network (HRNet). 
For the low-frequency band, we design a two-stage cascade network: a \textbf{H}ighlight \textbf{D}etection \textbf{Net}work (HDNet) first predicts a soft highlight mask, which then guides a \textbf{H}ighlight \textbf{R}emoval \textbf{Net}work (HRNet) to suppress global highlights. 
%Using the global highlight removal result, we further 
As for the high-frequency band, we introduce a \textbf{D}iffusion-based \textbf{E}nhancement \textbf{M}odule (DEM) to achieve fine-grained textual details recovery. 
Experiments show that the diverse attributes of DocHR14K enhance model generalization across real-world scenarios, enabling models to handle complex visual variations in document images effectively. Furthermore, our method achieves state-of-the-art performance, not only showing a 5.01\% improvement in PSNR and a 13.17\% improvement in RMSE on DocHR14K, but also achieving a significant PSNR gain from 24.430dB to 27.558dB on the RD dataset compared to previous methods. Our main contributions are summarized as follows:
\vspace{-6pt}
\begin{itemize}
    \item We present \textbf{DocHR14K}, a large-scale, high-resolution dataset for document image highlight removal. Unlike existing datasets, it includes six document types and nine lighting conditions, better reflecting real-world scenarios.
    \item We design \textbf{L\textsuperscript{2}HRNet}, a location-aware Laplacian Pyramid-based network that leverages the proposed Highlight Location Prior (HLP) to guide highlight removal, and incorporate diffusion-based enhancemnt module (DEM) to restore text information.
    \item Extensive experiments demonstrate that DocHR14K effectively supports highlight removal under diverse lighting and content conditions. Additionally, our L\textsuperscript{2}HRNet outperforms previous methods across multiple benchmarks.
\end{itemize}\vspace{-6pt}

% Table 1: Comparison table
\begin{table*}[t]
\centering
\begin{tabular}{lccccccc}
\toprule
Datasets & Source & Number of Pairs & Mean Resolution & Method & Domain & Daily & Color \\
\midrule
SD1, SD2, RD & PRCV'21 & 30,000  & 512 × 512 & Synthetic + Illumination Toggling & Text In The Wild & No & No \\
SHIQ & CVPR'21 & 10,825 & 200 × 200 & Synthesis & Natural & No & No \\
PSD & TMM'21 & 13,380 & 738 × 512 & Cross Polarization & Natural & No & No \\
SSHR & ICCV'23 & 135,000 & 256 × 256 & Synthesis & Object-Level & No & No \\ 
% NSH & ACM MM'24 & 30,000 & N/A & Cross Polarization & Natural  & No & No\\
SHDocs & NIPS'24 & 19,104 & 1,224 × 1,024 & Cross Polarization & Document & No & No \\
\rowcolor[HTML]{EFEFEF}
DocHR14K & \textbf{Ours} & 14,902 & 2,319 × 1,553 & Cross Polarization & Document & 3,405 & 2,511 \\
\bottomrule
\end{tabular}
\vspace{3pt}
 \caption{Comparison of existing highlight removal datasets. (Domain = Application Domain; Daily = Daily-living lighting conditions; Color = Color Lighting Images)} \vspace{-15pt}
\label{tab:dataset_comparison}
\end{table*}
\vspace{-6pt}

\section{Related Works}
\label{sec:related_works}

\subsection{Image Highlight Removal Dataset}

Specular highlight removal has long been recognized as a challenging problem in imaging processing. With the advent of deep learning, several large-scale benchmark datasets have been curated to facilitate the development and evaluation of highlight removal algorithms. Wu \textit{et al.} \cite{wu2021single} introduced the first large-scale dataset (PSD) for highlight removal, constructed under controlled laboratory conditions using a cross-polarization setup. Concurrently, Fu \textit{et al.} \cite{fu2021multi} proposed SHIQ, a dataset targeting natural scenes, generated through post-processed multi-image highlight separation~\cite{guo2014robust}.  To extend the applicability to object-level scenarios, Fu \textit{et al.} \cite{fu2023towards} further presented SSHR, a synthetic dataset comprising 135K images rendered using 3D object models and high-dynamic-range (HDR) environment maps. More recently, Zhang \textit{et al.} \cite{zhang2024highlightremover} introduced NSH, a real-world dataset captured under natural lighting using a multi-illumination acquisition system. In the context of document images, Hou~\textit{et al.}\cite{hou2021text} proposed the first dataset for text image highlight removal, comprising image pairs either synthetically generated or captured by turning off the light source. However, these approaches either fail to reproduce realistic highlight distortions or suffer from inconsistent global illumination between image pairs, limiting their effectiveness for supervised learning. To further advance the field, Leong \textit{et al.} ~\cite{leong2024shdocs} constructed SHDocs, a cross-polarization-based dataset tailored for highlight removal in printed materials. Despite its precise alignment and high capture quality, SHDocs is restricted to grayscale imaging, a limited set of document types, and fixed LED lighting conditions, thereby hindering its generalization to broader real-world scenarios.

In contrast, our dataset is the first high-resolution document highlight dataset that includes a wide variety of document types and lighting conditions, providing a solid foundation for advancing future research in real-world document image highlight removal.

\subsection{Image Highlight Removal Method}
Highlight removal methods have evolved significantly in recent years.Wu \textit{et al.} \cite{wu2021single} proposed a GAN-based framework that integrates global contextual reasoning with highlight detection cues to guide restoration. Fu \textit{et al.} \cite{fu2021multi} designed a joint highlight detection and removal network based on a physically-inspired highlight formation model, leveraging dilated convolutions to enhance contextual awareness. Subsequently, Fu \textit{et al.} \cite{fu2023towards} introduced a three-stage decomposition network that separates input images into albedo, shading, and specular residual components for more structured object-level highlight removal. To better capture global dependencies, Guo \textit{et al.} \cite{guo2024dual} employed a Transformer-based architecture that models inter-channel correlations and long-range spatial interactions, resulting in coherent and perceptually consistent outputs. More recently, Zhang \textit{et al.} ~\cite{zhang2024highlightremover} proposed a spatially valid pixel learning framework that selectively restores highlight-contaminated regions using context-aware fusion and location-sensitive feature transformations.

Unlike previous works, we present the first network specifically tailored for high-resolution document images. Our method exploits Laplacian Pyramid decomposition \cite{burt1987laplacian} to separate and process frequency components, incorporates a residual-based highlight location prior for spatial guidance, and applies a residual diffusion model to restore fine text structures. Together, these innovations enable accurate and detail-preserving highlight removal under complex real-world conditions.

\begin{figure}[htbp]
    \centering
    \begin{subfigure}[b]{0.32\linewidth}
        \centering
        \includegraphics[angle=0,width=\linewidth]{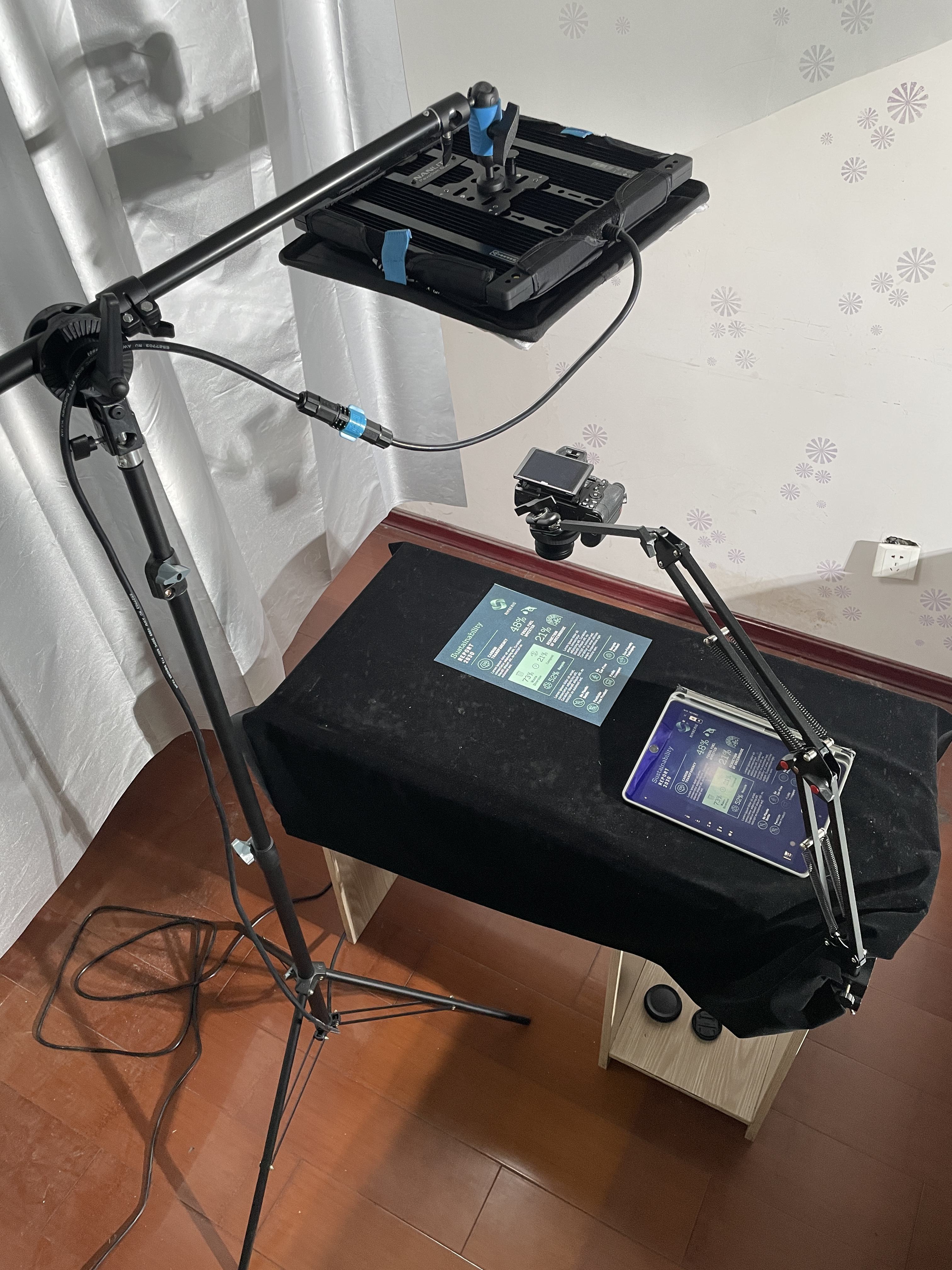}
        \caption{}
        \label{fig:vertical}
    \end{subfigure}
    \hfill
    \begin{subfigure}[b]{0.32\linewidth}
        \centering
        \includegraphics[angle=0,width=\linewidth]{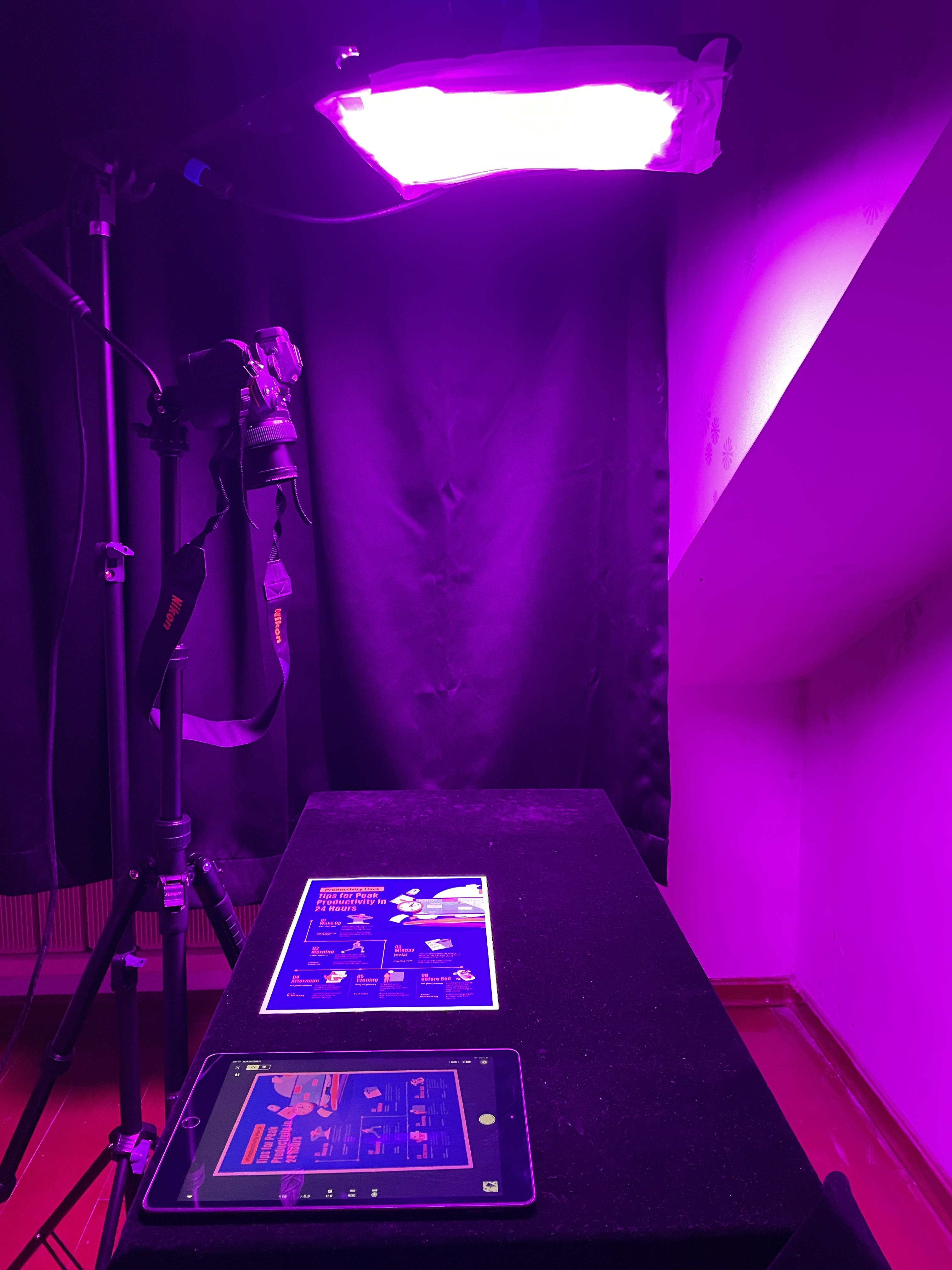}
        \caption{}
        \label{fig:non_vertical_diffuse}
    \end{subfigure}
    \hfill
    \begin{subfigure}[b]{0.32\linewidth}
        \centering
        \includegraphics[angle=0,width=\linewidth]{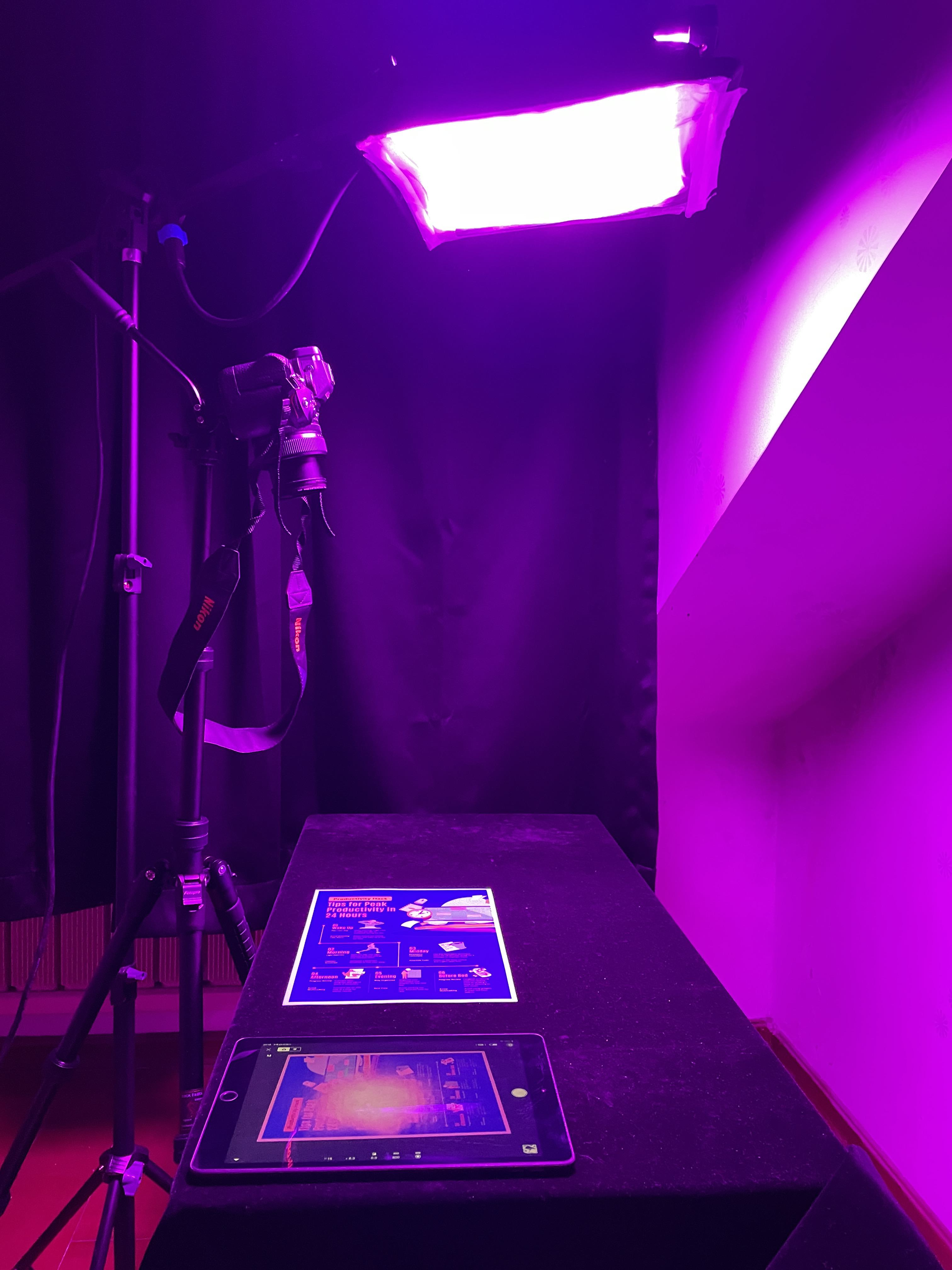}
        \caption{}
        \label{fig:non_vertical_highlight}
    \end{subfigure}
    \begin{subfigure}[b]{0.45\linewidth}
        \centering
        \includegraphics[angle=0,width=\linewidth]{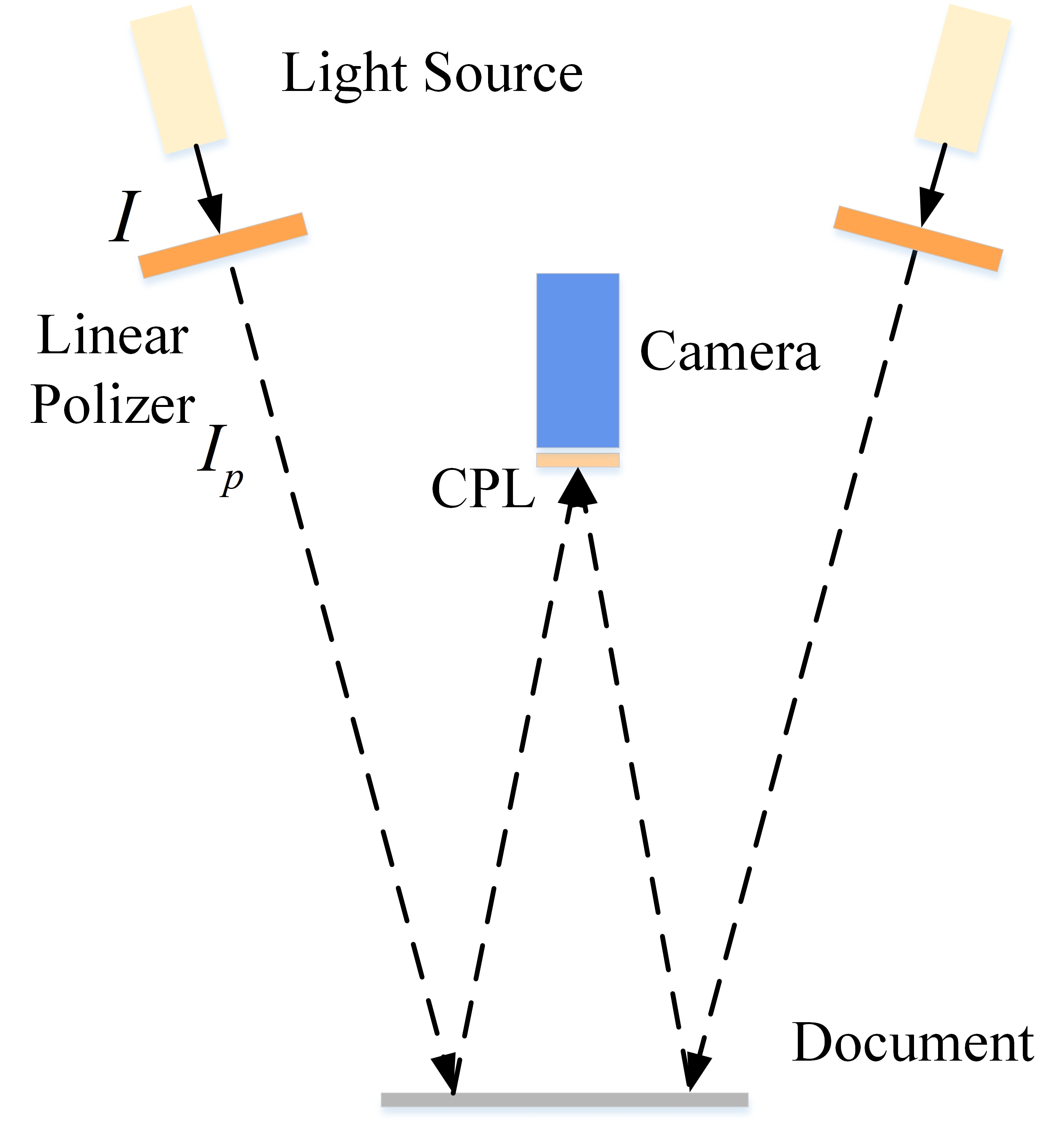}
        \caption{}
        \label{fig:cross_polarized}
    \end{subfigure}
        \begin{subfigure}[b]{0.45\linewidth}
        \centering
        \includegraphics[angle=0,width=\linewidth]{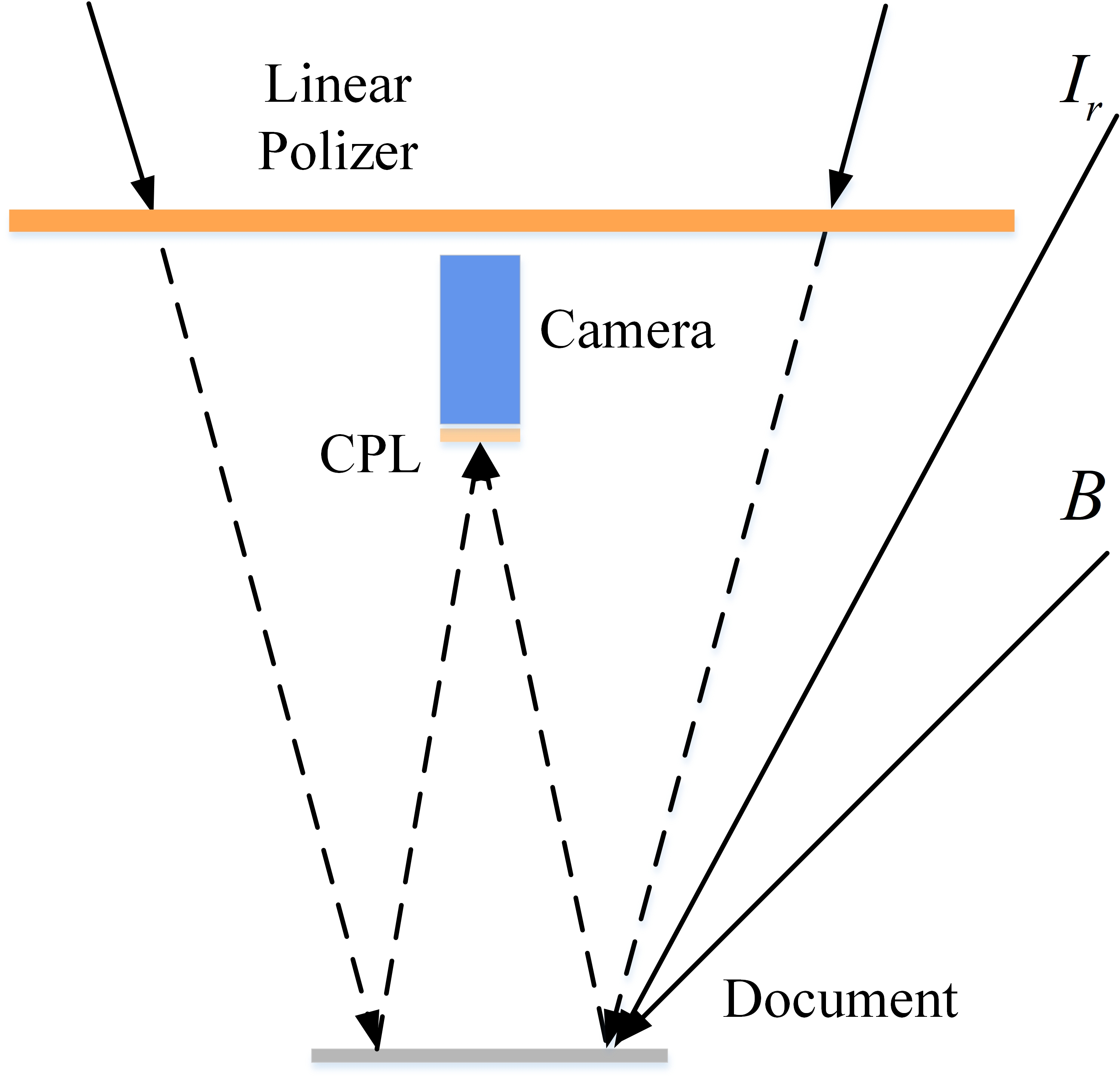}
        \caption{}
        \label{fig:improved_cross_polarized}
    \end{subfigure}
    \caption{Illustration of image capturing. (a)  Vertical shooting under a color temperature of 5600K. (b) Diffuse image captured under purple lighting by 15 degrees. (c) Corresponding highlight image captured by adjusting the angle of the linear polarizer. (d)  Illustration of collection in the laboratory environment. $I$ is the light from the source and $I_p$ denotes the polarized light. (e)  Illustration of collection in the daily living environment. $I_r$ represents residual incident light, and $B$ denotes ambient light. }
    \label{fig:enviroment}
    \vspace{-12pt}
\end{figure}

% \vspace{-12pt}
\section{DocHR14K Dataset}

\label{sec:dataset}

%-------------------------------------------------------------------------
%-------------------------------------------------------------------------

Existing highlight removal datasets either focus on natural images without textual content or lack diversity in document types and lighting conditions. As highlighted in~\Cref{tab:dataset_comparison}, current datasets do not effectively support document highlight removal tasks, as they lack coverage of diverse lighting conditions needed for real-world scenarios. %, which are crucial for improving text visibility and readability under varying lighting conditions. 
Therefore, we have assembled a collection of 14,902 image pairs from real-world scenarios. These images, with an average resolution of \resolution, span six representative document categories and are captured under nine standard lighting colors from five distinct angles. The following presents the detailed descriptions of the dataset construction process.

% %-------------------------------------------------------------------------

\subsection{Dataset Construction}

%Capturing images without highlights by simply removing the light source is impractical, as it significantly alters overall illumination, which degrades image quality. Current techniques, such as cross-polarization~\cite{wu2021single}, while effective in controlled laboratory settings, struggle in everyday environments and risk introducing a domain gap when applied to real-world scenarios. 
To support highlight removal under diverse lighting conditions, it is necessary to collect pairs of document images, one with highlights and one without, captured under varying illumination settings, including daily-living lighting. However, current techniques such as cross-polarization~\cite{wu2021single, wolff1991constraining, wen2021polarization,leong2024shdocs}, while effective in controlled lab environments, often struggle in everyday settings.
Hence, we first refine the cross-polarization method to better suit daily living conditions for collecting image pairs. We then capture images in both laboratory and real-world settings, covering a wide range of document types and lighting conditions to enhance the diversity of the dataset. Finally, the image post-processing is applied to maintain the quality and consistency of the dataset. Details of these strategies are elaborated as follows.

%-------------------------------------------------------------------------
% Figure 2: Donut chart examples with individual legends
\begin{figure*}[t]
    \centering
    \begin{subfigure}[b]{0.2\linewidth}
        \centering
        \includegraphics[angle=0,width=\linewidth]{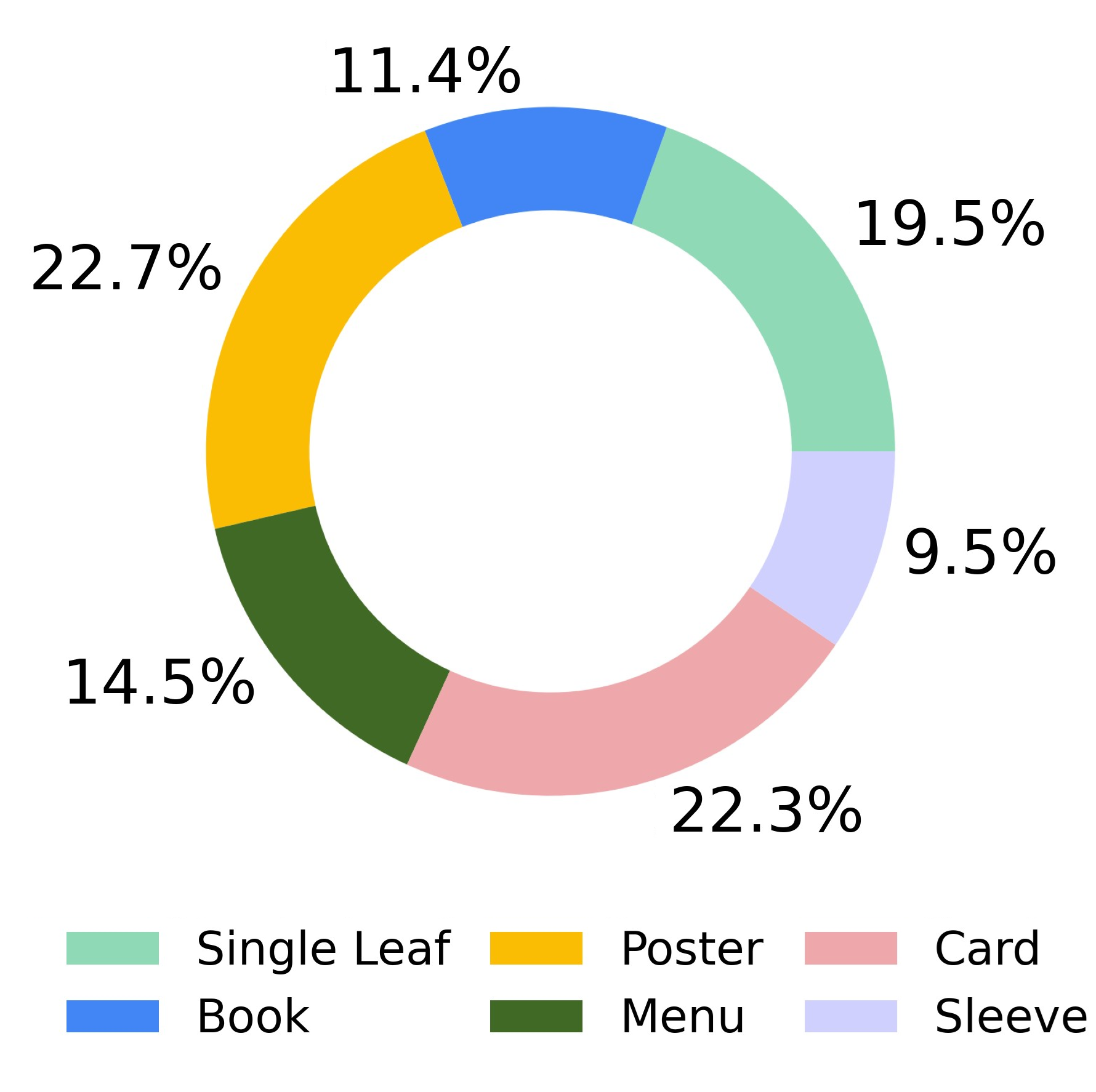}
        \caption{Document Category}
        \label{fig:statics-a}
    \end{subfigure}
    \hfill
    \begin{subfigure}[b]{0.2\linewidth}
        \centering
        \includegraphics[angle=0,width=\linewidth]{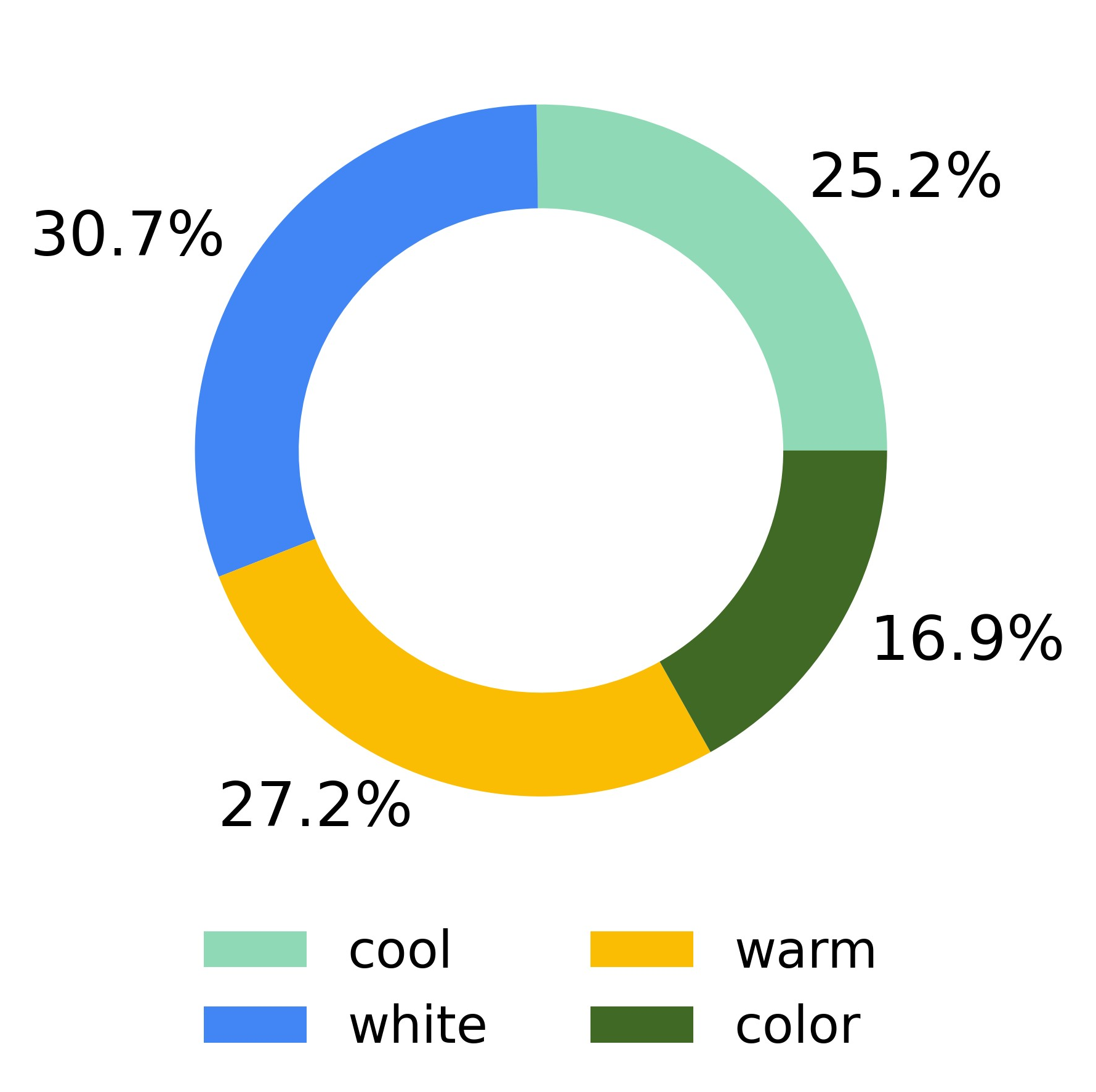}
        \caption{Light Category}
        \label{fig:statics-b}
    \end{subfigure}
    \hfill
    \begin{subfigure}[b]{0.2\linewidth}
        \centering
        \includegraphics[angle=0,width=\linewidth]{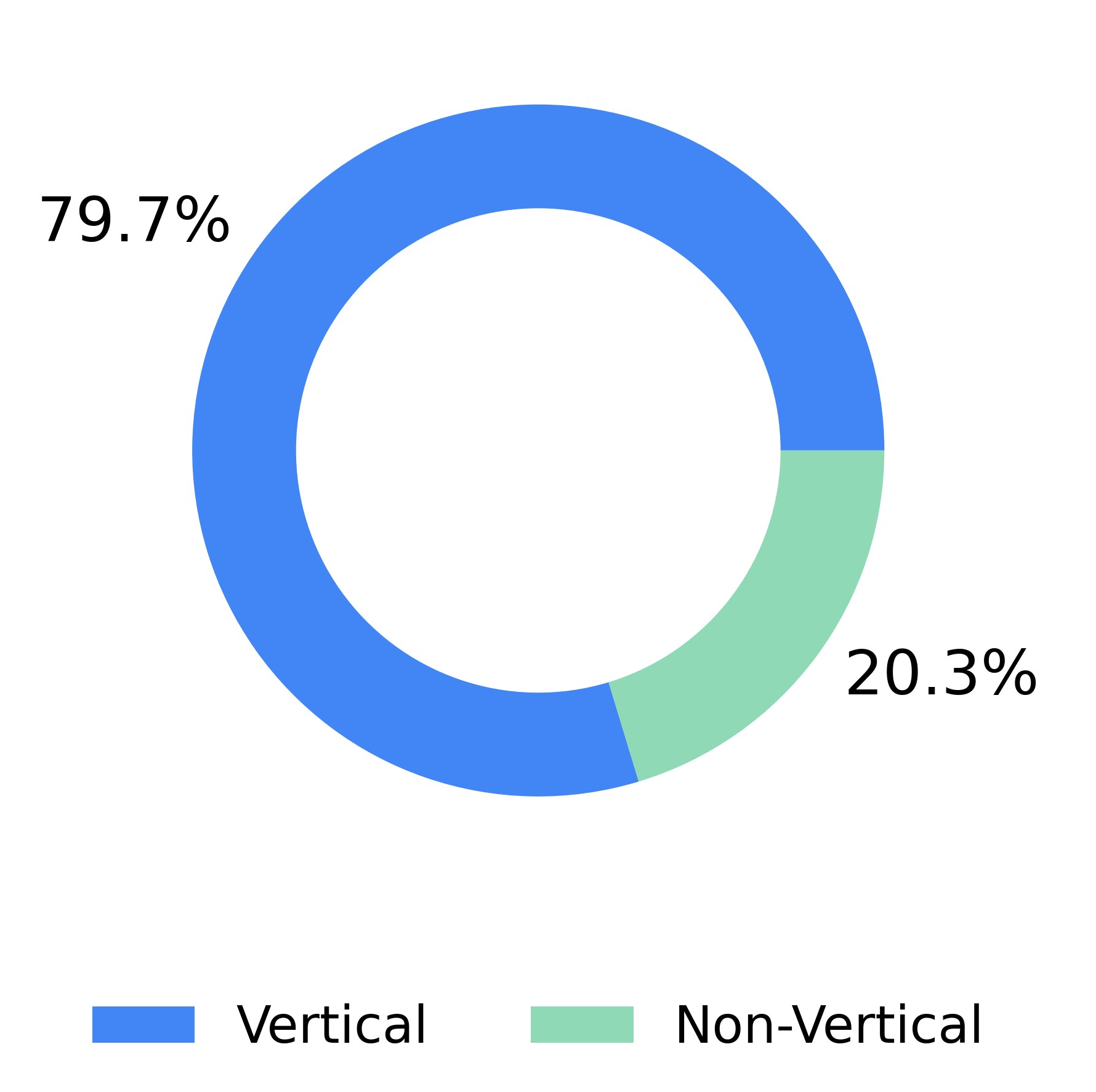}
        \caption{Shooting Angle}
        \label{fig:statics-c}
    \end{subfigure}
    \hfill
    \begin{subfigure}[b]{0.2\linewidth}
        \centering
        \includegraphics[angle=0,width=\linewidth]{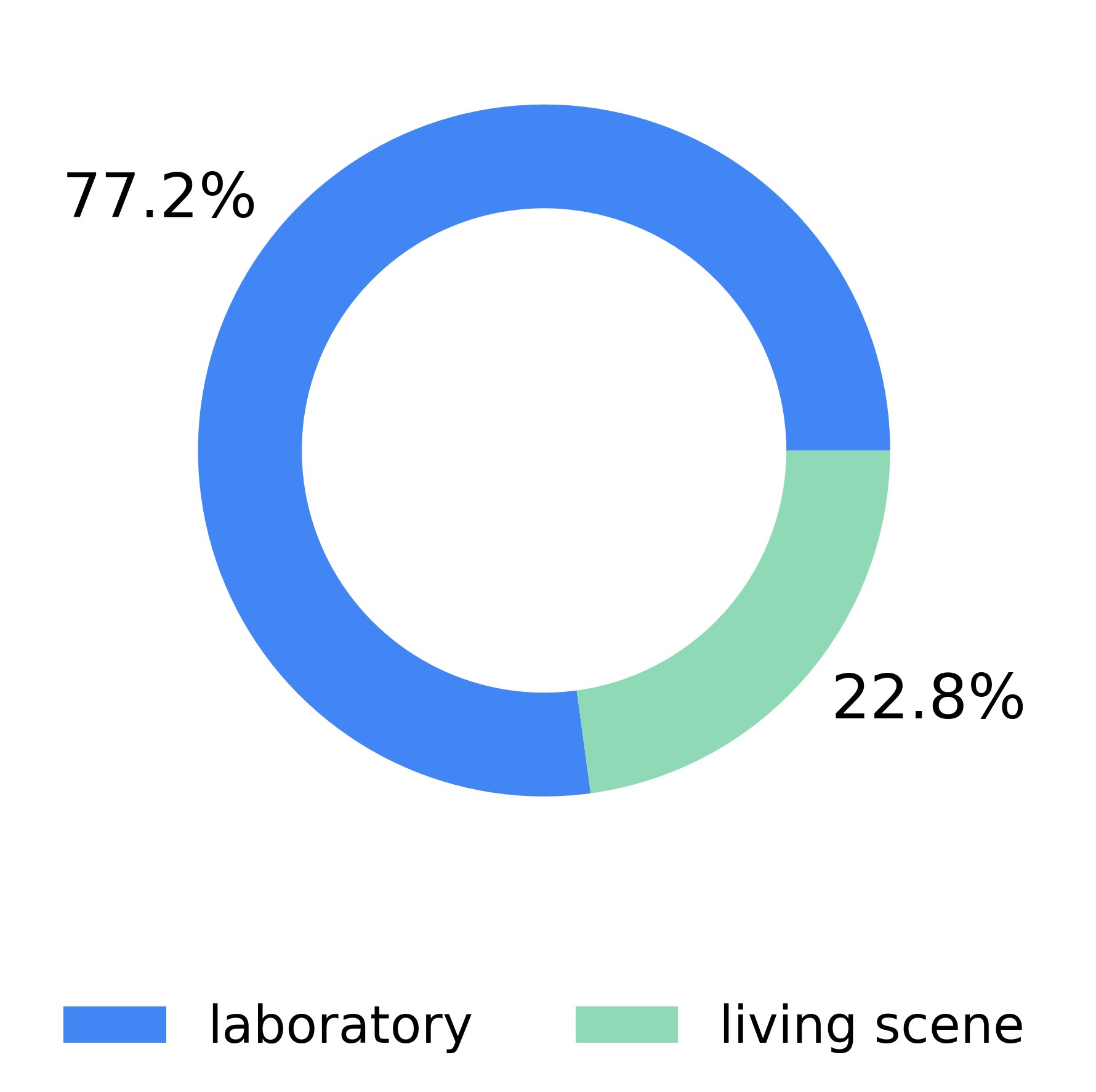}
        \caption{Shooting Environment}
        \label{fig:statics-d}
    \end{subfigure}
    \vspace{-6pt}
    \caption{ Data distribution of DocHR14K.} \vspace{-12pt}
    \label{fig:statics}
\end{figure*}

%-------------------------------------------------------------------------
\noindent \textbf{Cross-Polarization Improvement.} As shown in \Cref{fig:cross_polarized}, the prevailing cross-polarization technique \cite{wu2021single} modifies incident light by positioning a linear polarizer directly in front of the light source, transforming it into polarized light. Subsequently, a circular polarizing lens (CPL) is mounted in front of the camera lens to selectively filter this polarized light, thus capturing a diffuse image. Nevertheless, such an approach proves impractical in real-life settings where light sources, such as ceiling-mounted fluorescent tubes, are not readily accessible for modifications. This limits its applicability in common environments outside controlled setups. %This limitation significantly challenges the application of traditional cross-polarization methods outside controlled laboratory environments.

To this end, we re-evaluate the cross-polarization process, identifying the polarization of incident light as the critical step. As shown in \Cref{fig:improved_cross_polarized}, our improved methodology involves placing the linear polarizer close to the subject of the photograph rather than the light source. This adjustment ensures that all light interacting with the polarizer becomes polarized. When the axis of this linear polarizer is aligned perpendicular to the CPL in front of the camera, a diffuse image is successfully produced \cite{wu2021single}. To manage any residual highlights resulting from light bypassing the linear polarizer, we employ targeted image post-processing techniques. This step is efficient, as highlights typically recur in consistent locations under uniform lighting conditions. Our approach offers two significant benefits: i) it facilitates the capture of diffuse images even in environments with multiple light sources, and ii) it removes the need to attach polarizers to each light source, which is often impractical in real-world settings.%it can be applied in scenarios where modifying the light source directly is unfeasible.  % it is viable in situations where modifying the light source directly is unfeasible.

\par

\noindent \textbf{Dataset Diversification and Realization.}
 Laboratory-based image collection provides controlled lighting conditions that enable the simulation of various light colors and intensities. For our dataset, we employ the Nanlite Pavo Slim 60c as our light source and a Nikon Z5 camera for imaging. To mitigate misalignment caused by unwanted movements, we utilize a metal bracket to secure the camera for vertical shots, capturing image pairs remotely via a mobile app, as depicted in~\Cref{fig:vertical}. For non-vertical angles that are more challenging, including 15°, 30°, 45°, and angles greater than 45°, we use a tripod to capture 3,027 image pairs, as shown in~\Cref{fig:non_vertical_diffuse}. %For angles challenging for vertical shooting-15, 30, 45, and beyond 45 degrees-we capture 3027 image pairs using a tripod, shown in~\Cref{fig:non_vertical_diffuse}. 

 During these sessions, a linear polarizer is positioned in front of the LED panel, with a CPL filter placed in front of the camera. Initially, the linear polarizer is adjusted to achieve a diffuse image, illustrated in~\Cref{fig:non_vertical_diffuse}. Subsequently, the polarizer is rotated by 90 degrees to capture images with specular highlights, as demonstrated in~\Cref{fig:non_vertical_highlight}. In natural settings, we select four typical lighting conditions: LED lamps, bar lights, ring lights, and fluorescent tubes. A 50 cm square linear polarizer is strategically placed above the document to polarize the incident light. The polarizer is first adjusted to secure a diffuse image, followed by a vertical rotation to capture the highlight image. 

%-------------------------------------------------------------------------
\noindent \textbf{Post Processing.} After collecting image pairs, we first organize highlight-diffuse pairs into corresponding folders and crop out areas with unwanted highlight residuals. Next, we remove misaligned pairs by calculating the residual image and checking for image misalignment outside the highlight region. Each image pair is then annotated with its document category and light color, providing fine-grained labels to support related tasks. Additionally, quality control is made by stratified sampling based on document category, lighting type, language, shooting angle, and environment. \footnote{Detail of quality control can be found in \textbf{Section A} of the supplementary material.}

\subsection{Characteristics of DocHR14K Dataset}
%Several unique and challenging characteristics distinguish our DocHR14K dataset from existing highlight removal datasets. 
Our DocHR14K dataset stands out from existing highlight removal datasets due to several distinctive and challenging characteristics.
Please refer to \Cref{fig:image_samples} for visualizations of the resultant examples from our dataset.

\noindent \textbf{(1) High-resolution image pairs.}
%As shown in \Cref{tab:dataset_comparison}, our dataset is a high-resolution highlight removal dataset with an average resolution of \resolution, which is the highest among existing datasets. 
As shown in \Cref{tab:dataset_comparison}, our dataset is a high-resolution highlight removal dataset, with the highest average resolution (\resolution) among existing datasets.

\noindent \textbf{(2) Diverse document categories.}
As illustrated in \Cref{fig:statics-a}, we provide 6 document types with corresponding labels: single leaf, book, poster, menu, card, and plastic-sleeved document. Each image is labeled with its corresponding document categories.

\noindent \textbf{(3) Varied light sources.}
As shown in \Cref{fig:statics-b}, our dataset has 4 primary categories of lighting conditions: cold light, white light, warm light, which are frequently encountered in real-life settings, and color lighting conditions. %Additionally, we incorporated six representative colors by selecting hue-based lighting every 60 degrees: red (0 degrees), yellow (60 degrees), green (120 degrees), cyan (180 degrees), blue (240 degrees), and purple (300 degrees). 
The color lighting category further comprises six representative colors, selected at 60-degree intervals in the HSV color space: red (0 degrees), yellow (60 degrees), green (120 degrees), cyan (180 degrees), blue (240 degrees), and purple (300 degrees). 
Each image is labeled with its corresponding color information.

\noindent \textbf{(4) Varied shooting angles.}
As shown in \Cref{fig:statics-c}, We collect 3,027 image pairs captured at non-vertical angles to simulate scenarios where a vertical digital copy of the document is hard to obtain. Specifically, we provide 4 progressively increasing angles, ranging from 15 to 60 degrees, to reflect diverse real-world conditions.

\noindent \textbf{(5) Daily-living lighting conditions.}
As shown in \Cref{fig:statics-d}, We collect 3,405 image pairs in everyday lighting scenarios, including desk lamps, bar lights, ring lights, and fluorescent tubes. This is the first highlight removal dataset that includes common lighting conditions benefiting from our improved cross-polarization pipeline.\footnote{Real world illustrations of the improved cross-polarization method are given in \textbf{Section C} of the supplementary material.}

\section{L\textsuperscript{2}HRNet}

\begin{figure*}[htbp]  % [htbp] 指定浮动行为
    \centering  % 图片居中
    \includegraphics[width=0.8\textwidth]{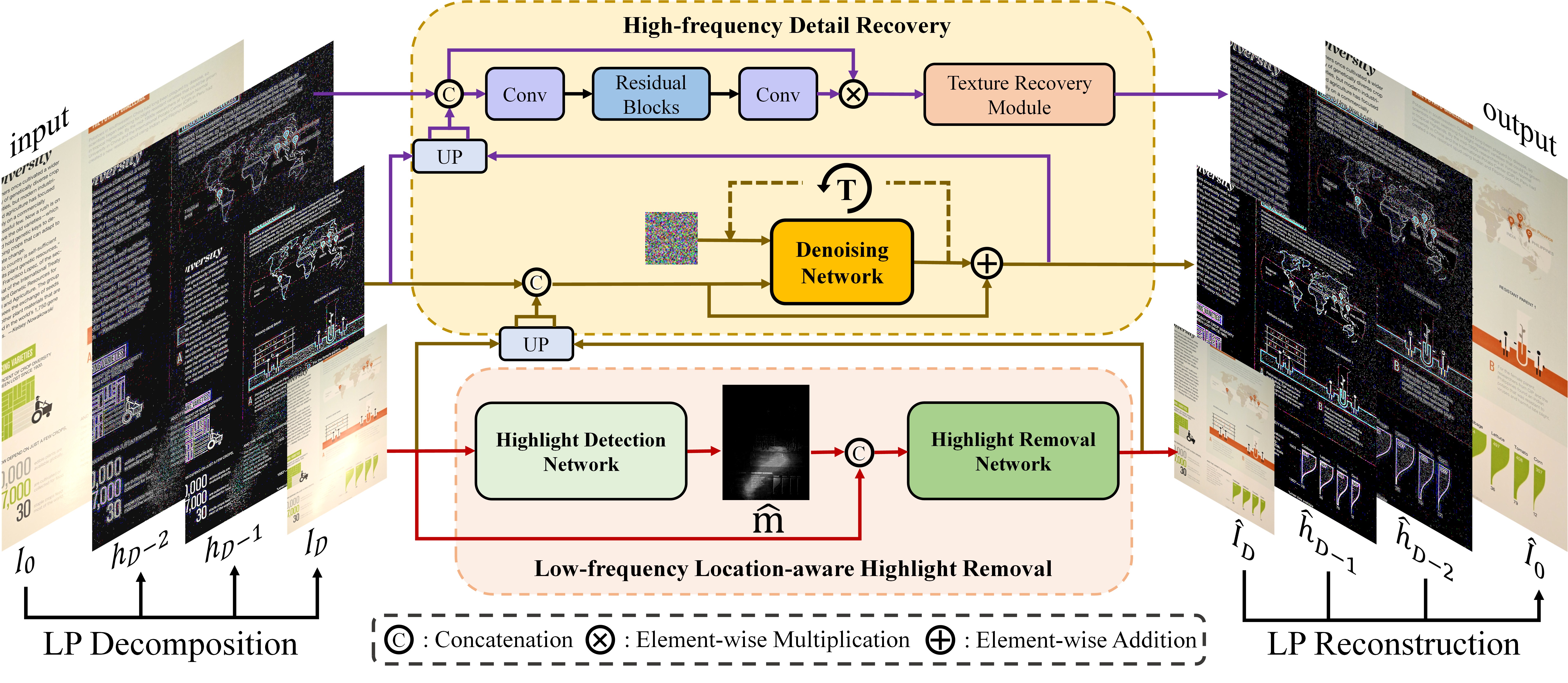}  % 插入图片，并设置宽度为文本宽度的 50%
    \caption{Framework of the proposed L\textsuperscript{2}HRNet. The input image $ I_0 \in \mathbb{R}^{h \times w \times 3}$ is first decomposed into high- and low-frequency bands using the Laplacian pyramid. \textcolor{red}{Red arrows}: For the low-frequency component $I_D \in \mathbb{R}^{\frac{h}{2^D} \times \frac{w}{2^D} \times c}$, highlight location prior is incorporated to achieve global highlight removal. \textcolor{brown}{Brown arrows}: For the high-frequency component $h_{D-1}\in \mathbb{R}^{\frac{h}{2^{D-1}} \times \frac{w}{2^{D-1}} \times c}$, the diffusion-based enhancement module, is applied to restore fine details, such as text edges. \textcolor{darkpurple}{Purple arrows}: Convolution layers, residual blocks, and a texture recovery module further enhance finer details in other high-frequency components (e.g., $h_{D-2}, \dots$). The final highlight-free image $\hat{I}_0 \in \mathbb{R}^{h \times w \times 3}$ is reconstructed by merging the processed high- and low-frequency outputs. In the image example, $D$ is set to 2.} \vspace{-10pt} % 添加图片标题
    \label{fig:framework}  % 为图片设置标签，以便在文中引用
\end{figure*}

In this section, we first present the Highlight Location Prior (HLP), which generates structurally aligned pseudo highlight masks. Then, we describe the structure of our tailored Location-Aware Laplacian Pyramid-based Highlight Removal Network (L\textsuperscript{2}HRNet), which performs global highlight removal in the low-frequency band guided by the estimated highlight mask, and refines fine textural details in the high-frequency band via the Diffusion-based Enhancement Module (DEM).

\begin{table}[h]
\centering
\caption{
Accuracy (ACC) and balanced error rate (BER) of the binarized residual map and state-of-the-art highlight detection results on the RD and SHIQ datasets.
We observe that the residual image with contrast stretching achieves the best performance, closely approximating the ground-truth mask. The best results are in \textbf{bold}.
}
\begin{tabular}{lccc}
\toprule
\textbf{Dataset} & \textbf{Method} & \textbf{ACC}~$\uparrow$ & \textbf{BER}~$\downarrow$ \\
\midrule
\multirow{3}{*}{RD}
& Wu \textit{et al.} \cite{wu2023joint}        & 0.89 & 50.48 \\
& Input + Otsu (baseline)         & 0.37 & 36.3 \\
& Residual + Otsu                 & 0.97 & \textbf{4.58} \\
& \textbf{Residual + Stretch + Otsu} & \textbf{0.98} & 5.79 \\
\midrule
\multirow{5}{*}{SHIQ}
& JSHDR \cite{fu2021multi}        & 0.93  & 5.92 \\
& Wu \textit{et al.} \cite{wu2023joint}        & \textbf{0.97} & 5.92 \\
& Input + Otsu (baseline)         & 0.59 & 21.9 \\
& Residual + Otsu                 & 0.93 & 3.90 \\
& \textbf{Residual + Stretch + Otsu} & \textbf{0.97} & \textbf{2.16} \\
\bottomrule
\end{tabular}
\label{tab:eva_mask}\vspace{-12pt}
\end{table}

\begin{figure}[t]
  \centering

  % ----- 第一行 -----
  \begin{subfigure}[b]{0.19\linewidth}
    \includegraphics[width=\linewidth]{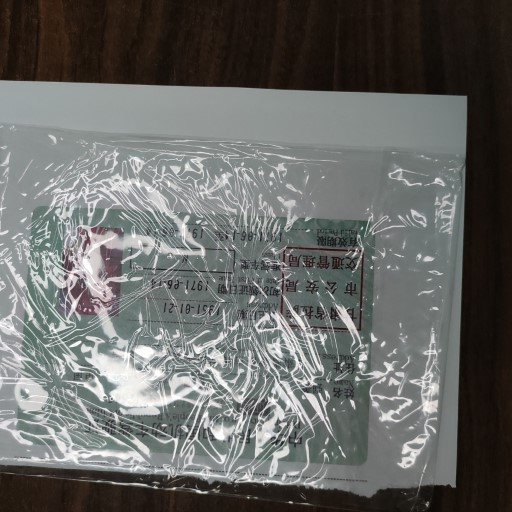}
  \end{subfigure}
  \begin{subfigure}[b]{0.19\linewidth}
    \includegraphics[width=\linewidth]{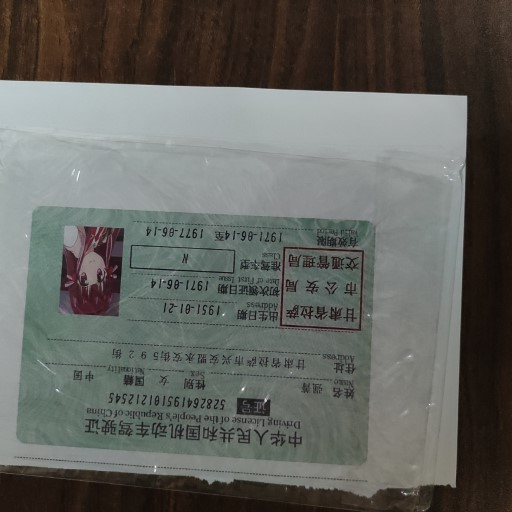}
  \end{subfigure}
  \begin{subfigure}[b]{0.19\linewidth}
    \includegraphics[width=\linewidth]{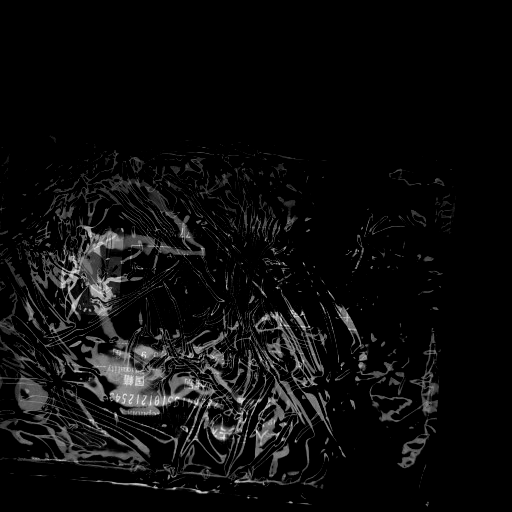}
  \end{subfigure}
  \begin{subfigure}[b]{0.19\linewidth}
    \includegraphics[width=\linewidth]{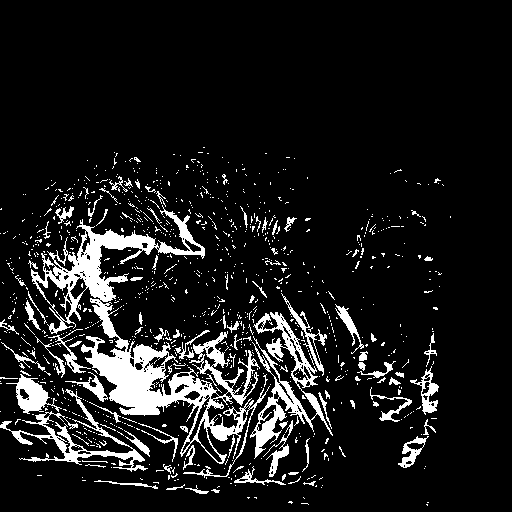}
  \end{subfigure}
  \begin{subfigure}[b]{0.19\linewidth}
    \includegraphics[width=\linewidth]{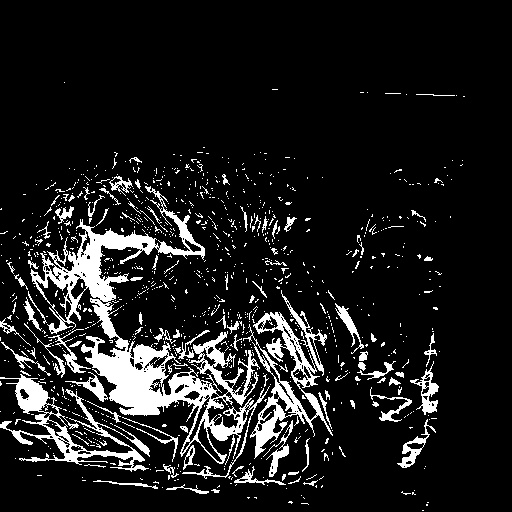}
  \end{subfigure}

  \vspace{1mm}

  % ----- 第二行 -----
  \begin{subfigure}[b]{0.19\linewidth}
    \includegraphics[width=\linewidth]{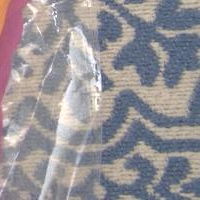}
    \caption*{\footnotesize Input}
  \end{subfigure}
  \begin{subfigure}[b]{0.19\linewidth}
    \includegraphics[width=\linewidth]{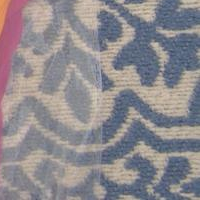}
    \caption*{\footnotesize GT}
  \end{subfigure}
  \begin{subfigure}[b]{0.19\linewidth}
    \includegraphics[width=\linewidth]{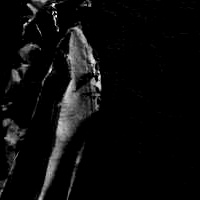}
    \caption*{\footnotesize Residual}
  \end{subfigure}
  \begin{subfigure}[b]{0.19\linewidth}
    \includegraphics[width=\linewidth]{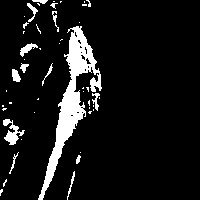}
    \caption*{\footnotesize Otsu Mask}
  \end{subfigure}
  \begin{subfigure}[b]{0.19\linewidth}
    \includegraphics[width=\linewidth]{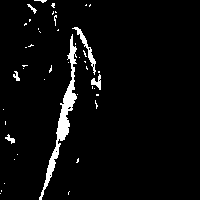}
    \caption*{\footnotesize GT Mask}
  \end{subfigure}

  \caption{
  Visualization of the highlight residual prior on two document images. The top row presents sample from the RD dataset, while the bottom row shows sample from the SHIQ dataset.  The residual map highlights regions of specular reflection, and the corresponding binary mask (Otsu Mask) shows strong structural alignment with ground-truth masks.
  }
  \label{fig:residual_prior_vis}\vspace{-12pt}
\end{figure}

\subsection{Highlight Location Prior}
\label{sec:residual_prior}

Precise localization of specular highlights is critical for effective highlight removal. However, fine-grained manual annotation is both labor-intensive and time-comsuming. To address this, we propose a simple yet effective highlight location prior derived directly from aligned image pairs.

We compute a residual map by subtracting the highlight-free grayscale image from the highlighted one and retaining only the positive values. 
We compute the residual map \textbf{m} as follows:
\begin{align}
\textbf{m} = \max(I_D - I_{gtD}, 0), 
\label{eq:residual_map}
\end{align}
where \(I_{\text{gtD}}\) represents the low-frequency component of the ground truth image.
This residual naturally captures the spatial structure introduced by highlight regions. We apply contrast stretching by clipping the residual map below a lower percentile threshold \(\alpha\), where \(\alpha\) is set to 80\% in our experiments. This suppresses weak background noise and enhances the visibility of strong highlight regions. The resulting grayscale residual is used directly as a soft spatial prior—without any hard thresholding—enabling smooth supervision and preserving boundary transitions.

To assess the effectiveness of the residual prior, we additionally apply Otsu\cite{otsu1975threshold} thresholding to obtain binary highlight masks for evaluation, as the metrics require binary masks. However, during training, we retain the continuous-valued residual map to preserve richer information. For comparison, we also apply Otsu thresholding directly to the original highlight image. %Although our method does not use these binary masks during training, 
This comparison helps demonstrate that even a simple residual map—without any learned model—can achieve strong spatial alignment with ground-truth highlights. Two widely used metrics \cite{fu2021multi, wu2023joint}, accuracy (ACC) and balanced error rate (BER), are used to evaluate the generated masks. As shown in \Cref{tab:eva_mask}, residual-based masks significantly outperform the baseline and even surpass recent learning-based detectors \cite{fu2021multi, wu2023joint} on SHIQ dataset \cite{fu2021multi}. Additionally, as illustrated in \Cref{fig:residual_prior_vis}, the residual map effectively captures the highlight regions, and the resulting binary masks align closely with the ground truth, which further validates the effectiveness of our residual-based approach.

%This design reflects a minimalist principle: the structural difference between two real images is a natural and reliable indicator of highlight location—no annotation, no classifier, just signal. 

% While residual-based analysis has been used in other low-level tasks, we are the first to exploit grayscale residuals between highlight and clean document pairs as a direct and effective spatial prior for highlight localization.

\vspace{-3pt}
\subsection{Laplacian Pyramid}
As illustrated on the left side of \Cref{fig:framework}, we first use the LP to decompose the document image into high-frequency components $[h_0, h_1, \dots, h_{D-1}]$, which preserve textures and edges, and a low-frequency component $I_D$ that retains overall global illumination information. Here, $D$ represents the total number of decomposition levels. Then, we process the low-frequency component $I_D$ with HLP guidance to achieve location-aware highlight removal, resulting in $\hat{I}_D$, shown in the lower middle of \Cref{fig:framework}. Furthermore, the high-frequency components $[h_0, h_1, \dots, h_{D-1}]$ undergo high-frequency-specific processing to restore text clarity and edge details, producing outputs $[\hat{h}_0, \hat{h}_1, \dots, \hat{h}_{D-1}]$, depicted in the upper middle of \Cref{fig:framework}. Finally, we reconstruct the highlight-free image $\hat{I}_0$ by iteratively combining these processed layers, yielding the final output displayed on the right side of \Cref{fig:framework}. 

\vspace{-3pt}
\subsection{Low-frequency Location-Aware Highlight Removal}

As illustrated in \Cref{fig:framework}, we design a two-stage cascaded architecture to remove global highlights in the low-frequency band. The first stage estimates a spatially continuous highlight prior, while the second stage performs spatially guided restoration.

\textbf{Stage 1: Highlight Prior Estimation.}
The spatial prior $\hat{\mathbf{m}}$ is supervised using pseudo masks generated from residual maps between highlight and highlight-free images (see Sec.~\ref{sec:residual_prior}). We use a Highlight Detection Network (HDNet) to predict the soft highlight location prior from the low-frequency input $I_D$: \begin{align} \hat{\mathbf{m}} = \text{HDNet}({I_D}). \end{align}

\textbf{Stage 2: Prior-guided Highlight Removal.}
The estimated prior is concatenated with the low-frequency input and passed to a Highlight Removal Network (HRNet), enabling spatially adaptive highlight removal: \begin{equation} \hat{{I}}_D = \text{HRNet}(\text{Concat}([{I_D}, \hat{\mathbf{m}}])). \end{equation}

This two-stage approach effectively estimates and utilizes the prior to guide highlight removal in the low-frequency band.

%Both HDNet and HRNet adopt the Laplacian pyramid-based restoration backbone proposed in \cite{li2023high}, which provides a lightweight yet effective framework for high-resolution document enhancement.

\vspace{-3pt}
\subsection{High-frequency Detail Recovery}

While our cascade network can effectively achieve global illumination recovery, fine-grained text edge informations are missing in the high-frequency band, which are important for OCR and text recognition by human. Therefore, in the high-frequency band, we employ Diffusion-based Enhancement Module (DEM), which has shown effectiveness in restoring fine document details in previous works \cite{yang2023docdiff, cicchetti2024naf}, to accurately recover high-frequency elements. Additionally, to optimize memory usage and improve inference speed, we apply the diffusion model only to the high-frequency component $h_{D-1}$, which has a manageable resolution. The process is illustrated by the brown arrows in \Cref{fig:framework}. 

The diffusion model loss $L_{DM}$ is defined as:
\begin{align}
    L_{DM} = \mathbb{E} \left\| x_0 - f_\theta \left( \sqrt{\bar{\alpha}_t} x_{\text{0}} + \sqrt{1 - \bar{\alpha}_t} \epsilon, t, y \right) \right\|_2^2,
\end{align}
where $x_0 = {h}^{gt}_{D-1} - h_{D-1}$ represents the target residual with ${h}^{gt}_{D-1}$ denoting the LP decomposition component from layer $D-1$ of the ground truth image. The conditioning input $y = [h_{D-1}; \text{Up}(I_D);\\ \text{Up}(\hat{I}_D)]$, where $\text{Up}(\cdot)$ denotes the upsampling operation. $f_\theta$ is the denoising network that predicts $x_0$, $\bar{\alpha}_t = \prod_{s=1}^t \alpha_s$ is the accumulated noise scaling factor, $\epsilon \sim \mathcal{N}(0, \mathbf{I})$ is Gaussian noise, and $t$ denotes the timestep in the diffusion process. It is worth noting that directly predicting $x_0$ can improve detail recovery quality but slightly reduces diversity according to \cite{yang2023docdiff, cicchetti2024naf}. However, this is acceptable for highlight removal tasks since it is the recovery task instead of the creation task. To further increase efficiency, we utilize the DPM-Solver \cite{lu2022dpm} to minimize the number of iterations in the reverse process, enabling faster convergence without compromising high-frequency detail recovery.
For other high-frequency components (e.g., $h_{D-2}, h_{D-3}, \dots$), we progressively refine details through convolution layers, residual blocks, and a texture recovery module, as shown by the purple arrows in \Cref{fig:framework}.

\vspace{-3pt}
\subsection{Loss Function}

HDNet is trained in a supervised manner using the pseudo mask $\mathbf{m}$ generated by HLP:
\begin{equation}
\mathcal{L}_{\text{mask}} = \left\| \mathbf{m} - \hat{\mathbf{m}} \right\|_1 + \beta_1 \cdot \mathcal{L}_{\text{TV}}(\hat{\mathbf{m}}),
\end{equation}
where $\beta_1$ represents the balancing weight for the Total Variation loss~\cite{rudin1992nonlinear}, which could smooth predictions and reduce artifacts and $\beta_1$ is chosen as 0.00005 in our experiment.

To preserve the original details, we use the Mean Squared Error loss ($L_{MSE}$) \cite{ren2016faster}, and Structural Similarity Index Measure loss ($L_{SSIM}$) \cite{wang2004image} to compare the recovered image $\hat{I}_0$ with the ground truth image $I_{\text{gt}}$. The diffusion loss $L_{DM}$ is utilized to enhance high-frequency detail recovery. Additionally, we incorporate a structure consistency loss $L_{\text{structure}}$ \cite{simonyan2014very} to ensure structural consistency:
\begin{align} 
L_{\text{structure}} = \left\| \text{VGG}(I_{\text{gt}}) - \text{VGG}(\hat{I}_0) \right\|_2^2, 
\end{align}
where \text{VGG(·)} is the feature extractor from the pre-trained VGG19 model. The total loss function can be expressed as follows:
\begin{align}
L_{\text{total}} = L_{MSE} + \lambda_{1} L_{SSIM} + \lambda_{2} L_{DM} + \lambda_{3} L_{\text{structure}} + \lambda_{4} L_{\text{mask}},
\end{align}
where $\lambda_{1}$, $\lambda_{2}$, $\lambda_{3}$, $\lambda_{4}$ are the weights to balance each item.

\label{sec:method}

\vspace{-3pt}
\section{Experiment}

\label{sec:experiment}

%-------------------------------------------------------------------------

\begin{figure*}[tp]  
    \begin{subfigure}[b]{0.138\textwidth}
        \centering
        \includegraphics[angle=90, width=\linewidth]{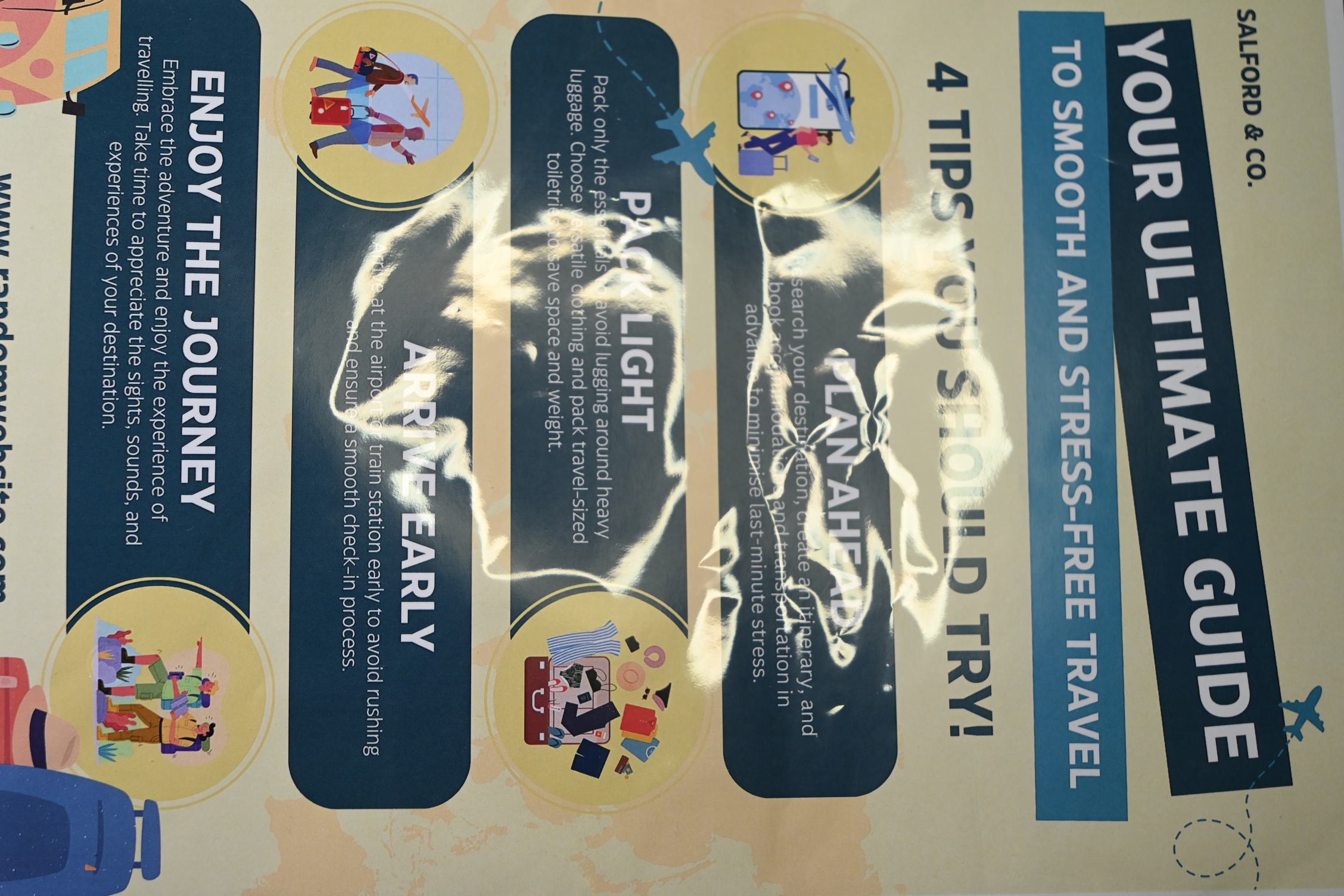} % 替换为实际文件路径
    \end{subfigure}
    \begin{subfigure}[b]{0.138\textwidth}
        \centering
        \includegraphics[angle=90, width=\linewidth]{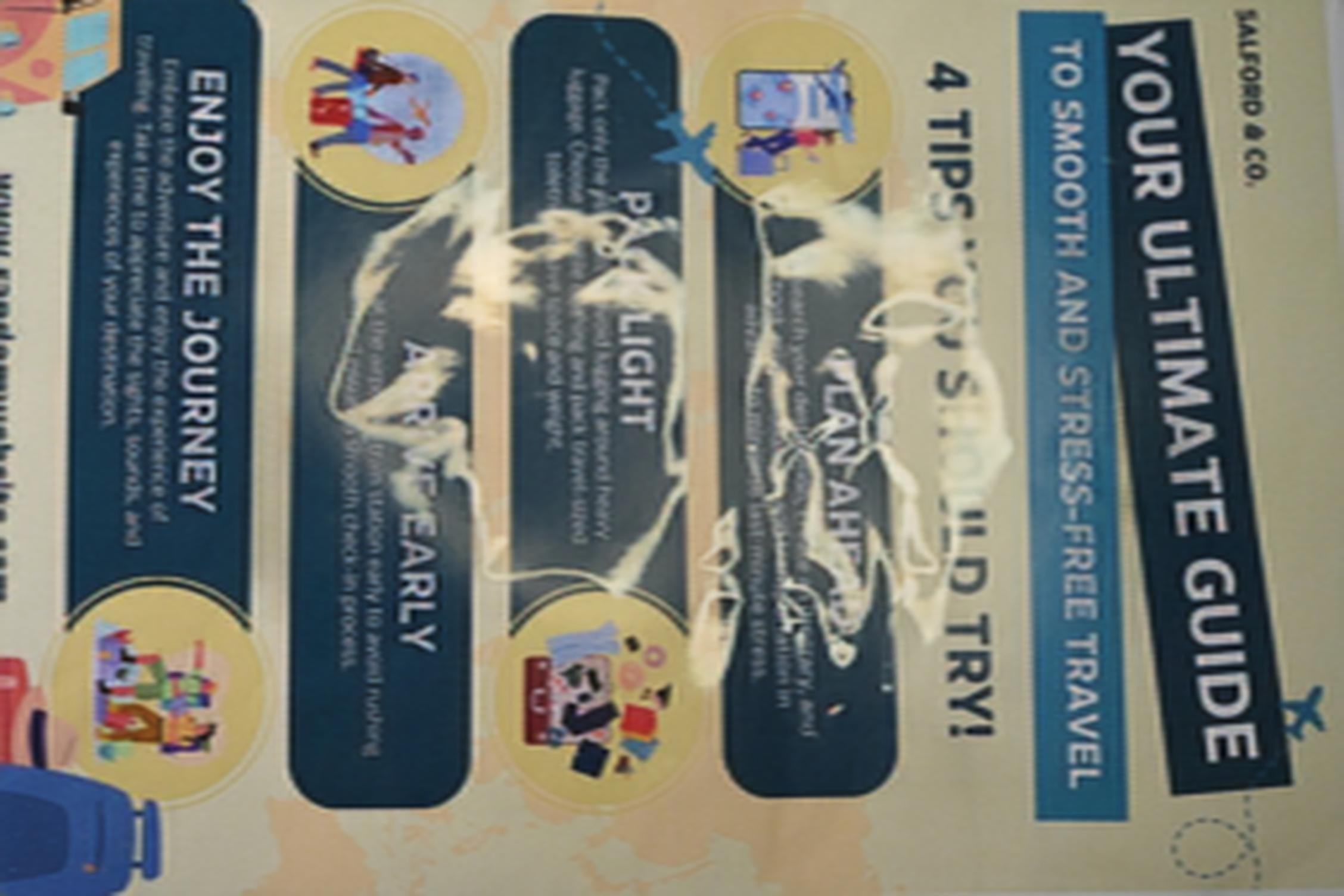} % 替换为实际文件路径
    \end{subfigure}
    \begin{subfigure}[b]{0.138\textwidth}
        \centering
        \includegraphics[angle=90, width=\linewidth]{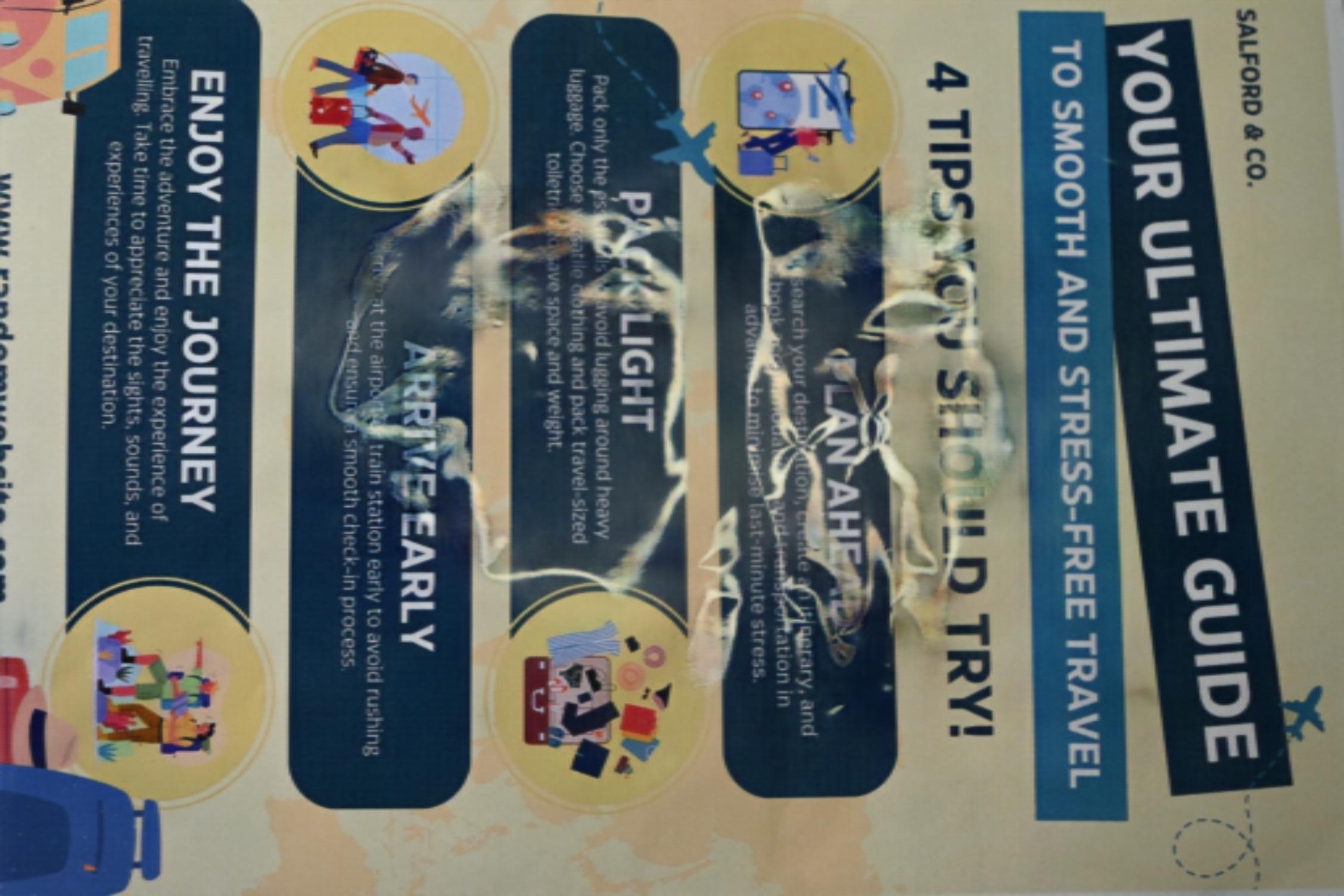} % 替换为实际文件路径
    \end{subfigure}
    \begin{subfigure}[b]{0.138\textwidth}
        \centering
        \includegraphics[angle=90, width=\linewidth]{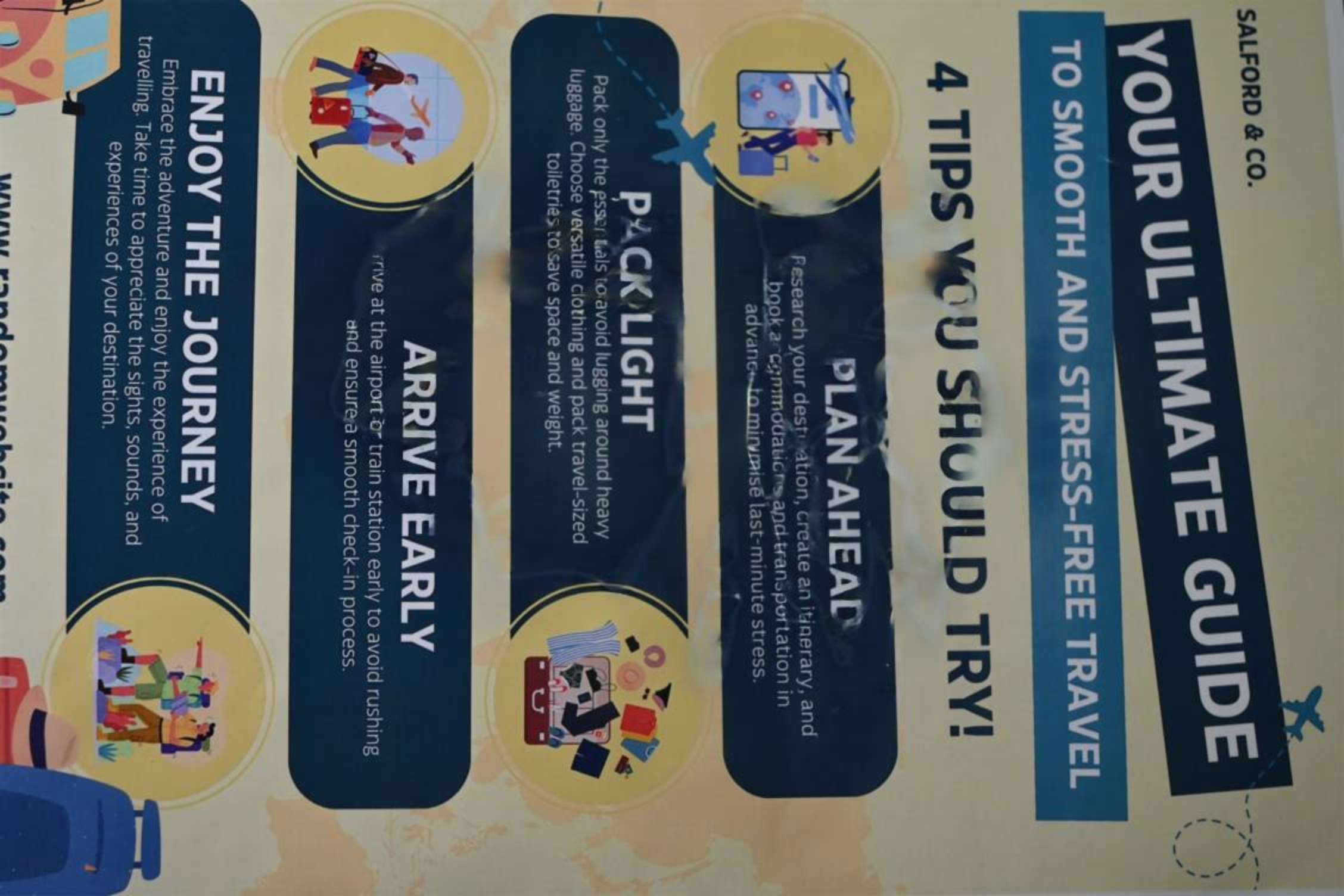} % 替换为实际文件路径
    \end{subfigure}
    \begin{subfigure}[b]{0.138\textwidth}
        \centering
        \includegraphics[angle=90, width=\linewidth]{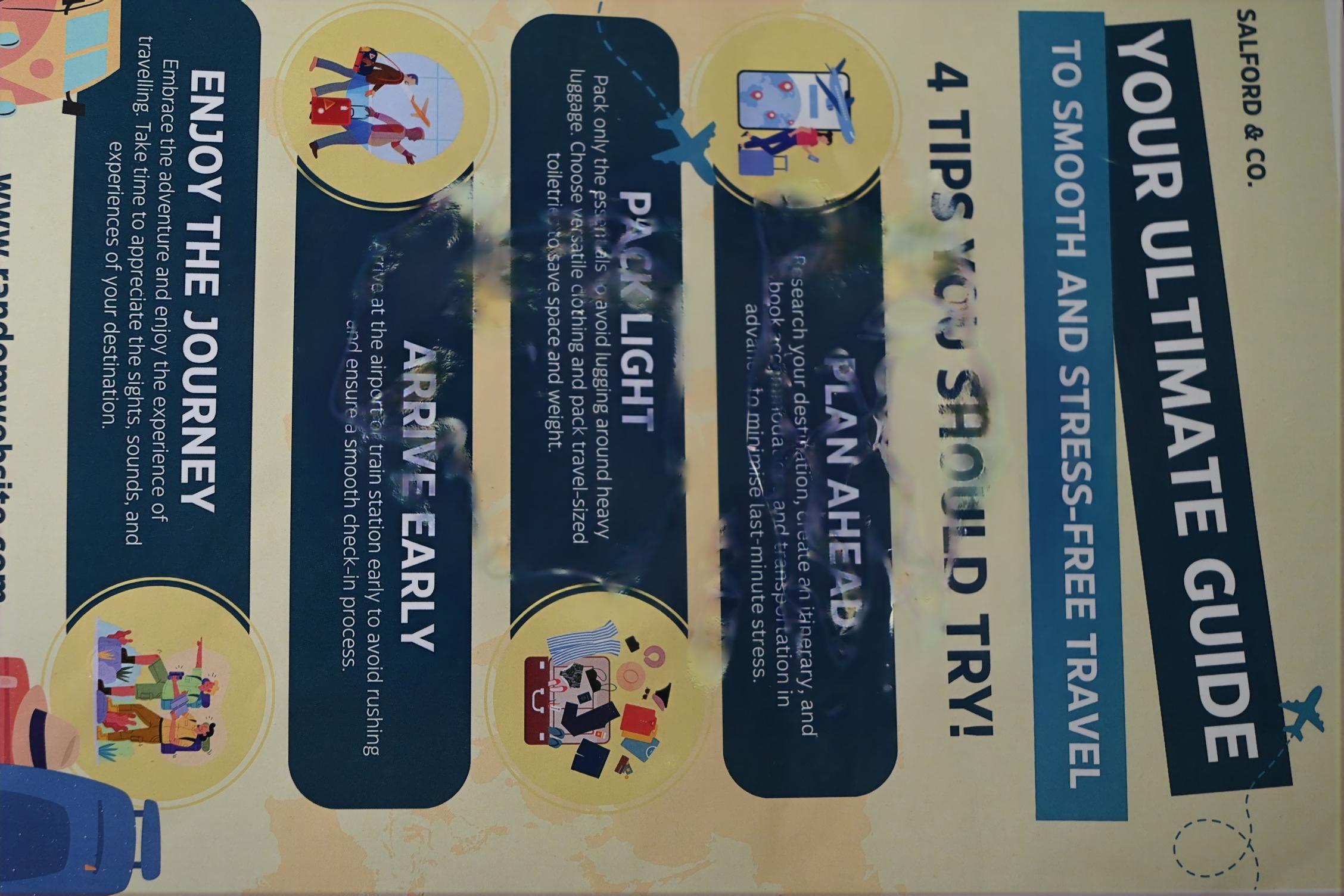} % 替换为实际文件路径
    \end{subfigure}
    \begin{subfigure}[b]{0.138\textwidth}
        \centering
        \includegraphics[angle=90, width=\linewidth]{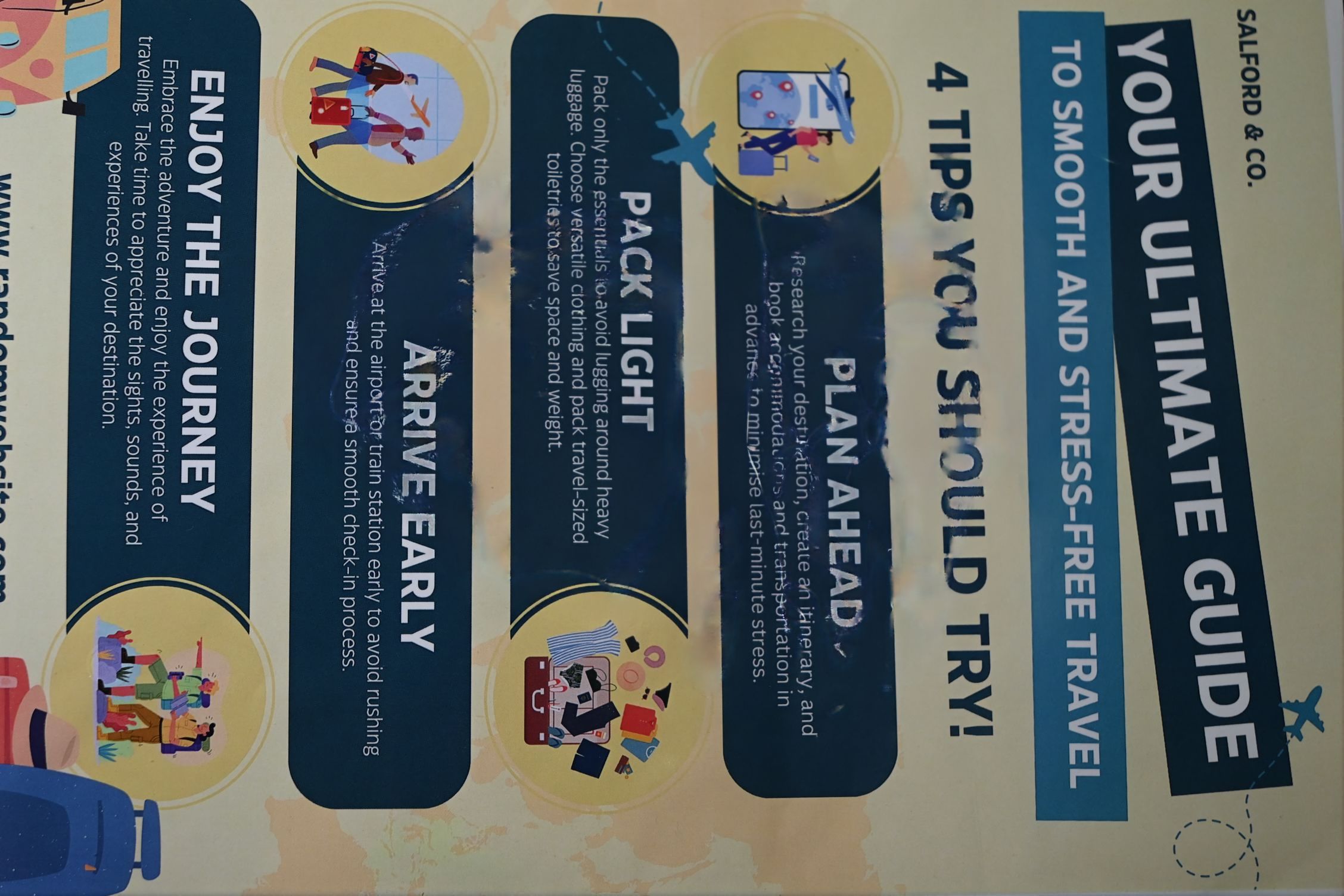} % 替换为实际文件路径
    \end{subfigure}
    \begin{subfigure}[b]{0.138\textwidth}
        \centering
        \includegraphics[angle=90, width=\linewidth]{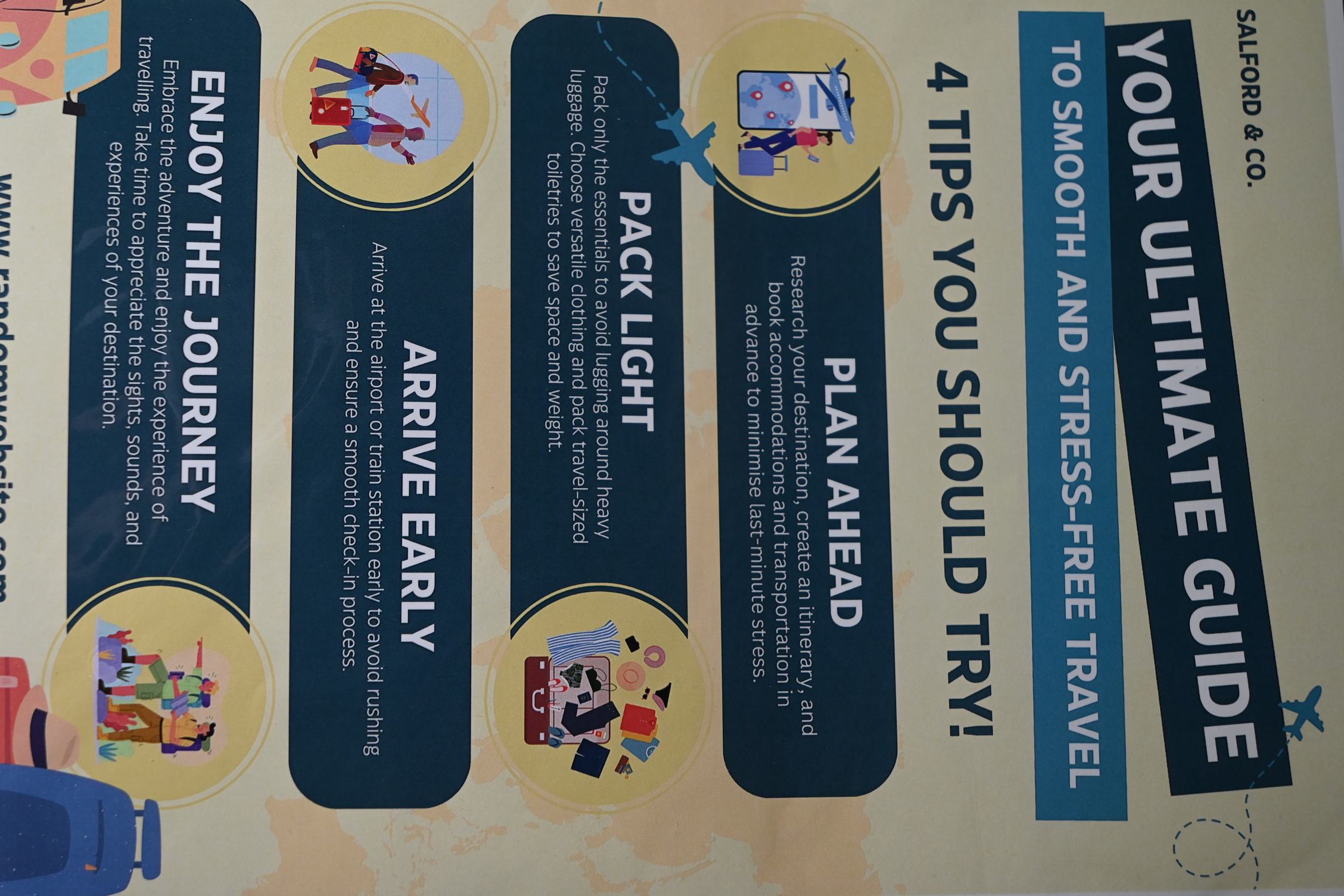} % 替换为实际文件路径
    \end{subfigure} \\
    
%-------------------------------------------------------------------------
    % the third row   
    \begin{subfigure}[b]{0.138\textwidth}
        \centering
        \includegraphics[angle=90, width=\linewidth]{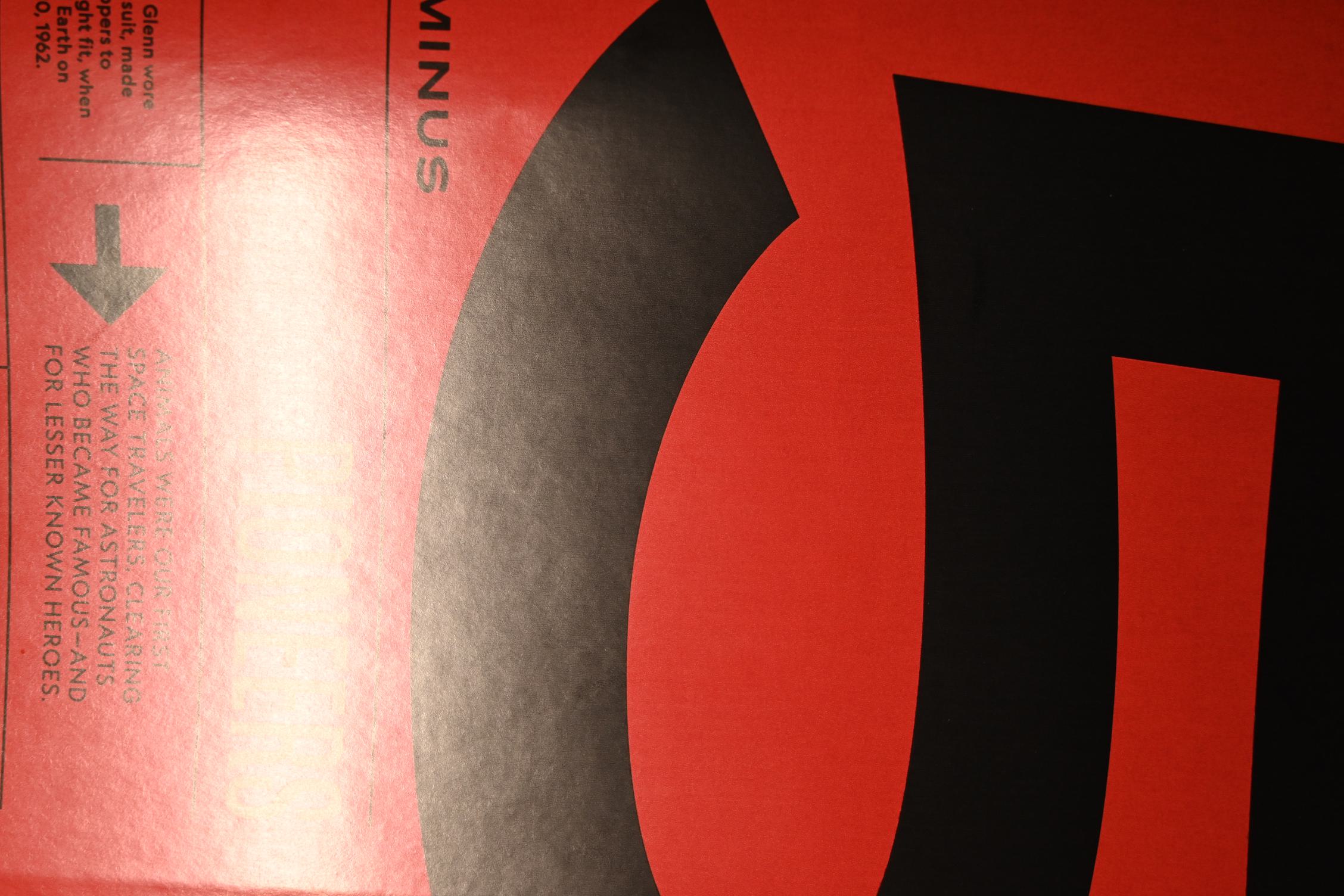}
        % \caption*{Input}
    \end{subfigure}
    \begin{subfigure}[b]{0.138\textwidth}
        \centering
        \includegraphics[angle=90, width=\linewidth]{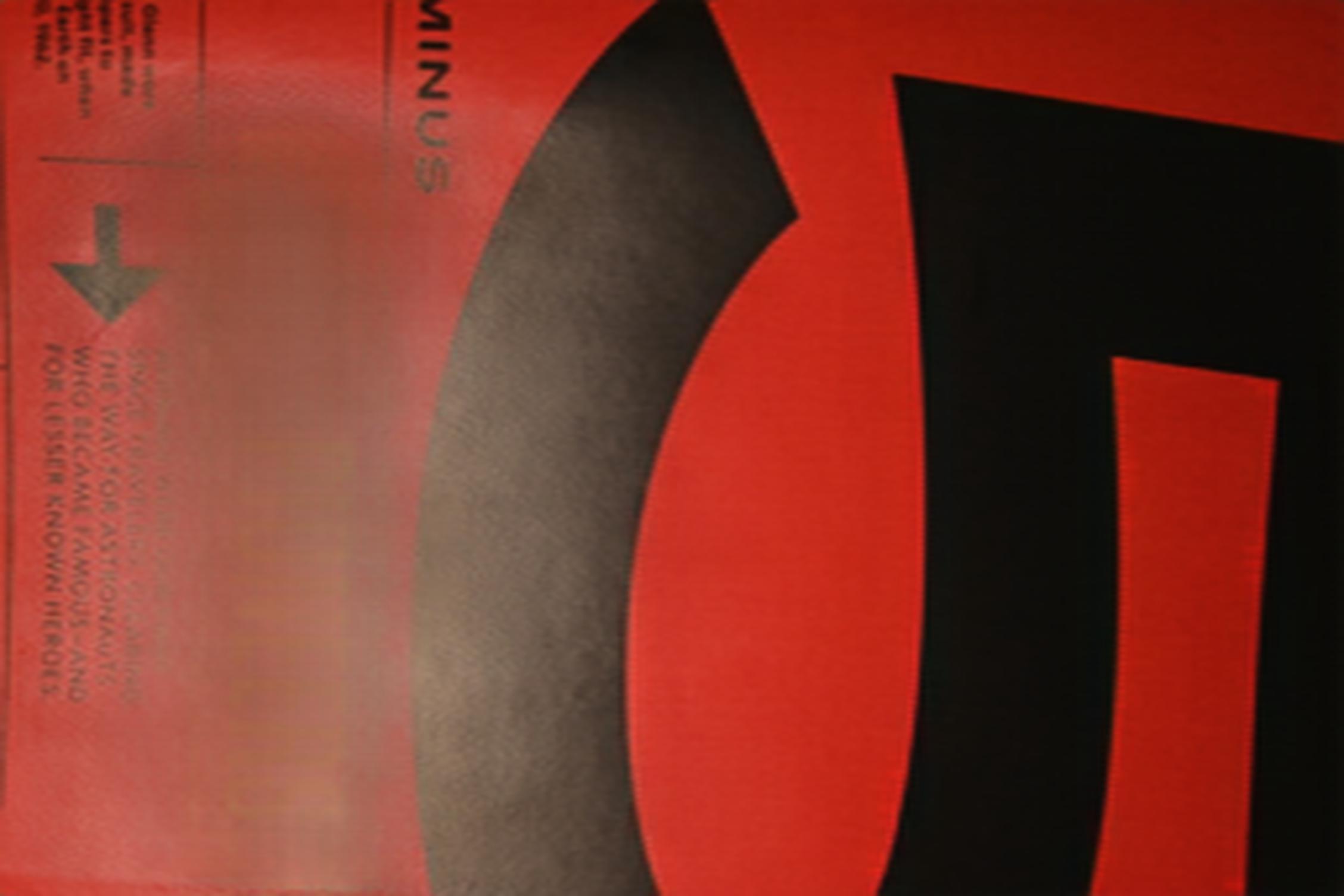}
        % \caption{JSHDR}
    \end{subfigure}
    \begin{subfigure}[b]{0.138\textwidth}
        \centering
        \includegraphics[angle=90, width=\linewidth]{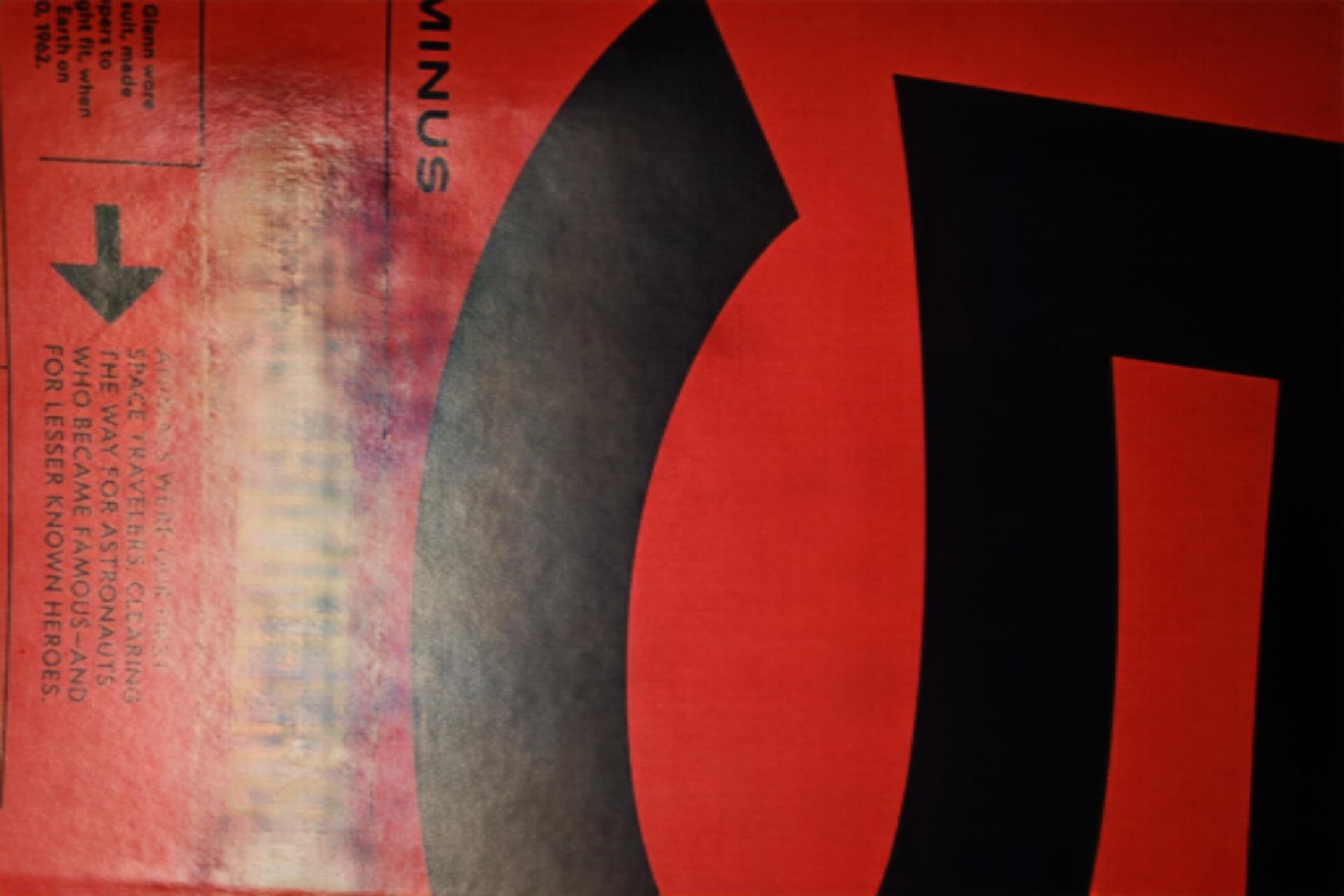}
        % \caption{TSHRNet}
    \end{subfigure}
    \begin{subfigure}[b]{0.138\textwidth}
        \centering
        \includegraphics[angle=90, width=\linewidth]{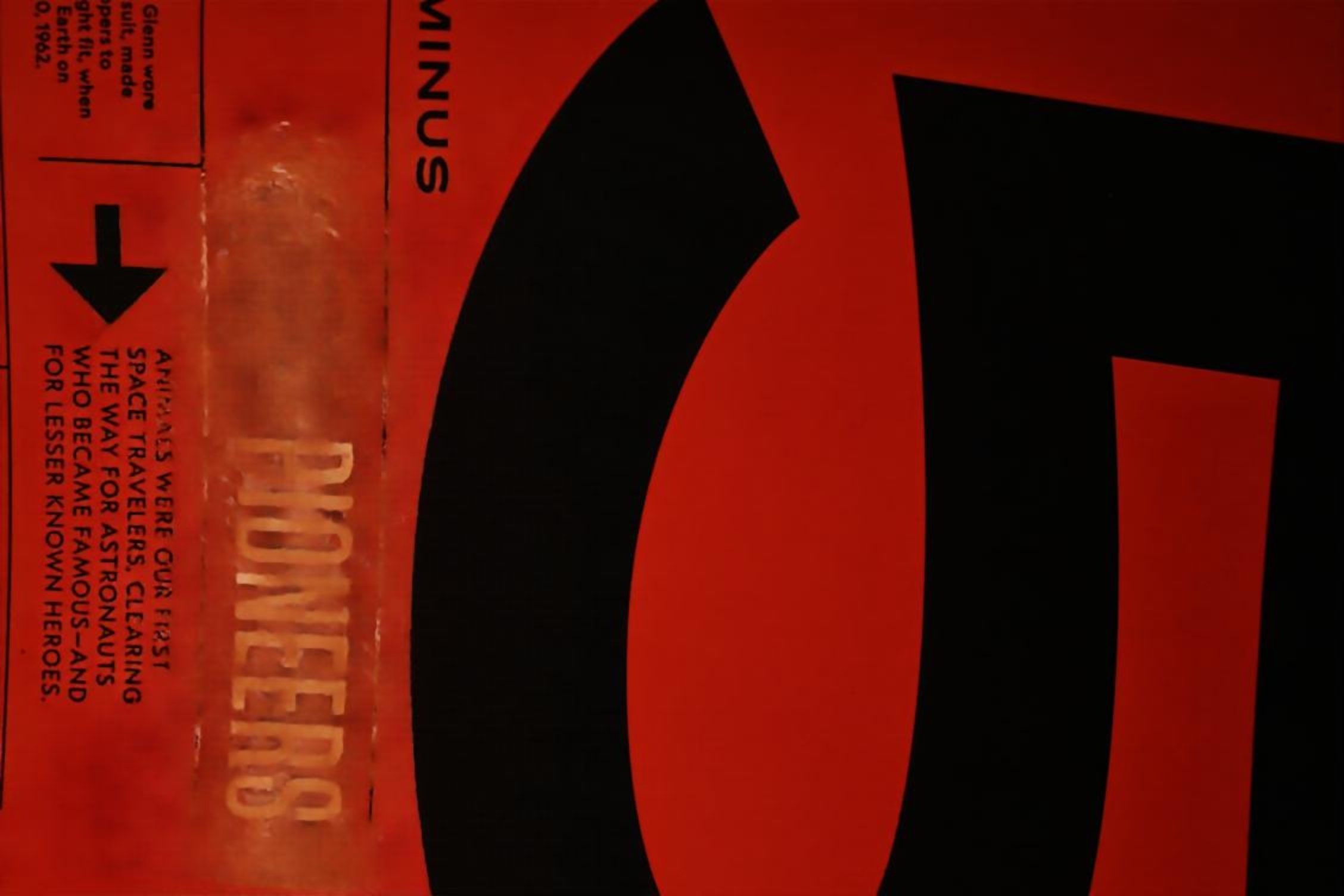}
        % \caption{DHAN-SHR}
    \end{subfigure}
    \begin{subfigure}[b]{0.138\textwidth}
        \centering
        \includegraphics[angle=90, width=\linewidth]{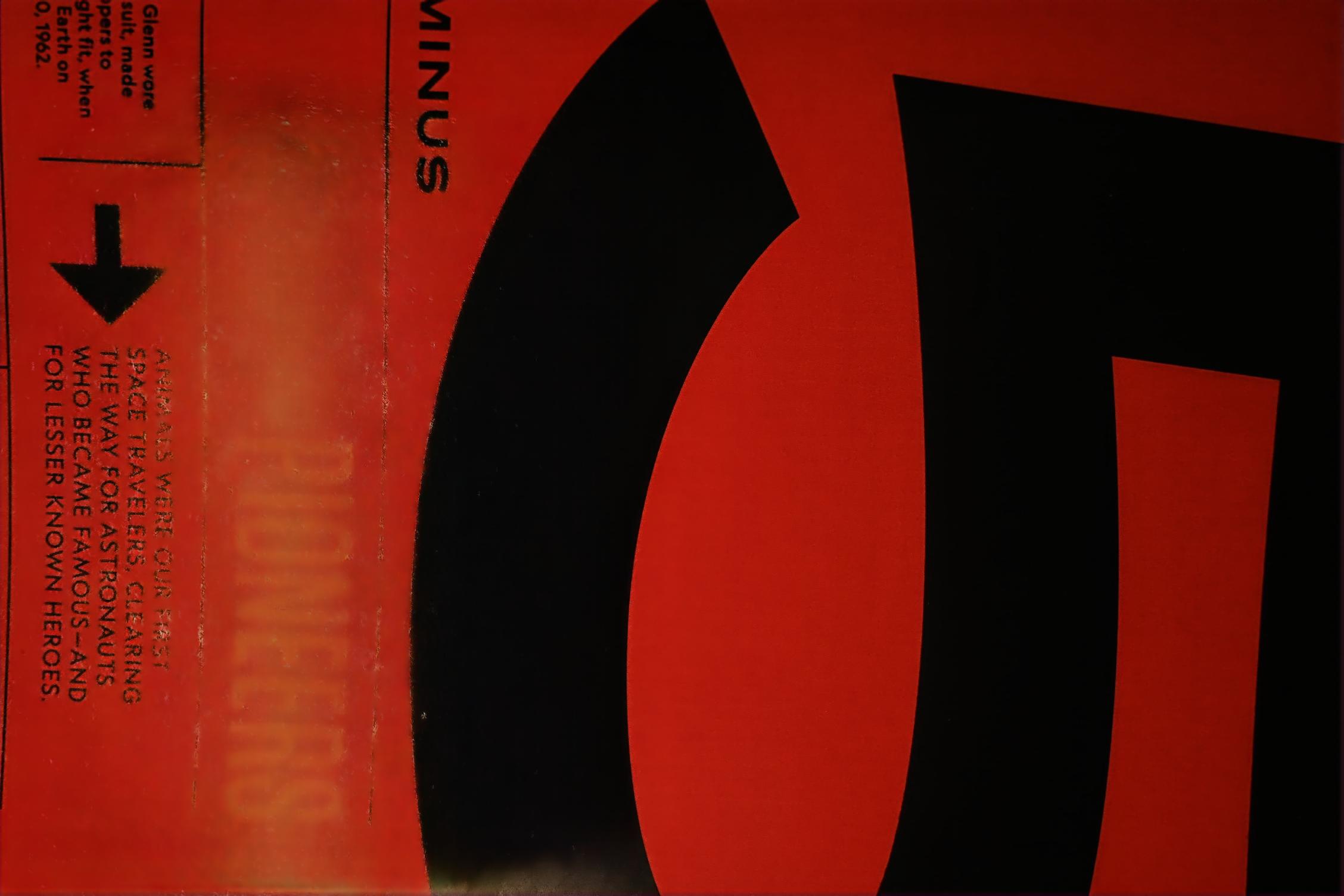}
        % \caption{DocShadowNet}
    \end{subfigure}
    \begin{subfigure}[b]{0.138\textwidth}
        \centering
        \includegraphics[angle=90, width=\linewidth]{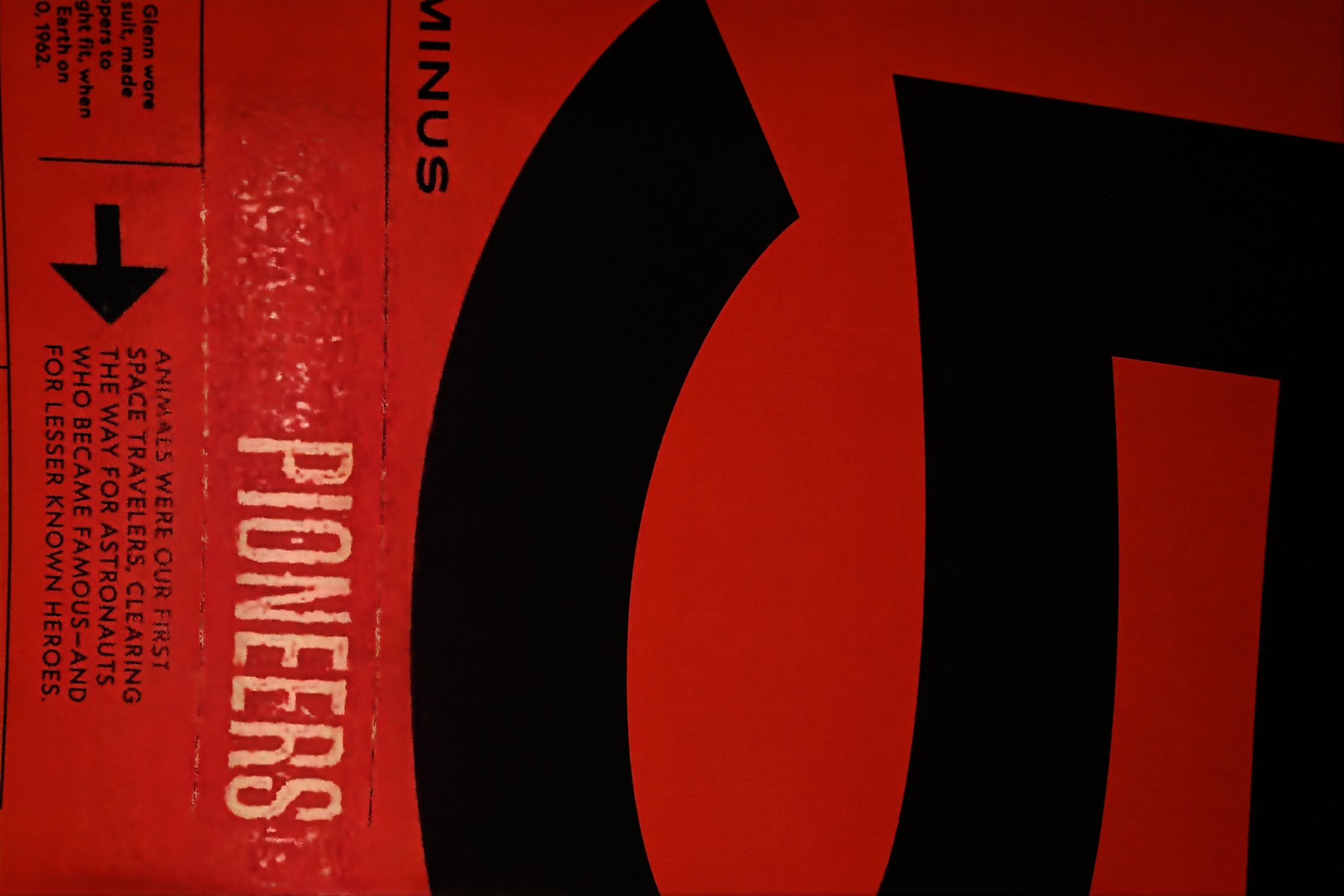}
        % \caption*{Ours}
    \end{subfigure}
        \begin{subfigure}[b]{0.138\textwidth}
        \centering
        \includegraphics[angle=90, width=\linewidth]{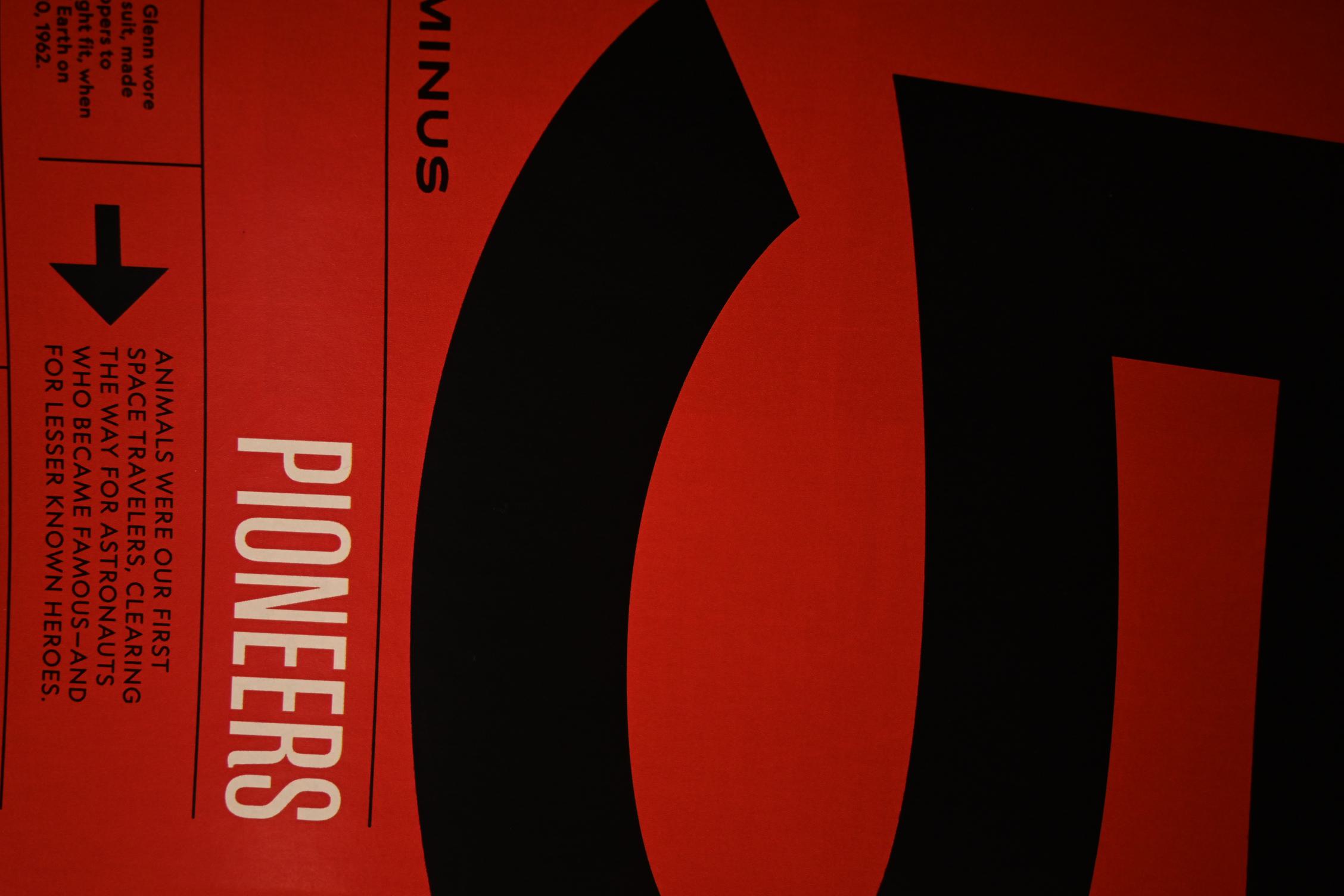}
        % \caption*{Reference}
    \end{subfigure} \\
%-------------------------------------------------------------------------

    %the fourth row  
    \begin{subfigure}[b]{0.138\textwidth}
        \centering
        \includegraphics[angle=0, width=\linewidth]{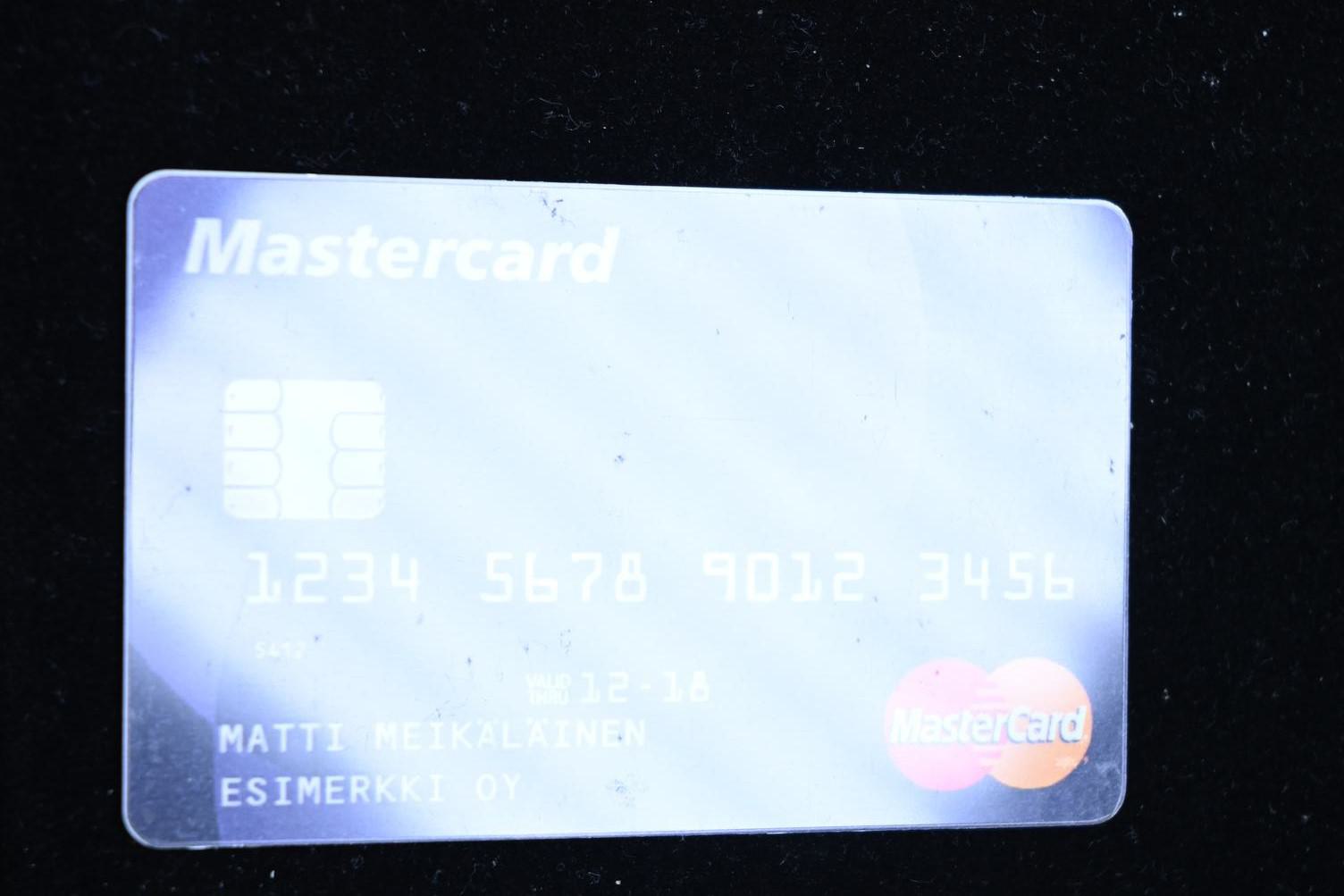}
        \caption*{Input}
    \end{subfigure}
    \begin{subfigure}[b]{0.138\textwidth}
        \centering
        \includegraphics[angle=0, width=\linewidth]{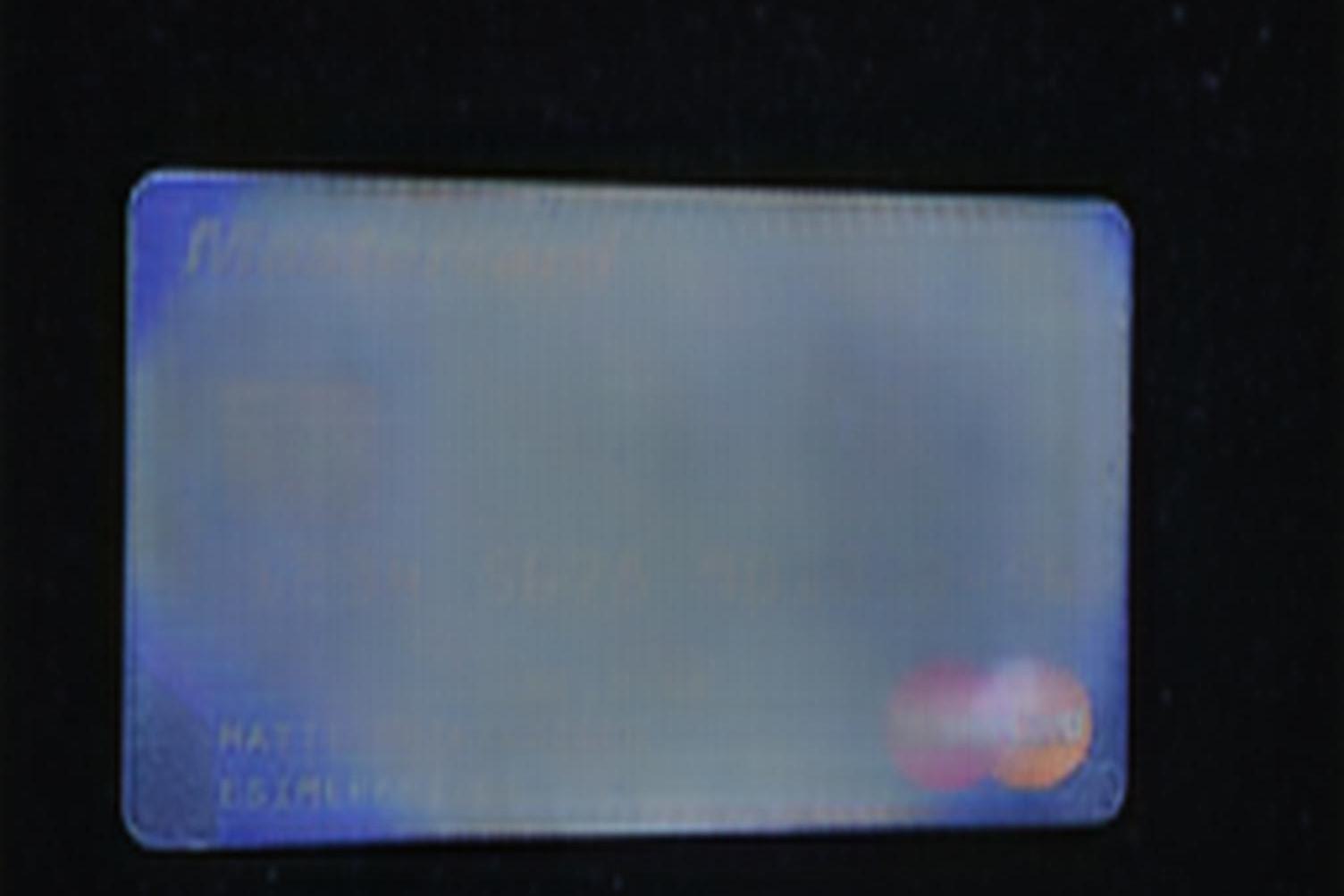}
        \caption{JSHDR}
    \end{subfigure}
    \begin{subfigure}[b]{0.138\textwidth}
        \centering
        \includegraphics[angle=0, width=\linewidth]{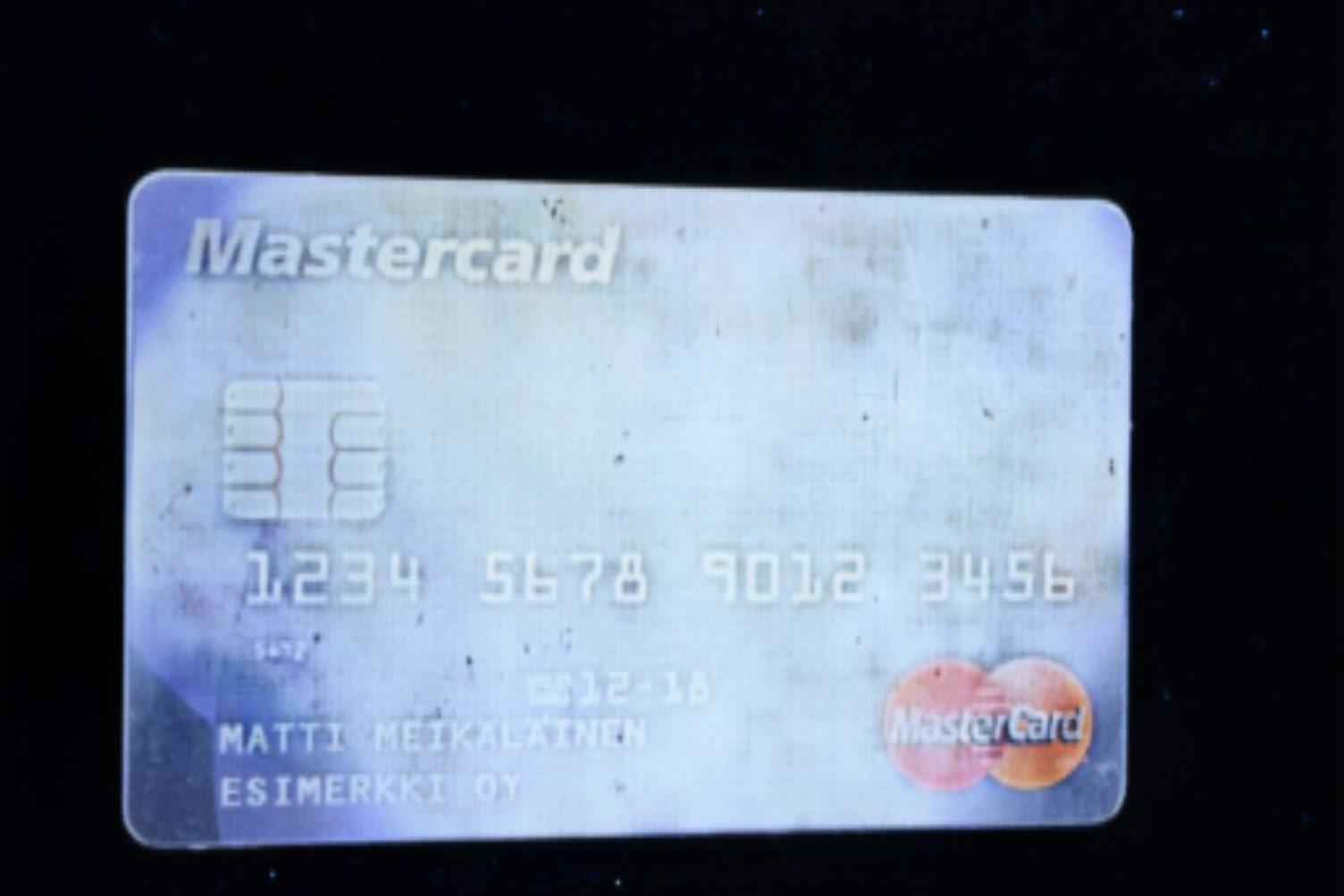}
        \caption{TSHRNet}
    \end{subfigure}
    \begin{subfigure}[b]{0.138\textwidth}
        \centering
        \includegraphics[angle=0, width=\linewidth]{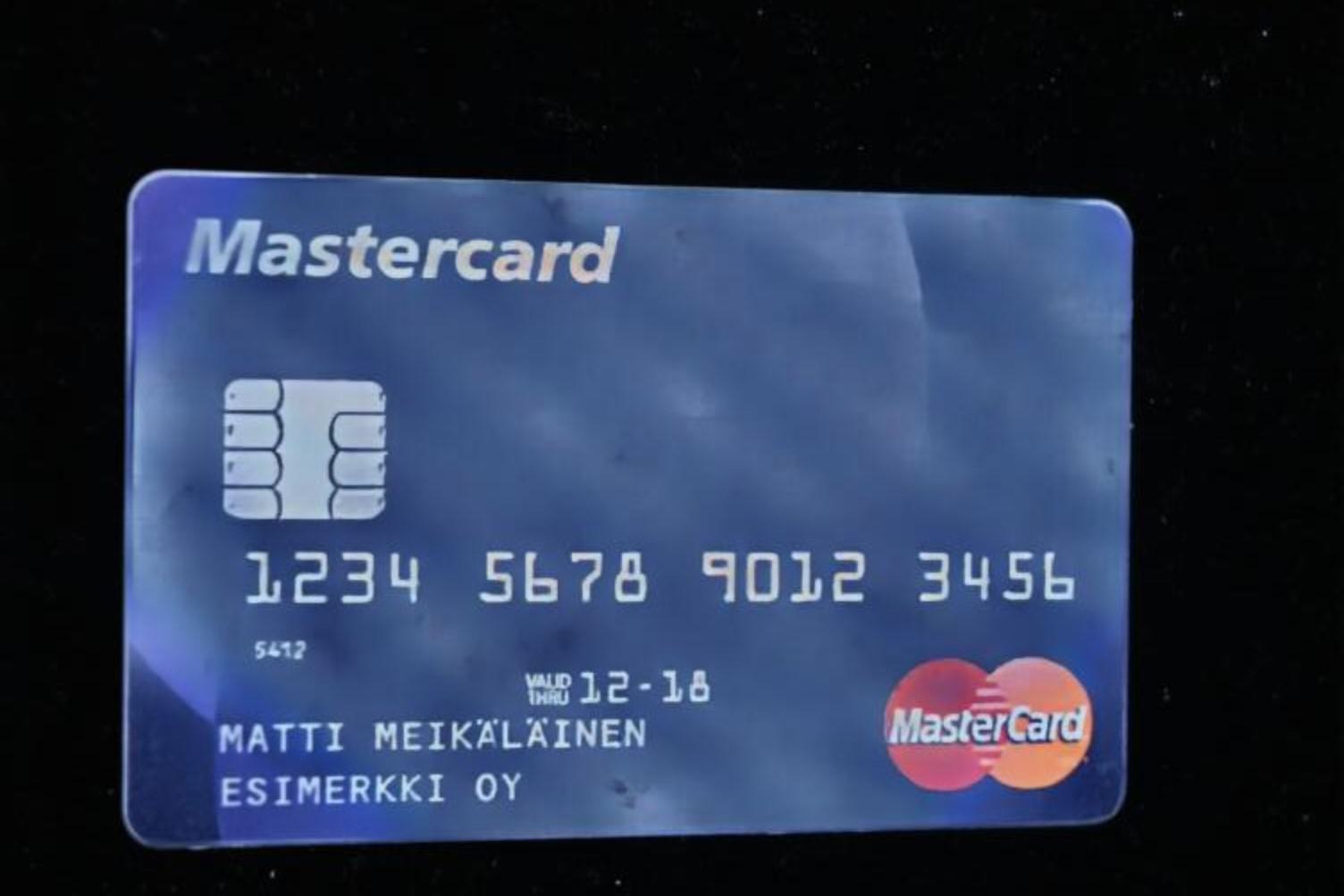}
        \caption{DHAN-SHR}
    \end{subfigure}
    \begin{subfigure}[b]{0.138\textwidth}
        \centering
        \includegraphics[angle=0, width=\linewidth]{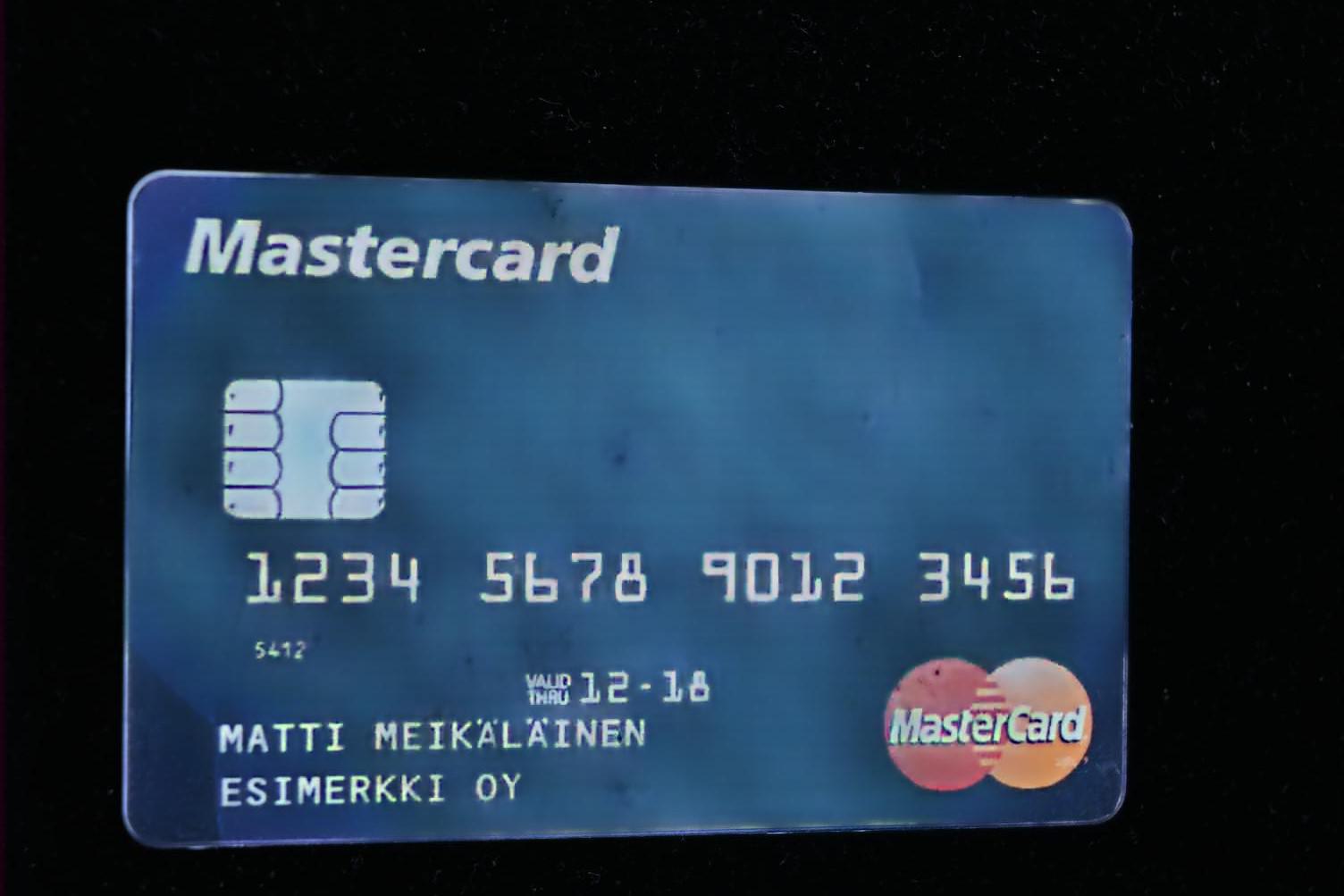}
         \caption{DocShadowNet}
    \end{subfigure}
    \begin{subfigure}[b]{0.138\textwidth}
        \centering
        \includegraphics[angle=0, width=\linewidth]{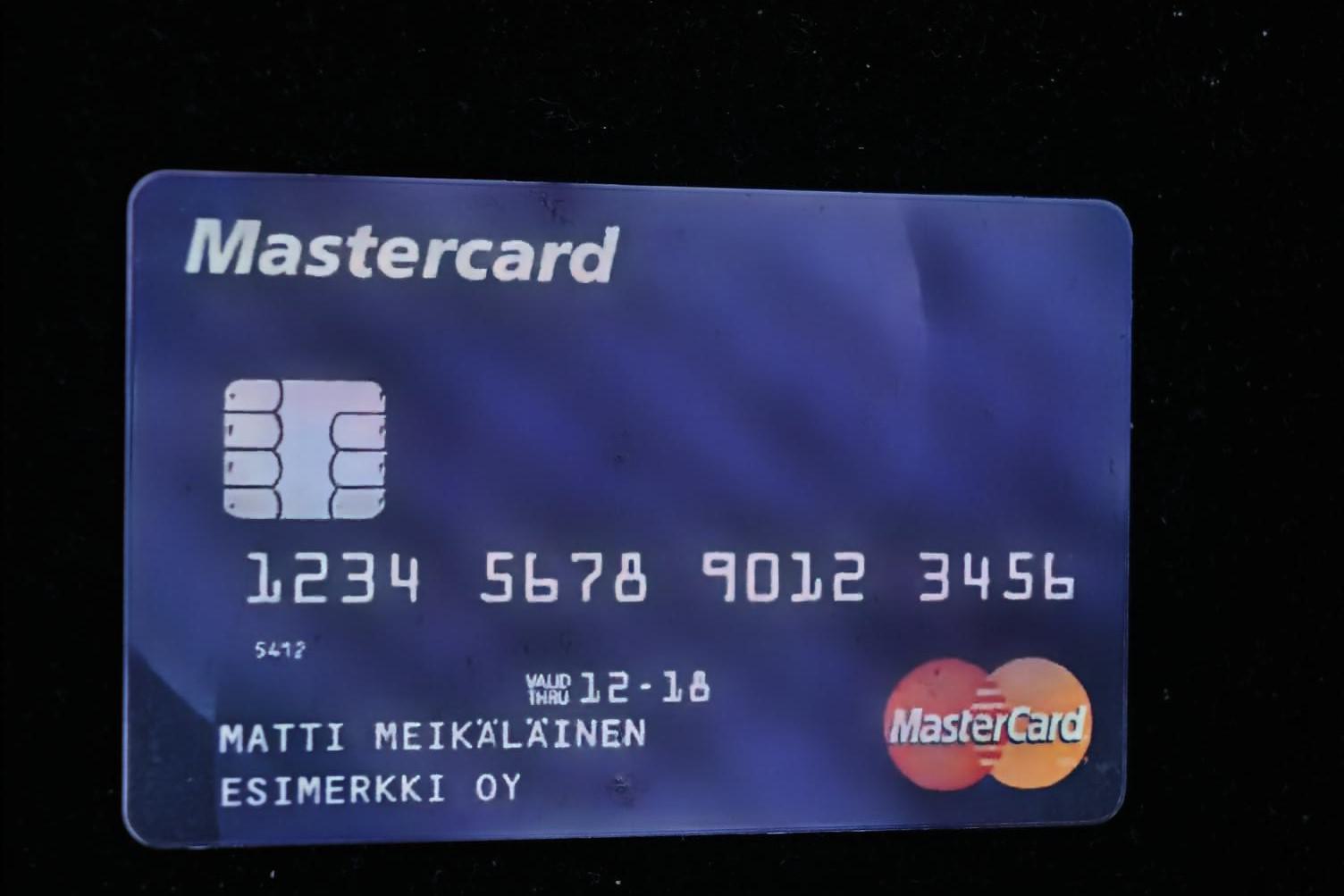}
        \caption*{Ours}
    \end{subfigure}
        \begin{subfigure}[b]{0.138\textwidth}
        \centering
        \includegraphics[angle=0, width=\linewidth]{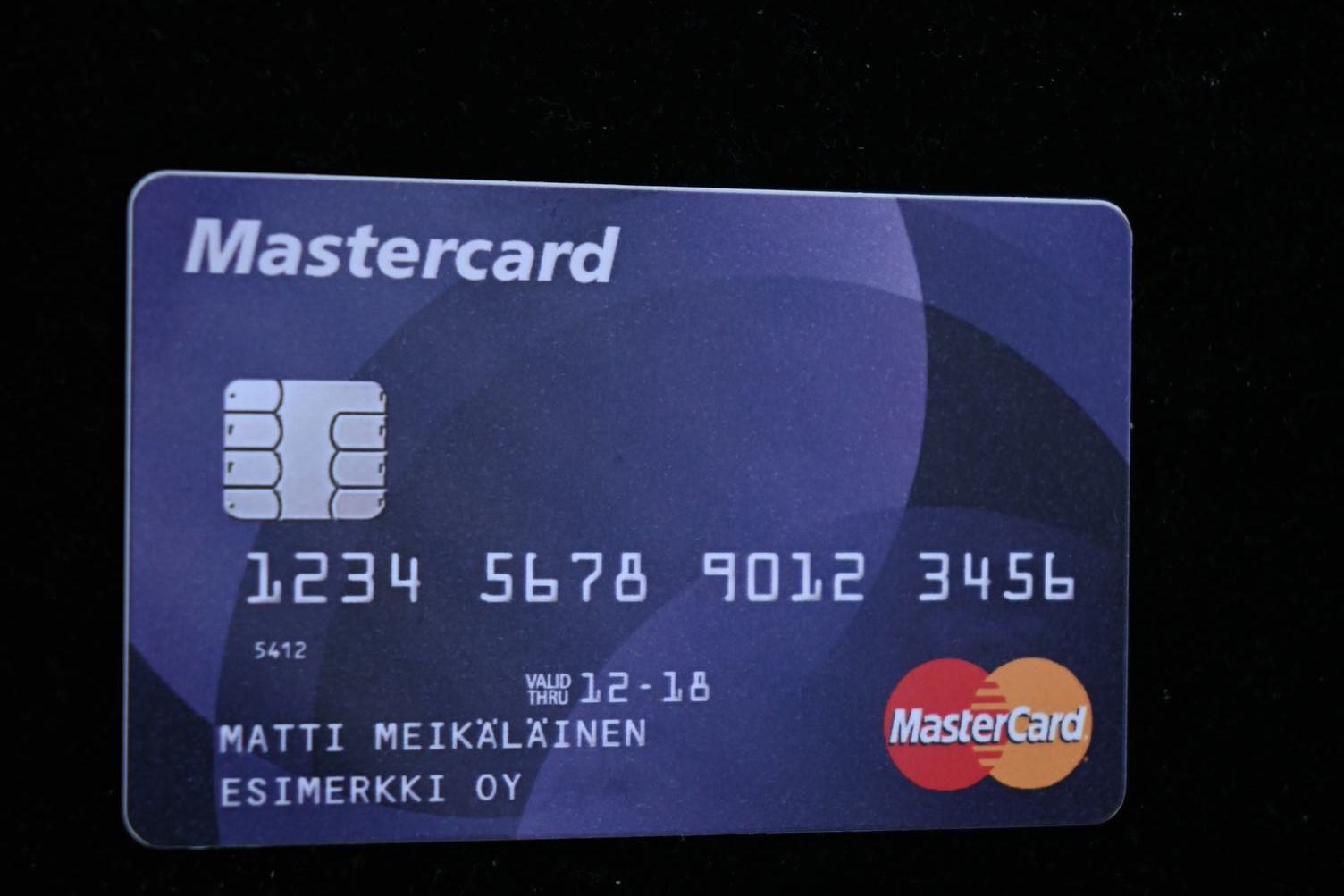}
        \caption*{GT}
    \end{subfigure} \\[-12pt]

    \caption{Qualitative comparisons of document highlight removal on the DocHR14K dataset. From left to right: the input highlight image, the estimated results of (a) JSHDR \cite{fu2021multi}, (b) TSHRNet \cite{fu2023towards}, (c) DHAN-SHR \cite{guo2024dual}, (d) DocShadowNet \cite{li2023high}, ours, and the ground truth image, respectively. Zoom in for the best view.} \vspace{-4pt}
    \label{fig:Qualitatively Comparison}
\end{figure*}
%-------------------------------------------------------------------------

\subsection{ Experimental Setups}
\noindent \textbf{Implementation Details.} Our network is implemented using PyTorch and trained on six NVIDIA RTX 4090 GPUs for a duration of two and a half days. We initialize the learning rate at $2 \times {10}^{-4}$ and set the batch size to $6$.  We train the L\textsuperscript{2}HRNet for an extensive 1000 epochs in an end-to-end manner. The coefficients of our network, $\lambda_{1}$, $\lambda_{2}$, $\lambda_{3}$ and $\lambda_{4}$ are experimentally determined to be 0.4, 1.0, 0.1 and 0.5, respectively. For the denoising network $f_\theta$, we use a lightweight variant of nonlinear activation-free network \cite{cicchetti2024naf, chen2022simple}. To enhance robustness and generalization, we apply random flips and rotations for data augmentation.

\noindent \textbf{Benchmark Datasets and Evaluation Metrics.}
%Apart from our DocHR14K dataset, We work with two benchmark datasets for highlight removal experiments
In addition to DocHR14K, we conduct highlight removal experiments on two benchmark datasets: (1)
RD~\cite{liang2021high} dataset includes 1800 training and 255 testing pairs (highlight images and highlight-free images).
(2) SHIQ dataset \cite{fu2021multi}, a widely used benchmark for image specular highlight removal, which comprises of 9825 training and 1000 testing pairs of highlight and highlight-free
images. 
All methods are evaluated  with three common-used full-reference metrics, \textit{i.e.} PSNR, SSIM \cite{wang2004image}, and RMSE. PSNR focuses on pixel-level accuracy, SSIM emphasizes perceptual quality and structural similarity, and RMSE provides a straightforward numerical error to evaluate the quality.

\begin{table*}[htbp]
\centering
\caption{Quantitative comparisons on three datasets (DocHR14K, RD, SHDocs) in terms of PSNR, SSIM, and RMSE. The best results are in \textbf{bold}, and the second-best are \underline{underlined}.} \vspace{-6pt}
\label{tab:multi_dataset_comparison}
\renewcommand{\arraystretch}{1.2}
\setlength{\tabcolsep}{5pt}
\begin{tabular}{ll|ccc|ccc|ccc}
\toprule
\textbf{Methods} & \textbf{Venue/Year} &
\multicolumn{3}{c|}{\textbf{DocHR14K}} &
\multicolumn{3}{c|}{\textbf{RD}} &
\multicolumn{3}{c}{\textbf{SHIQ}} \\
\cmidrule(lr){3-5} \cmidrule(lr){6-8} \cmidrule(lr){9-11}
& & PSNR~$\uparrow$ & SSIM~$\uparrow$ & RMSE~$\downarrow$ 
  & PSNR~$\uparrow$ & SSIM~$\uparrow$ & RMSE~$\downarrow$
  & PSNR~$\uparrow$ & SSIM~$\uparrow$ & RMSE~$\downarrow$ \\
\midrule
Yang \textit{et al.}~\cite{yang2010real} & ECCV/2010 &
15.505 & 0.688 & 49.670 &
12.094 & 0.493 & 66.660 &
23.168  & 0.772 & 23.147
 \\

Akashi \textit{et al.}~\cite{akashi2014separation} & ACCV/2014 &
18.519 & 0.640 & 32.908 &
15.849 & 0.394 &  42.314 &
21.953 & 0.633 & 22.156
\\

SpecularityNet~\cite{wu2021single} & TMM/2021 &
22.783 & 0.836 & 19.943 &
22.325 & 0.874 & 20.944 &
30.008 & 0.933  & 9.776
 \\

JSHDR~\cite{fu2021multi} & CVPR/2021 &
18.449 & 0.695 & 32.665 &
12.112 & 0.482 & 65.778 &
\textbf{34.131} & 0.860 & - 
 \\

TSHRNet~\cite{fu2023towards} & ICCV/2023 &
21.057 & 0.781 & 24.508 &
19.082 & 0.835 & 29.482 &
32.086  & 0.928 & \underline{7.117}
 \\

DHAN-SHR~\cite{guo2024dual} & ACM MM/2024 &
25.439 & 0.871 & 14.984 &
23.674 & \underline{0.894} & 17.772 &
32.425 & 0.930  & 7.134
 \\

HighlightRemover~\cite{zhang2024highlightremover} & ACM MM/2024 &
\underline{26.650} & 0.918 & \underline{12.754} &
24.377 & 0.884 & 16.622 &
30.231 & 0.947  & 10.226
 \\

DocShadowNet~\cite{li2023high} & ICCV/2023 &
26.600 & \underline{0.919} & 13.015 &
\underline{24.430}  & 0.886 & \underline{16.158} &
31.982 & \underline{0.951} & 7.494 
 \\

\midrule
\rowcolor[HTML]{EFEFEF}
\textbf{L\textsuperscript{2}HRNet} & Ours &
\textbf{27.995} & \textbf{0.928} & \textbf{11.301} &
\textbf{27.558} & \textbf{0.905} & \textbf{11.921} &
\underline{32.710} & \textbf{0.961} & \textbf{6.780}  
\\
\bottomrule
\end{tabular}\vspace{-12pt}
\end{table*}

\subsection{Comparison with State-of-the-Art Methods}

\noindent \textbf{Quantitative Comparison}. As shown in \Cref{tab:multi_dataset_comparison}, we evaluate six highlight removal methods: Yang \textit{et al.} \cite{yang2010real}, Akashi \textit{et al.} \cite{akashi2014separation}, SpecularityNet \cite{wu2021single}, JSHDR \cite{fu2021multi}, TSHRNet \cite{fu2023towards}, and DHAN-SHR \cite{guo2024dual}. Additionally, we assess a document shadow removal method DocShadowNet \cite{li2023high}, which is a laplacian pyramid based image restoration method. \Cref{tab:multi_dataset_comparison} shows the quantitative results on the testing sets over DocHR14K, RD and SHIQ, respectively. The results demonstrate that our method outperforms all the metrics on DocHR14K and RD dataset, which consist of real-world document or text-related images, demonstrating the effectiveness of our approach for document image highlight removal. As for the SHIQ dataset, a widely used benchmark for natural image highlight removal, our method achieves the best performance in SSIM and RMSE, further showing its applicability beyond document images. %, and ranks second in PSNR. 
These results highlight the  superior performance and  strong generalization ability of our method on different dataset.

\noindent \textbf{Qualitative Comparison}. \Cref{fig:Qualitatively Comparison} presents the visual results of four SOTA methods compared with our L\textsuperscript{2}HRNet on DocHR14K.\footnote{Due to space limitations, we present a subset of qualitative results in \Cref{fig:Qualitatively Comparison}. Additional results are provided in \textbf{Section D} of the supplementary material.} Note that the results from JSHDR are obtained using the executable file provided by the authors. %The outcomes show that methods designed for low-resolution inputs (i.e., JSHDR and TSHRNet) lose high-frequency details critical for text clarity, leading to blurred text. 
The outcomes show that methods designed for low-resolution inputs (i.e., JSHDR and TSHRNet) struggle to effectively remove highlights, resulting in residual highlights and a loss of important high-frequency details, possibly because they are designed for low-resolution images.
On the other hand, while DHAN-SHR and DocShadowNet can handle high-resolution images, they lack explicit highlight localization mechanisms, making them suboptimal in terms of both highlight removal and text detail restoration. In contrast, our L\textsuperscript{2}HRNet achieves superior global highlight suppression and sharper textual recovery, which can be attributed to the spatial guidance provided by the residual-based highlight location prior and the refinement capability of the proposed diffusion enhancement module. \footnote{Additional qualitative comparison results of our method against SOTA methods in RD and SHIQ dataset are given in \textbf{Section D} of the supplementary material.}

\vspace{-3pt}
\subsection{Ablation Study and Discussions}

\noindent \textbf{Effectiveness of Different Components in L\textsuperscript{2}HRNet.} %To assess the effectiveness of each component in our proposed L\textsuperscript{2}HRNet, we conduct an ablation study on the DocHR14K dataset. As shown in ~\Cref{tab:method ablation}, we evaluate three key components: the highlight detection network (HDNet), the diffusion-based enhancement module (DEM), and the highlight-aware supervision loss $\mathcal{L}_{\text{mask}}$. By selectively removing specific components, we could discern their individual contributions,affirming that the integration of these elements is vital for document highlight removal.
To assess the effectiveness of each component, we conduct an ablation study on DocHR14K, evaluating three key components: the highlight detection network (HDNet), the diffusion-based enhancement module (DEM), and the highlight-aware supervision loss $\mathcal{L}_{\text{mask}}$.
Removing $\mathcal{L}_{\text{mask}}$ results in a decrease in performance, demonstrating the benefit of highlight location prior. Excluding the DEM also leads to a reduction in performance, likely due to its role in restoring high-frequency details essential for document highlight removal. Similarly, removing both HDNet and \(\mathcal{L}_{\text{mask}}\) results in a performance drop, underscoring the importance of the detection network and the associated supervision in improving highlight removal. Note that removing HDNet also eliminates \(\mathcal{L}_{\text{mask}}\), as \(\mathcal{L}_{\text{mask}}\) depends on the output of HDNet. The best results are achieved when all components are included, highlighting the crucial role of each component. % in achieving optimal highlight removal.

%Starting from a baseline with only HDNet, we observe a PSNR of 27.164, SSIM of 0.923, and RMSE of 12.267. Adding the mask-based supervision $\mathcal{L}_{\text{mask}}$ yields notable improvements across all metrics, demonstrating the benefit of integrating the highlight location prior. Adding the DEM further enhances performance, achieving a PSNR of 27.995, SSIM of 0.928, and RMSE of 11.301, as it effectively refines the high-frequency text details that is crucial for accurate highlight removal. The results show that the combined contribution of HDNet, the location prior, and DEM effectively enhances highlight removal. 
% To further evaluate the importance of HDNet, we also conduct an experiment removing both HDNet and \(\mathcal{L}_{\text{mask}}\). It is important to note that removing HDNet also eliminates the use of \(\mathcal{L}_{\text{mask}}\), as the loss depends on the output of HDNet. Removing these components results in a decrease in performance, underscoring the importance of the detection network and the associated supervision in improving highlight removal.
%We also provide a experiment of diffusion alone, which achieves performance comparable to that of HDNet combined with $\mathcal{L}{\text{mask}}$, indicating the effectiveness of our diffusion-based enhancement module. 
%Finally, enabling all three components leads to the best results, achieving a PSNR of 27.995, SSIM of 0.928, and RMSE of 11.301. These findings confirm that each proposed module contributes positively, and their combination brings complementary benefits for robust highlight removal.

\begin{table}[h]
\vspace{-6pt}
\centering
\caption{Quantitative results of ablation study for L\textsuperscript{2}HRNet on DocHR14K. The best results are in \textbf{bold}.} \vspace{-6pt}
\begin{tabular}{ccc|ccc}
\toprule
\multicolumn{3}{c|}{\textbf{Model Configurations}} & \textit{PSNR}~$\uparrow$ & \textit{SSIM}~$\uparrow$ & \textit{RMSE}~$\downarrow$ \\
\cmidrule(lr){1-3} 
HDNet  & $\mathcal{L}_{\text{mask}}$ & DEM &  &  &  \\
\midrule
\checkmark &  &             & 27.164 & 0.923 & 12.267 \\
\checkmark & \checkmark &   & 27.585 & 0.926 & 11.738 \\
  &   & \checkmark &  27.482 & 0.924 & 11.985 \\
  \checkmark &   & \checkmark &  27.546 & 0.925 & 11.871 \\
\checkmark & \checkmark & \checkmark & \textbf{27.995} & \textbf{0.928} & \textbf{11.301}  \\
\bottomrule
\end{tabular}
\label{tab:method ablation}\vspace{-6pt}
\end{table}

\noindent \textbf{Effectiveness of Different Characteristics of DocHR14K.} 
To evaluate the effectiveness of the different characteristics of our new dataset, three sets of experiments are given, which represent the document categories, color lighting, and images captured by different angles. For each characteristic ablation study, we first create a ``w/o'' training set, then identify the corresponding ``full'' training set with the same remaining variables (e.g., number of images). To make a fair comparison, all the experiments are realized based on DocShadowNet \cite{li2023high}, a SOTA method for document image restoration, with the same testing set. 
As shown in \Cref{tab:dataset ablation}, training the model with additional dataset characteristics consistently improves performance over the baseline, demonstrating the importance of its diverse attributes. The results indicate that models trained on DocHR14K are better suited for handling a broad range of real-world scenarios, underscoring the advantages of DocHR14K for robust document highlight removal under various conditions. 
%-------------------------------------------------------------------------

\vspace{-6pt} % Adjust this value as needed
\begin{table}[h]
\centering
\caption{Quantitative results of ablation study regarding different characteristics of DocHR14K. The best results are in \textbf{bold}.}
\vspace{-8pt}
\begin{tabular}{ccccc}
\toprule
Ablation Study &  PSNR $\uparrow$ & SSIM $\uparrow$ & RMSE $\downarrow$  \\
\midrule
w/o cards  & 24.732 & 0.907 & 16.393 \\
\rowcolor[HTML]{EFEFEF}
full document category& \textbf{26.533} & \textbf{0.919} & \textbf{13.145}  \\
\midrule
w/o color lighting  & 23.703 & 0.895 & 18.619 \\
\rowcolor[HTML]{EFEFEF}
full light category  & \textbf{24.660} & \textbf{0.906} & \textbf{16.341} \\
\midrule
w/o non-vertical shooting  & 20.908 & 0.850 & 25.727 \\
\rowcolor[HTML]{EFEFEF}
full shooting angle & \textbf{22.015} & \textbf{0.870} & \textbf{23.173} \\
% \midrule
% Ours  & \textbf{28.244} & \textbf{0.929} & \textbf{11.055} \\
\bottomrule
\end{tabular}
\label{tab:dataset ablation}\vspace{-8pt}
\end{table}

\noindent \textcolor{black}{\textbf{Discussion on Downstream Application.} }
We further conduct experiments to evaluate the downstream application of DocHR14K and L\textsuperscript{2}HRNet. We first evaluate five mainstream LLMs' text recognition performance on document images disturbed by the highlight from DocHR14K. \footnote{More details about LLMs and text-similarity are given in \textbf{Section B} of the supplementary material.}  We observe that only GPT-4o \cite{achiam2023gpt} could obtain performance close to human recognition under the highlight disturbance, with a text-similarity of 80.9\% provided by o1-preview, while the rest are below 50\%. Next, we evaluate whether de-highlighted document results could enhance the performance of LLMs. \Cref{tab:OCR results from different methods.} reports the outcomes from two SOTA methods JSHDR \cite{fu2021multi}, DocShadowNet \cite{li2023high} and ours. Experiments show that only L\textsuperscript{2}HRNet could improve the recognition performance of GPT-4o, from 80.9\% to 83.5\%, which indicates that L\textsuperscript{2}HRNet not only effectively recovers the global illumination feature but also the local underlying text information and could provide less highlight disturbed document images for downstream tasks.

\vspace{-5pt}
% \vspace{-10pt} % Adjust this value as needed
\begin{table}[h]
\centering
\caption{Comparisons of OCR results from SOTA methods and ours. The best results are in \textbf{bold}.}
\vspace{-5pt}
\begin{tabular}{cccc}
\toprule
Method &  JSHDR  & DocShadowNet & Ours  \\
\midrule
Text-Similarity $\uparrow$ & 18.3\%  & 78.8\% & \textbf{83.5\%}  \\
% \midrule
% Ours  & \textbf{28.244} & \textbf{0.929} & \textbf{11.055} \\
\bottomrule
\end{tabular}
\label{tab:OCR results from different methods.}\vspace{-12pt}
\end{table}

%-------------------------------------------------------------------------
\section{Conclusion}
\label{sec:conclusion}

In this paper, we advance the document image highlight removal task by proposing a large-scale real-world dataset and a location-aware network. Specifically, We present DocHR14K, a high-resolution dataset comprising 14,902 image pairs across six document categories and various lighting conditions. This dataset provides a comprehensive foundation for developing and benchmarking highlight removal methods tailored to document scenarios. Moreover, we propose L\textsuperscript{2}HRNet, a location-aware framework that removes highlights in the low-frequency band with spatial guidance and restores fine-grained textural details in the high-frequency band via the diffusion-based enhancement module. Extensive experiments demonstrate that the diverse characteristics of our DocHR14K contribute to improved highlight removal across various conditions and show the effectiveness of our L\textsuperscript{2}HRNet, which achieves state-of-the-art performance in multiple benchmarks with a large margin.

\bibliographystyle{ACM-Reference-Format}
\bibliography{reference}

%%% -*-BibTeX-*-
%%% Do NOT edit. File created by BibTeX with style
%%% ACM-Reference-Format-Journals [18-Jan-2012].

\begin{thebibliography}{47}

%%% ====================================================================
%%% NOTE TO THE USER: you can override these defaults by providing
%%% customized versions of any of these macros before the \bibliography
%%% command.  Each of them MUST provide its own final punctuation,
%%% except for \shownote{} and \showURL{}.  The latter two
%%% do not use final punctuation, in order to avoid confusing it with
%%% the Web address.
%%%
%%% To suppress output of a particular field, define its macro to expand
%%% to an empty string, or better, \unskip, like this:
%%%
%%% \newcommand{\showURL}[1]{\unskip}   % LaTeX syntax
%%%
%%% \def \showURL #1{\unskip}           % plain TeX syntax
%%%
%%% ====================================================================

\ifx \showCODEN    \undefined \def \showCODEN     #1{\unskip}     \fi
\ifx \showISBNx    \undefined \def \showISBNx     #1{\unskip}     \fi
\ifx \showISBNxiii \undefined \def \showISBNxiii  #1{\unskip}     \fi
\ifx \showISSN     \undefined \def \showISSN      #1{\unskip}     \fi
\ifx \showLCCN     \undefined \def \showLCCN      #1{\unskip}     \fi
\ifx \shownote     \undefined \def \shownote      #1{#1}          \fi
\ifx \showarticletitle \undefined \def \showarticletitle #1{#1}   \fi
\ifx \showURL      \undefined \def \showURL       {\relax}        \fi
% The following commands are used for tagged output and should be
% invisible to TeX
\providecommand\bibfield[2]{#2}
\providecommand\bibinfo[2]{#2}
\providecommand\natexlab[1]{#1}
\providecommand\showeprint[2][]{arXiv:#2}

\bibitem[Achiam et~al\mbox{.}(2023)]%
        {achiam2023gpt}
\bibfield{author}{\bibinfo{person}{Josh Achiam}, \bibinfo{person}{Steven Adler}, \bibinfo{person}{Sandhini Agarwal}, \bibinfo{person}{Lama Ahmad}, \bibinfo{person}{Ilge Akkaya}, \bibinfo{person}{Florencia~Leoni Aleman}, \bibinfo{person}{Diogo Almeida}, \bibinfo{person}{Janko Altenschmidt}, \bibinfo{person}{Sam Altman}, \bibinfo{person}{Shyamal Anadkat}, {et~al\mbox{.}}} \bibinfo{year}{2023}\natexlab{}.
\newblock \showarticletitle{Gpt-4 technical report}.
\newblock \bibinfo{journal}{\emph{arXiv preprint arXiv:2303.08774}} (\bibinfo{year}{2023}).
\newblock


\bibitem[Akashi and Okatani(2015)]%
        {akashi2014separation}
\bibfield{author}{\bibinfo{person}{Yasuhiro Akashi} {and} \bibinfo{person}{Takayuki Okatani}.} \bibinfo{year}{2015}\natexlab{}.
\newblock \showarticletitle{Separation of reflection components by sparse non-negative matrix factorization}. In \bibinfo{booktitle}{\emph{ACCV}}. \bibinfo{pages}{611--625}.
\newblock


\bibitem[Bajcsy et~al\mbox{.}(1996)]%
        {bajcsy1996detection}
\bibfield{author}{\bibinfo{person}{Ruzena Bajcsy}, \bibinfo{person}{Sang~Wook Lee}, {and} \bibinfo{person}{Ale{\v{s}} Leonardis}.} \bibinfo{year}{1996}\natexlab{}.
\newblock \showarticletitle{Detection of diffuse and specular interface reflections and inter-reflections by color image segmentation}.
\newblock \bibinfo{journal}{\emph{IJCV}} \bibinfo{volume}{17}, \bibinfo{number}{3} (\bibinfo{year}{1996}), \bibinfo{pages}{241--272}.
\newblock


\bibitem[Bissacco et~al\mbox{.}(2013)]%
        {bissacco2013photoocr}
\bibfield{author}{\bibinfo{person}{Alessandro Bissacco}, \bibinfo{person}{Mark Cummins}, \bibinfo{person}{Yuval Netzer}, {and} \bibinfo{person}{Hartmut Neven}.} \bibinfo{year}{2013}\natexlab{}.
\newblock \showarticletitle{Photoocr: Reading text in uncontrolled conditions}. In \bibinfo{booktitle}{\emph{ICCV}}. \bibinfo{pages}{785--792}.
\newblock


\bibitem[Burt and Adelson(1987)]%
        {burt1987laplacian}
\bibfield{author}{\bibinfo{person}{Peter~J Burt} {and} \bibinfo{person}{Edward~H Adelson}.} \bibinfo{year}{1987}\natexlab{}.
\newblock \showarticletitle{The Laplacian pyramid as a compact image code}.
\newblock In \bibinfo{booktitle}{\emph{Readings in computer vision}}. \bibinfo{publisher}{Elsevier}, \bibinfo{pages}{671--679}.
\newblock


\bibitem[Chen et~al\mbox{.}(2022)]%
        {chen2022simple}
\bibfield{author}{\bibinfo{person}{Liangyu Chen}, \bibinfo{person}{Xiaojie Chu}, \bibinfo{person}{Xiangyu Zhang}, {and} \bibinfo{person}{Jian Sun}.} \bibinfo{year}{2022}\natexlab{}.
\newblock \showarticletitle{Simple baselines for image restoration}. In \bibinfo{booktitle}{\emph{ECCV}}. \bibinfo{pages}{17--33}.
\newblock


\bibitem[Cicchetti and Comminiello(2024)]%
        {cicchetti2024naf}
\bibfield{author}{\bibinfo{person}{Giordano Cicchetti} {and} \bibinfo{person}{Danilo Comminiello}.} \bibinfo{year}{2024}\natexlab{}.
\newblock \showarticletitle{NAF-DPM: A Nonlinear Activation-Free Diffusion Probabilistic Model for Document Enhancement}.
\newblock \bibinfo{journal}{\emph{arXiv preprint arXiv:2404.05669}} (\bibinfo{year}{2024}).
\newblock


\bibitem[Fu et~al\mbox{.}(2021)]%
        {fu2021multi}
\bibfield{author}{\bibinfo{person}{Gang Fu}, \bibinfo{person}{Qing Zhang}, \bibinfo{person}{Lei Zhu}, \bibinfo{person}{Ping Li}, {and} \bibinfo{person}{Chunxia Xiao}.} \bibinfo{year}{2021}\natexlab{}.
\newblock \showarticletitle{A multi-task network for joint specular highlight detection and removal}. In \bibinfo{booktitle}{\emph{CVPR}}. \bibinfo{pages}{7752--7761}.
\newblock


\bibitem[Fu et~al\mbox{.}(2023)]%
        {fu2023towards}
\bibfield{author}{\bibinfo{person}{Gang Fu}, \bibinfo{person}{Qing Zhang}, \bibinfo{person}{Lei Zhu}, \bibinfo{person}{Chunxia Xiao}, {and} \bibinfo{person}{Ping Li}.} \bibinfo{year}{2023}\natexlab{}.
\newblock \showarticletitle{Towards high-quality specular highlight removal by leveraging large-scale synthetic data}. In \bibinfo{booktitle}{\emph{ICCV}}. \bibinfo{pages}{12857--12865}.
\newblock


\bibitem[Guo et~al\mbox{.}(2018)]%
        {guo2018single}
\bibfield{author}{\bibinfo{person}{Jie Guo}, \bibinfo{person}{Zuojian Zhou}, {and} \bibinfo{person}{Limin Wang}.} \bibinfo{year}{2018}\natexlab{}.
\newblock \showarticletitle{Single image highlight removal with a sparse and low-rank reflection model}. In \bibinfo{booktitle}{\emph{ECCV}}. \bibinfo{pages}{268--283}.
\newblock


\bibitem[Guo et~al\mbox{.}(2014)]%
        {guo2014robust}
\bibfield{author}{\bibinfo{person}{Xiaojie Guo}, \bibinfo{person}{Xiaochun Cao}, {and} \bibinfo{person}{Yi Ma}.} \bibinfo{year}{2014}\natexlab{}.
\newblock \showarticletitle{Robust separation of reflection from multiple images}. In \bibinfo{booktitle}{\emph{CVPR}}. \bibinfo{pages}{2187--2194}.
\newblock


\bibitem[Guo et~al\mbox{.}(2024)]%
        {guo2024dual}
\bibfield{author}{\bibinfo{person}{Xiaojiao Guo}, \bibinfo{person}{Xuhang Chen}, \bibinfo{person}{Shenghong Luo}, \bibinfo{person}{Shuqiang Wang}, {and} \bibinfo{person}{Chi-Man Pun}.} \bibinfo{year}{2024}\natexlab{}.
\newblock \showarticletitle{Dual-Hybrid Attention Network for Specular Highlight Removal}. In \bibinfo{booktitle}{\emph{ACM MM}}. \bibinfo{pages}{10173–10181}.
\newblock


\bibitem[Ho et~al\mbox{.}(2020)]%
        {ho2020denoising}
\bibfield{author}{\bibinfo{person}{Jonathan Ho}, \bibinfo{person}{Ajay Jain}, {and} \bibinfo{person}{Pieter Abbeel}.} \bibinfo{year}{2020}\natexlab{}.
\newblock \showarticletitle{Denoising diffusion probabilistic models}.
\newblock \bibinfo{journal}{\emph{NIPS}} (\bibinfo{year}{2020}), \bibinfo{pages}{6840--6851}.
\newblock


\bibitem[Hou et~al\mbox{.}(2021)]%
        {hou2021text}
\bibfield{author}{\bibinfo{person}{Shiyu Hou}, \bibinfo{person}{Chaoqun Wang}, \bibinfo{person}{Weize Quan}, \bibinfo{person}{Jingen Jiang}, {and} \bibinfo{person}{Dong-Ming Yan}.} \bibinfo{year}{2021}\natexlab{}.
\newblock \showarticletitle{Text-aware single image specular highlight removal}. In \bibinfo{booktitle}{\emph{PRCV}}. \bibinfo{pages}{115--127}.
\newblock


\bibitem[Hu et~al\mbox{.}(2024)]%
        {hu2024highlight}
\bibfield{author}{\bibinfo{person}{Kun Hu}, \bibinfo{person}{Zhaoyangfan Huang}, {and} \bibinfo{person}{Xingjun Wang}.} \bibinfo{year}{2024}\natexlab{}.
\newblock \showarticletitle{Highlight Removal Network Based on an Improved Dichromatic Reflection Model}. In \bibinfo{booktitle}{\emph{ICASSP}}. \bibinfo{pages}{2645--2649}.
\newblock


\bibitem[Kim et~al\mbox{.}(2013)]%
        {kim2013specular}
\bibfield{author}{\bibinfo{person}{Hyeongwoo Kim}, \bibinfo{person}{Hailin Jin}, \bibinfo{person}{Sunil Hadap}, {and} \bibinfo{person}{Inso Kweon}.} \bibinfo{year}{2013}\natexlab{}.
\newblock \showarticletitle{Specular reflection separation using dark channel prior}. In \bibinfo{booktitle}{\emph{CVPR}}. \bibinfo{pages}{1460--1467}.
\newblock


\bibitem[Kligler et~al\mbox{.}(2018)]%
        {kligler2018document}
\bibfield{author}{\bibinfo{person}{Netanel Kligler}, \bibinfo{person}{Sagi Katz}, {and} \bibinfo{person}{Ayellet Tal}.} \bibinfo{year}{2018}\natexlab{}.
\newblock \showarticletitle{Document enhancement using visibility detection}. In \bibinfo{booktitle}{\emph{CVPR}}. \bibinfo{pages}{2374--2382}.
\newblock


\bibitem[Leong et~al\mbox{.}(2024)]%
        {leong2024shdocs}
\bibfield{author}{\bibinfo{person}{Jovin Leong}, \bibinfo{person}{Koa Di}, \bibinfo{person}{Benjamin Cham}, {and} \bibinfo{person}{Shaun Heng}.} \bibinfo{year}{2024}\natexlab{}.
\newblock \showarticletitle{SHDocs: A dataset, benchmark, and method to efficiently generate high-quality, real-world specular highlight data with near-perfect alignment}.
\newblock \bibinfo{journal}{\emph{NIPS}} (\bibinfo{year}{2024}), \bibinfo{pages}{96924--96938}.
\newblock


\bibitem[Li et~al\mbox{.}(2022)]%
        {li2022dit}
\bibfield{author}{\bibinfo{person}{Junlong Li}, \bibinfo{person}{Yiheng Xu}, \bibinfo{person}{Tengchao Lv}, \bibinfo{person}{Lei Cui}, \bibinfo{person}{Cha Zhang}, {and} \bibinfo{person}{Furu Wei}.} \bibinfo{year}{2022}\natexlab{}.
\newblock \showarticletitle{Dit: Self-supervised pre-training for document image transformer}. In \bibinfo{booktitle}{\emph{ACM MM}}. \bibinfo{pages}{3530--3539}.
\newblock


\bibitem[Li et~al\mbox{.}(2020)]%
        {li2020docbank}
\bibfield{author}{\bibinfo{person}{Minghao Li}, \bibinfo{person}{Yiheng Xu}, \bibinfo{person}{Lei Cui}, \bibinfo{person}{Shaohan Huang}, \bibinfo{person}{Furu Wei}, \bibinfo{person}{Zhoujun Li}, {and} \bibinfo{person}{Ming Zhou}.} \bibinfo{year}{2020}\natexlab{}.
\newblock \showarticletitle{DocBank: A benchmark dataset for document layout analysis}.
\newblock \bibinfo{journal}{\emph{arXiv preprint arXiv:2006.01038}} (\bibinfo{year}{2020}).
\newblock


\bibitem[Li et~al\mbox{.}(2023)]%
        {li2023high}
\bibfield{author}{\bibinfo{person}{Zinuo Li}, \bibinfo{person}{Xuhang Chen}, \bibinfo{person}{Chi-Man Pun}, {and} \bibinfo{person}{Xiaodong Cun}.} \bibinfo{year}{2023}\natexlab{}.
\newblock \showarticletitle{High-resolution document shadow removal via a large-scale real-world dataset and a frequency-aware shadow erasing net}. In \bibinfo{booktitle}{\emph{ICCV}}. \bibinfo{pages}{12415--12424}.
\newblock


\bibitem[Liang et~al\mbox{.}(2021)]%
        {liang2021high}
\bibfield{author}{\bibinfo{person}{Jie Liang}, \bibinfo{person}{Hui Zeng}, {and} \bibinfo{person}{Lei Zhang}.} \bibinfo{year}{2021}\natexlab{}.
\newblock \showarticletitle{High-resolution photorealistic image translation in real-time: A laplacian pyramid translation network}. In \bibinfo{booktitle}{\emph{CVPR}}. \bibinfo{pages}{9392--9400}.
\newblock


\bibitem[Lu et~al\mbox{.}(2022)]%
        {lu2022dpm}
\bibfield{author}{\bibinfo{person}{Cheng Lu}, \bibinfo{person}{Yuhao Zhou}, \bibinfo{person}{Fan Bao}, \bibinfo{person}{Jianfei Chen}, \bibinfo{person}{Chongxuan Li}, {and} \bibinfo{person}{Jun Zhu}.} \bibinfo{year}{2022}\natexlab{}.
\newblock \showarticletitle{Dpm-solver: A fast ode solver for diffusion probabilistic model sampling in around 10 steps}.
\newblock \bibinfo{journal}{\emph{NIPS}} (\bibinfo{year}{2022}), \bibinfo{pages}{5775--5787}.
\newblock


\bibitem[Otsu et~al\mbox{.}(1975)]%
        {otsu1975threshold}
\bibfield{author}{\bibinfo{person}{Nobuyuki Otsu} {et~al\mbox{.}}} \bibinfo{year}{1975}\natexlab{}.
\newblock \showarticletitle{A threshold selection method from gray-level histograms}.
\newblock \bibinfo{journal}{\emph{Automatica}} \bibinfo{volume}{11}, \bibinfo{number}{285-296} (\bibinfo{year}{1975}), \bibinfo{pages}{23--27}.
\newblock


\bibitem[Qiao et~al\mbox{.}(2022)]%
        {qiao2022davarocr}
\bibfield{author}{\bibinfo{person}{Liang Qiao}, \bibinfo{person}{Hui Jiang}, \bibinfo{person}{Ying Chen}, \bibinfo{person}{Can Li}, \bibinfo{person}{Pengfei Li}, \bibinfo{person}{Zaisheng Li}, \bibinfo{person}{Baorui Zou}, \bibinfo{person}{Dashan Guo}, \bibinfo{person}{Yingda Xu}, \bibinfo{person}{Yunlu Xu}, \bibinfo{person}{Zhanzhan Cheng}, {and} \bibinfo{person}{Yi Niu}.} \bibinfo{year}{2022}\natexlab{}.
\newblock \showarticletitle{DavarOCR: A Toolbox for OCR and Multi-Modal Document Understanding}. In \bibinfo{booktitle}{\emph{ACM MM}}. \bibinfo{pages}{7355--7358}.
\newblock


\bibitem[Ren et~al\mbox{.}(2016)]%
        {ren2016faster}
\bibfield{author}{\bibinfo{person}{Shaoqing Ren}, \bibinfo{person}{Kaiming He}, \bibinfo{person}{Ross Girshick}, {and} \bibinfo{person}{Jian Sun}.} \bibinfo{year}{2016}\natexlab{}.
\newblock \showarticletitle{Faster R-CNN: Towards real-time object detection with region proposal networks}.
\newblock \bibinfo{journal}{\emph{TPAMI}} \bibinfo{volume}{39}, \bibinfo{number}{6} (\bibinfo{year}{2016}), \bibinfo{pages}{1137--1149}.
\newblock


\bibitem[Rudin et~al\mbox{.}(1992)]%
        {rudin1992nonlinear}
\bibfield{author}{\bibinfo{person}{Leonid~I Rudin}, \bibinfo{person}{Stanley Osher}, {and} \bibinfo{person}{Emad Fatemi}.} \bibinfo{year}{1992}\natexlab{}.
\newblock \showarticletitle{Nonlinear total variation based noise removal algorithms}.
\newblock \bibinfo{journal}{\emph{Physica D: nonlinear phenomena}} \bibinfo{volume}{60}, \bibinfo{number}{1-4} (\bibinfo{year}{1992}), \bibinfo{pages}{259--268}.
\newblock


\bibitem[Saharia et~al\mbox{.}(2022)]%
        {saharia2022image}
\bibfield{author}{\bibinfo{person}{Chitwan Saharia}, \bibinfo{person}{Jonathan Ho}, \bibinfo{person}{William Chan}, \bibinfo{person}{Tim Salimans}, \bibinfo{person}{David~J Fleet}, {and} \bibinfo{person}{Mohammad Norouzi}.} \bibinfo{year}{2022}\natexlab{}.
\newblock \showarticletitle{Image super-resolution via iterative refinement}.
\newblock \bibinfo{journal}{\emph{TPAMI}} \bibinfo{volume}{45}, \bibinfo{number}{4} (\bibinfo{year}{2022}), \bibinfo{pages}{4713--4726}.
\newblock


\bibitem[Shafer(1985)]%
        {shafer1985using}
\bibfield{author}{\bibinfo{person}{Steven~A Shafer}.} \bibinfo{year}{1985}\natexlab{}.
\newblock \showarticletitle{Using color to separate reflection components}.
\newblock \bibinfo{journal}{\emph{Color Research \& Application}} \bibinfo{volume}{10}, \bibinfo{number}{4} (\bibinfo{year}{1985}), \bibinfo{pages}{210--218}.
\newblock


\bibitem[Shi et~al\mbox{.}(2016)]%
        {shi2016end}
\bibfield{author}{\bibinfo{person}{Baoguang Shi}, \bibinfo{person}{Xiang Bai}, {and} \bibinfo{person}{Cong Yao}.} \bibinfo{year}{2016}\natexlab{}.
\newblock \showarticletitle{An end-to-end trainable neural network for image-based sequence recognition and its application to scene text recognition}.
\newblock \bibinfo{journal}{\emph{TPAMI}} \bibinfo{volume}{39}, \bibinfo{number}{11} (\bibinfo{year}{2016}), \bibinfo{pages}{2298--2304}.
\newblock


\bibitem[Simonyan and Zisserman(2014)]%
        {simonyan2014very}
\bibfield{author}{\bibinfo{person}{Karen Simonyan} {and} \bibinfo{person}{Andrew Zisserman}.} \bibinfo{year}{2014}\natexlab{}.
\newblock \showarticletitle{Very deep convolutional networks for large-scale image recognition}.
\newblock \bibinfo{journal}{\emph{arXiv preprint arXiv:1409.1556}} (\bibinfo{year}{2014}).
\newblock


\bibitem[Sohl-Dickstein et~al\mbox{.}(2015)]%
        {sohl2015deep}
\bibfield{author}{\bibinfo{person}{Jascha Sohl-Dickstein}, \bibinfo{person}{Eric Weiss}, \bibinfo{person}{Niru Maheswaranathan}, {and} \bibinfo{person}{Surya Ganguli}.} \bibinfo{year}{2015}\natexlab{}.
\newblock \showarticletitle{Deep unsupervised learning using nonequilibrium thermodynamics}. In \bibinfo{booktitle}{\emph{ICML}}. \bibinfo{pages}{2256--2265}.
\newblock


\bibitem[Song et~al\mbox{.}(2020)]%
        {song2020denoising}
\bibfield{author}{\bibinfo{person}{Jiaming Song}, \bibinfo{person}{Chenlin Meng}, {and} \bibinfo{person}{Stefano Ermon}.} \bibinfo{year}{2020}\natexlab{}.
\newblock \showarticletitle{Denoising diffusion implicit models}.
\newblock \bibinfo{journal}{\emph{arXiv preprint arXiv:2010.02502}} (\bibinfo{year}{2020}).
\newblock


\bibitem[Tan and Ikeuchi(2005)]%
        {tan2005separating}
\bibfield{author}{\bibinfo{person}{Robby~T Tan} {and} \bibinfo{person}{Katsushi Ikeuchi}.} \bibinfo{year}{2005}\natexlab{}.
\newblock \showarticletitle{Separating reflection components of textured surfaces using a single image}.
\newblock \bibinfo{journal}{\emph{TPAMI}} \bibinfo{volume}{27}, \bibinfo{number}{2} (\bibinfo{year}{2005}), \bibinfo{pages}{178--193}.
\newblock


\bibitem[Wang et~al\mbox{.}(2004)]%
        {wang2004image}
\bibfield{author}{\bibinfo{person}{Zhou Wang}, \bibinfo{person}{Alan~C Bovik}, \bibinfo{person}{Hamid~R Sheikh}, {and} \bibinfo{person}{Eero~P Simoncelli}.} \bibinfo{year}{2004}\natexlab{}.
\newblock \showarticletitle{Image quality assessment: from error visibility to structural similarity}.
\newblock \bibinfo{journal}{\emph{TIP}} \bibinfo{volume}{13}, \bibinfo{number}{4} (\bibinfo{year}{2004}), \bibinfo{pages}{600--612}.
\newblock


\bibitem[Wen et~al\mbox{.}(2021)]%
        {wen2021polarization}
\bibfield{author}{\bibinfo{person}{Sijia Wen}, \bibinfo{person}{Yinqiang Zheng}, {and} \bibinfo{person}{Feng Lu}.} \bibinfo{year}{2021}\natexlab{}.
\newblock \showarticletitle{Polarization guided specular reflection separation}.
\newblock \bibinfo{journal}{\emph{TIP}}  \bibinfo{volume}{30} (\bibinfo{year}{2021}), \bibinfo{pages}{7280--7291}.
\newblock


\bibitem[Wolff and Boult(1991)]%
        {wolff1991constraining}
\bibfield{author}{\bibinfo{person}{Lawrence~B. Wolff} {and} \bibinfo{person}{Terrance~E. Boult}.} \bibinfo{year}{1991}\natexlab{}.
\newblock \showarticletitle{Constraining object features using a polarization reflectance model}.
\newblock \bibinfo{journal}{\emph{TPAMI}} \bibinfo{volume}{13}, \bibinfo{number}{07} (\bibinfo{year}{1991}), \bibinfo{pages}{635--657}.
\newblock


\bibitem[Wu et~al\mbox{.}(2023)]%
        {wu2023joint}
\bibfield{author}{\bibinfo{person}{Zhongqi Wu}, \bibinfo{person}{Jianwei Guo}, \bibinfo{person}{Chuanqing Zhuang}, \bibinfo{person}{Jun Xiao}, \bibinfo{person}{Dong-Ming Yan}, {and} \bibinfo{person}{Xiaopeng Zhang}.} \bibinfo{year}{2023}\natexlab{}.
\newblock \showarticletitle{Joint specular highlight detection and removal in single images via Unet-Transformer}.
\newblock \bibinfo{journal}{\emph{CVM}} \bibinfo{volume}{9}, \bibinfo{number}{1} (\bibinfo{year}{2023}), \bibinfo{pages}{141--154}.
\newblock


\bibitem[Wu et~al\mbox{.}(2021)]%
        {wu2021single}
\bibfield{author}{\bibinfo{person}{Zhongqi Wu}, \bibinfo{person}{Chuanqing Zhuang}, \bibinfo{person}{Jian Shi}, \bibinfo{person}{Jianwei Guo}, \bibinfo{person}{Jun Xiao}, \bibinfo{person}{Xiaopeng Zhang}, {and} \bibinfo{person}{Dong-Ming Yan}.} \bibinfo{year}{2021}\natexlab{}.
\newblock \showarticletitle{Single-image specular highlight removal via real-world dataset construction}.
\newblock \bibinfo{journal}{\emph{TMM}}  \bibinfo{volume}{24} (\bibinfo{year}{2021}), \bibinfo{pages}{3782--3793}.
\newblock


\bibitem[Xu et~al\mbox{.}(2020)]%
        {xu2020layoutlm}
\bibfield{author}{\bibinfo{person}{Yiheng Xu}, \bibinfo{person}{Minghao Li}, \bibinfo{person}{Lei Cui}, \bibinfo{person}{Shaohan Huang}, \bibinfo{person}{Furu Wei}, {and} \bibinfo{person}{Ming Zhou}.} \bibinfo{year}{2020}\natexlab{}.
\newblock \showarticletitle{Layoutlm: Pre-training of text and layout for document image understanding}. In \bibinfo{booktitle}{\emph{ACM SIGKDD}}. \bibinfo{pages}{1192--1200}.
\newblock


\bibitem[Yamamoto and Nakazawa(2019)]%
        {yamamoto2019general}
\bibfield{author}{\bibinfo{person}{Takahisa Yamamoto} {and} \bibinfo{person}{Atsushi Nakazawa}.} \bibinfo{year}{2019}\natexlab{}.
\newblock \showarticletitle{General improvement method of specular component separation using high-emphasis filter and similarity function}.
\newblock \bibinfo{journal}{\emph{ITE Trans. Media Technol. Appl.}} \bibinfo{volume}{7}, \bibinfo{number}{2} (\bibinfo{year}{2019}), \bibinfo{pages}{92--102}.
\newblock


\bibitem[Yang et~al\mbox{.}(2014)]%
        {yang2014efficient}
\bibfield{author}{\bibinfo{person}{Qingxiong Yang}, \bibinfo{person}{Jinhui Tang}, {and} \bibinfo{person}{Narendra Ahuja}.} \bibinfo{year}{2014}\natexlab{}.
\newblock \showarticletitle{Efficient and robust specular highlight removal}.
\newblock \bibinfo{journal}{\emph{TPAMI}} \bibinfo{volume}{37}, \bibinfo{number}{6} (\bibinfo{year}{2014}), \bibinfo{pages}{1304--1311}.
\newblock


\bibitem[Yang et~al\mbox{.}(2010)]%
        {yang2010real}
\bibfield{author}{\bibinfo{person}{Qingxiong Yang}, \bibinfo{person}{Shengnan Wang}, {and} \bibinfo{person}{Narendra Ahuja}.} \bibinfo{year}{2010}\natexlab{}.
\newblock \showarticletitle{Real-time specular highlight removal using bilateral filtering}. In \bibinfo{booktitle}{\emph{ECCV}}. \bibinfo{pages}{87--100}.
\newblock


\bibitem[Yang et~al\mbox{.}(2023)]%
        {yang2023docdiff}
\bibfield{author}{\bibinfo{person}{Zongyuan Yang}, \bibinfo{person}{Baolin Liu}, \bibinfo{person}{Yongping Xxiong}, \bibinfo{person}{Lan Yi}, \bibinfo{person}{Guibin Wu}, \bibinfo{person}{Xiaojun Tang}, \bibinfo{person}{Ziqi Liu}, \bibinfo{person}{Junjie Zhou}, {and} \bibinfo{person}{Xing Zhang}.} \bibinfo{year}{2023}\natexlab{}.
\newblock \showarticletitle{Docdiff: Document enhancement via residual diffusion models}. In \bibinfo{booktitle}{\emph{ACM MM}}. \bibinfo{pages}{2795--2806}.
\newblock


\bibitem[Yoon et~al\mbox{.}(2006)]%
        {yoon2006fast}
\bibfield{author}{\bibinfo{person}{Kuk-Jin Yoon}, \bibinfo{person}{Yoojin Choi}, {and} \bibinfo{person}{In~So Kweon}.} \bibinfo{year}{2006}\natexlab{}.
\newblock \showarticletitle{Fast separation of reflection components using a specularity-invariant image representation}. In \bibinfo{booktitle}{\emph{ICIP}}. \bibinfo{pages}{973--976}.
\newblock


\bibitem[Zhang et~al\mbox{.}(2024)]%
        {zhang2024highlightremover}
\bibfield{author}{\bibinfo{person}{Ling Zhang}, \bibinfo{person}{Yidong Ma}, \bibinfo{person}{Zhi Jiang}, \bibinfo{person}{Weilei He}, \bibinfo{person}{Zhongyun Bao}, \bibinfo{person}{Gang Fu}, \bibinfo{person}{Wenju Xu}, {and} \bibinfo{person}{Chunxia Xiao}.} \bibinfo{year}{2024}\natexlab{}.
\newblock \showarticletitle{HighlightRemover: Spatially Valid Pixel Learning for Image Specular Highlight Removal}. In \bibinfo{booktitle}{\emph{ACM MM}}. \bibinfo{pages}{10046--10054}.
\newblock


\bibitem[Zhong et~al\mbox{.}(2019)]%
        {zhong2019publaynet}
\bibfield{author}{\bibinfo{person}{Xu Zhong}, \bibinfo{person}{Jianbin Tang}, {and} \bibinfo{person}{Antonio~Jimeno Yepes}.} \bibinfo{year}{2019}\natexlab{}.
\newblock \showarticletitle{Publaynet: largest dataset ever for document layout analysis}. In \bibinfo{booktitle}{\emph{ICDAR}}. \bibinfo{pages}{1015--1022}.
\newblock


\end{thebibliography}

%%
%% The next two lines define the bibliography style to be used, and
%% the bibliography file.

%%
%% If your work has an appendix, this is the place to put it.
% \appendix

%% If your work has an appendix, this is the place to put it.
% \appendix
\clearpage
\setcounter{section}{0} % 重置章节计数

In this supplementary material, \textbf{\Cref{sec: quality control}} first provides the details and results of the quality control for DocHR14K. Next, \textbf{\Cref{sec: evaluation of LLMs}}  details the assessment of LLMs' recognition performance under highlight disturbances. Then, \textbf{\Cref{sec: more samples from DocHR14K}} presents additional details for DocHR14K. Furthermore, \textbf{\Cref{sec: more qualitative comparisons}} provides more qualitative comparisons of three benchmark datasets to support the main paper. Finally, \textbf{\Cref{sec: Limitations and Future Work}} discusses the limitations of our approach and outlines directions for future work.

\section{Quality Control}
\label{sec: quality control}
The details of the quality control process are outlined below. To comprehensively evaluate our dataset, we employ stratified sampling based on document category, lighting type, language, shooting angle, and environment. We select 10\% of the image pairs from DocHR14K, resulting in 1,490 samples. These samples are then randomly divided into three groups and evaluated by three well-trained personnel by answering the following four questions:
\begin{itemize}
    \item \(\mathbf{Q_1}\): Is the highlighted image captured clearly without any blur?
    \item \(\mathbf{Q_2}\): Has the corresponding highlight been completely removed in the ground truth image?
    \item \(\mathbf{Q_3}\): Is the text in the non-highlighted document image clear and legible?
    \item \(\mathbf{Q_4}\): Are there any new artifacts or distortions in the non-highlighted document image?
\end{itemize}

%-------------------------------------------------------------------------
% \vspace{-10pt} % Adjust this value as needed
\begin{table}[!h]
\centering
\caption{Results of quality control evaluation. Ratings for $Q_1$, $Q_2$, and $Q_3$ represent the percentage of “Yes” responses, while $Q_4$ represents the percentage of “No” responses.}
\begin{tabular}{ccccc}
\toprule
Question  & $Q_{1}$ & $Q_{2}$ & $Q_{3}$  & $Q_{4}$ \\
\midrule
Rating   & 100\% & 98.3\% & 99.5\% & 98.7\% \\
\bottomrule
\end{tabular}
\label{tab: Results of quality control}
\end{table}
%-------------------------------------------------------------------------

As reported in \Cref{tab: Results of quality control}, $Q_{1}$ indicates that all the evaluated images are well-captured without any blurring. Additionally, $Q_2$ demonstrates that the majority of diffuse images are completely free from highlight residuals. Furthermore, $Q_{3}$, shows that almost all text information in the diffuse images is clear and legible, which is suitable to be the reference. Finally, $Q_{4}$ presents that only minimal artifacts or distortions, such as illumination or color inconsistencies, are introduced in the diffuse images. In summary, \Cref{tab: Results of quality control} indicates that DocHR14K achieves high standards in image clarity, highlight removal, text preservation, and artifact minimization, which also proves the effectiveness of our highlight-diffuse image-collecting procedure.

%-------------------------------------------------------------------------
\section{Evaluation of LLMs}

% %-------------------------------------------------------------------------
 
Recently, large language models (LLMs), particularly vision language models, have demonstrated strong proficiency in processing both image and text information. Hence, we conduct experiments to evaluate the performance of LLMs under the disturbance of highlight. 

\begin{figure*}[h]
    \centering
    \vspace{-12pt}
% \vspace{-0.4cm} % Space between rows
    % 1-1
    \begin{subfigure}[b]{1\textwidth}
        \centering
        \includegraphics[width=\linewidth]{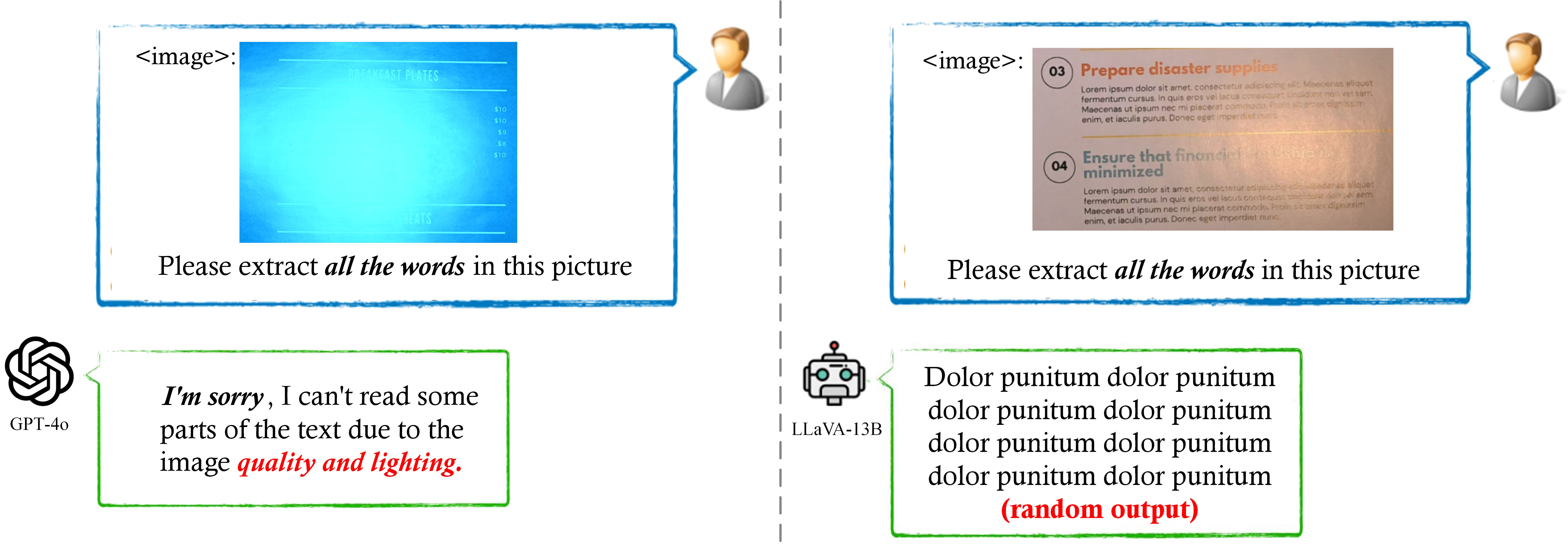}
        % \caption{Illustration of GPT-4o.}
        \label{fig: Illustration of GPT-4o}
    \end{subfigure}

    \caption{Illustration of LLMs' evaluation. GPT-4o could report abnormal conditions with responses like ``I'm sorry, I can't read some parts of the text due to the image quality and lighting.'' In contrast, other LLMs tend to provide unrelated or inaccurate answers when encountering highlight disturbances. }\vspace{-12pt}
    \label{fig: Illustration of LLMs}
\end{figure*}

In detail, we select 50 samples from DocHR14K and send them to the LLMs with the prompt to let them give the recognition outcome. We note that some of the LLMs will give a combination of the recognition answer and meaningless result on the document image degraded by highlight. To make a fair comparison, we use the SOTA model GPT ``o1-preview" to assess the text similarity between the output result and the ground truth text information recognized by humans. Five LLMs are evaluated: LLaVA-7B, LLaVA-13B, mPLUG-Owl, Qwen-VL, and GPT-4o . As exhibited in \Cref{tab: Text-Similarity of the results from different LLM}, GPT-4o achieves similar results compared with humans, while other LLMs are less than 50\%. Exhibited in \Cref{fig: Illustration of LLMs}, We also observe that GPT-4o prefers to overlook the degraded regions and gives answers like ``I'm sorry, I can't read some parts of the text due to the image quality and lighting.'' In contrast, other LLMs tend to give unrelated answers when the highlight disturbs severely. The experiment above indicates that GPT-4o could achieve better recognition performance under the highlight disturbance during text recognition. 
\begin{table*}[!htbp]
\vspace{-12pt}
\centering
\caption{Results of text-similarity across different LLMs. The best results are in \textbf{bold}, and the second-best results are \underline{underlined}.}
\begin{tabular}{cccccc}
\toprule
Models  & LLaVA-7B & LLaVA-13B & mPLUG-Owl 2 & Qwen-VL &  GPT-4o\\
\midrule
Text-Similarity $\uparrow$  & 0.295 & 0.352 & 0.209 & \underline{0.446} & \textbf{0.809}\\
% \midrule
% Ours  & \textbf{28.244} & \textbf{0.929} & \textbf{11.055} \\
\bottomrule
\end{tabular}
\label{tab: Text-Similarity of the results from different LLM}
\end{table*}
\label{sec: evaluation of LLMs}
% \vspace{-10pt} % Adjust this value as needed

\vspace{-6pt}
%-------------------------------------------------------------------------
\section{More Details of DocHR14K}
\label{sec: more samples from DocHR14K}
\begin{figure}[h]
  \centering

  % ----- 第一行 -----
  \begin{subfigure}[b]{0.48\linewidth}
    \includegraphics[width=\linewidth]{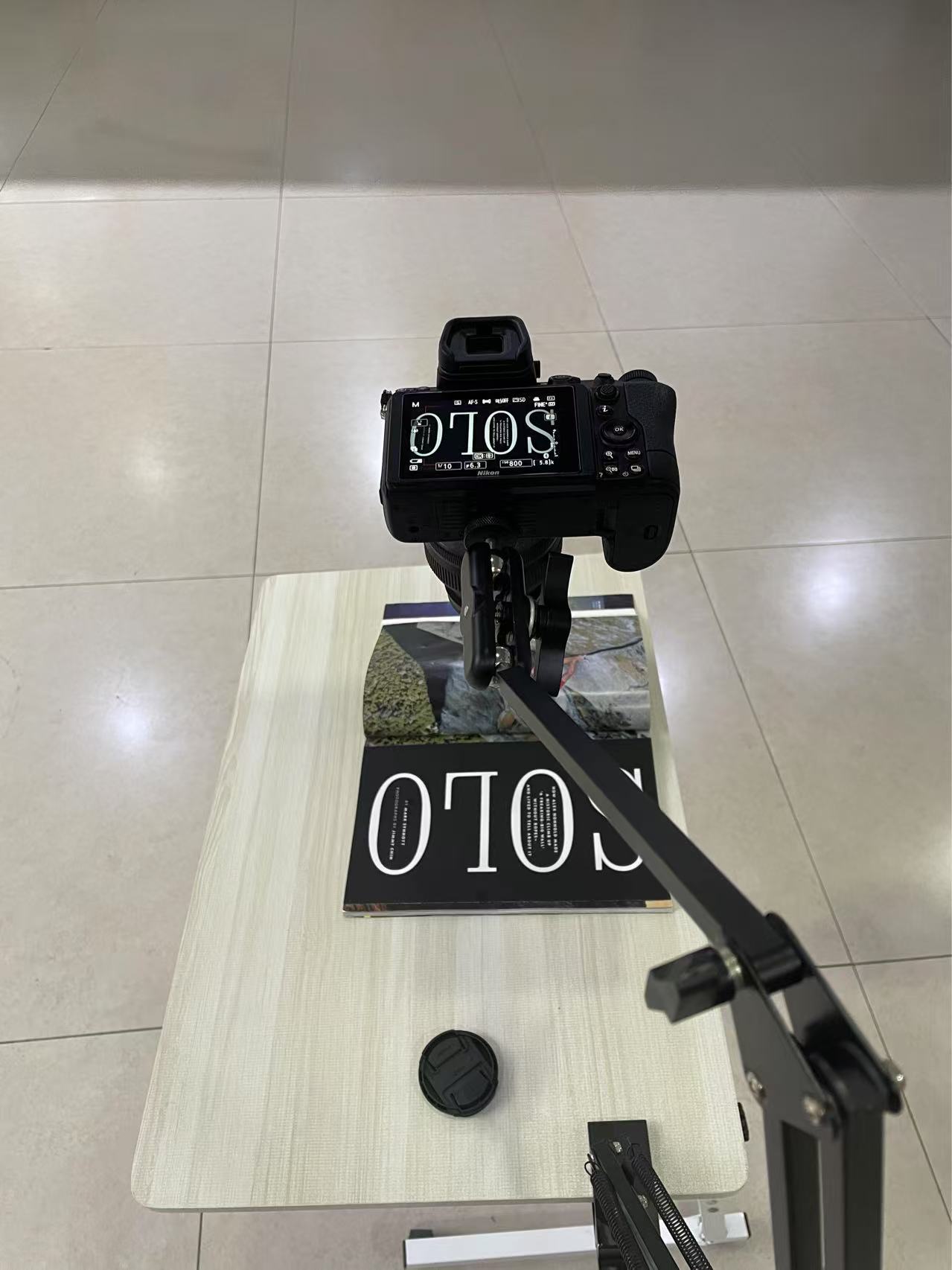}
  \end{subfigure}
  \begin{subfigure}[b]{0.48\linewidth}
    \includegraphics[width=\linewidth]{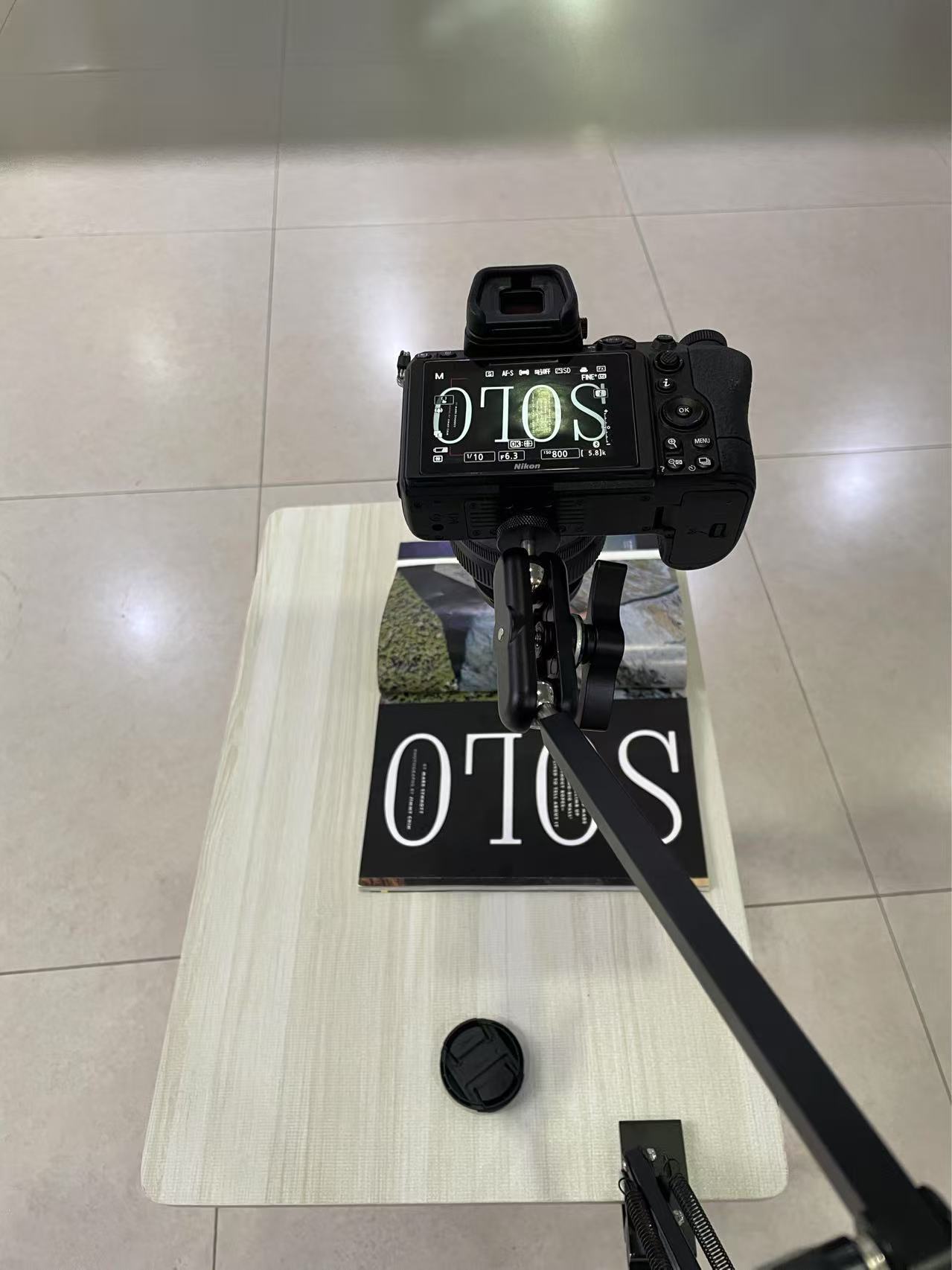}
  \end{subfigure}
  % ----- 第二行 -----
  \begin{subfigure}[b]{0.48\linewidth}
    \includegraphics[angle=0,width=\linewidth]{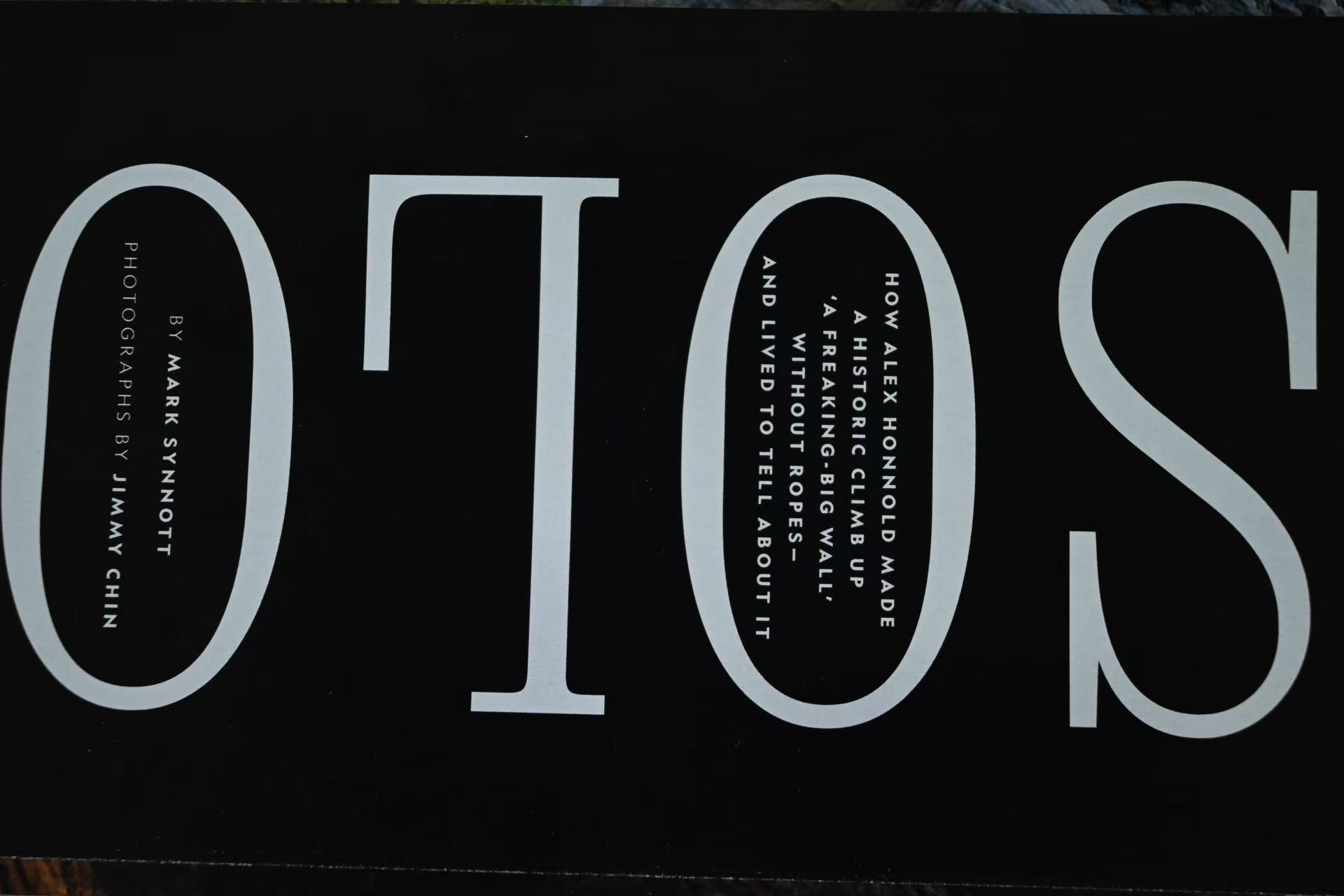}
    \caption{}
  \end{subfigure}
  \begin{subfigure}[b]{0.48\linewidth}
    \includegraphics[angle=0,width=\linewidth]{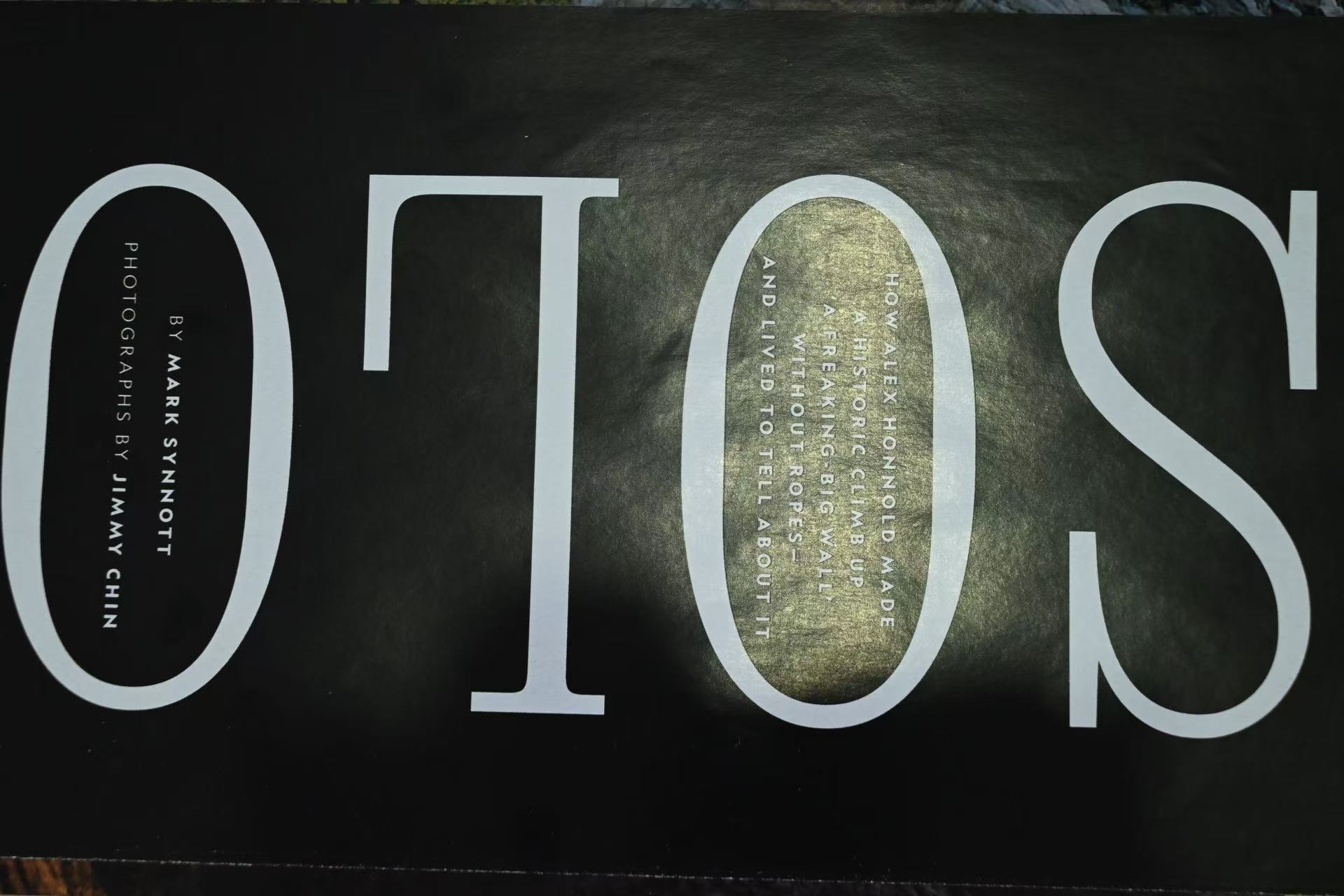}
    \caption{}
  \end{subfigure}

  \caption{Illustration of our refined cross-polarization data acquisition strategy under real-world conditions. Subfigures (a)  depict the setup and resulting image for capturing highlight-free document images. Subfigures (b) show the corresponding configuration and result for collecting images with highlights.}
  \label{fig:refined_cp_vis}
\end{figure}

 For training our L\textsuperscript{2}HRNet, DocHR14K is randomly divided into 11,922 pairs for training, 1,490 pairs for validation, and 1,490 pairs for testing. To enhance the model's applicability in real-world scenarios, we incorporate digital card images sourced from the open-access platform Wikimedia Commons, including bank cards, ID cards, and driver's licenses. These digital images were printed onto PVC materials commonly used for ID cards to simulate real-world usage conditions. To support the main paper, we also provide the real-world illustration of our refined cross-polarization, which is shown in \Cref{fig:refined_cp_vis}. Our dataset and source code will be released after the paper is accepted.

% %-------------------------------------------------------------------------

\vspace{-6pt}
\section{More Qualitative Comparisons}
\label{sec: more qualitative comparisons}

\begin{figure*}[tp]  
    % 1st
    \begin{subfigure}[b]{0.138\textwidth}
        \centering
        \includegraphics[angle=-90, width=\linewidth]{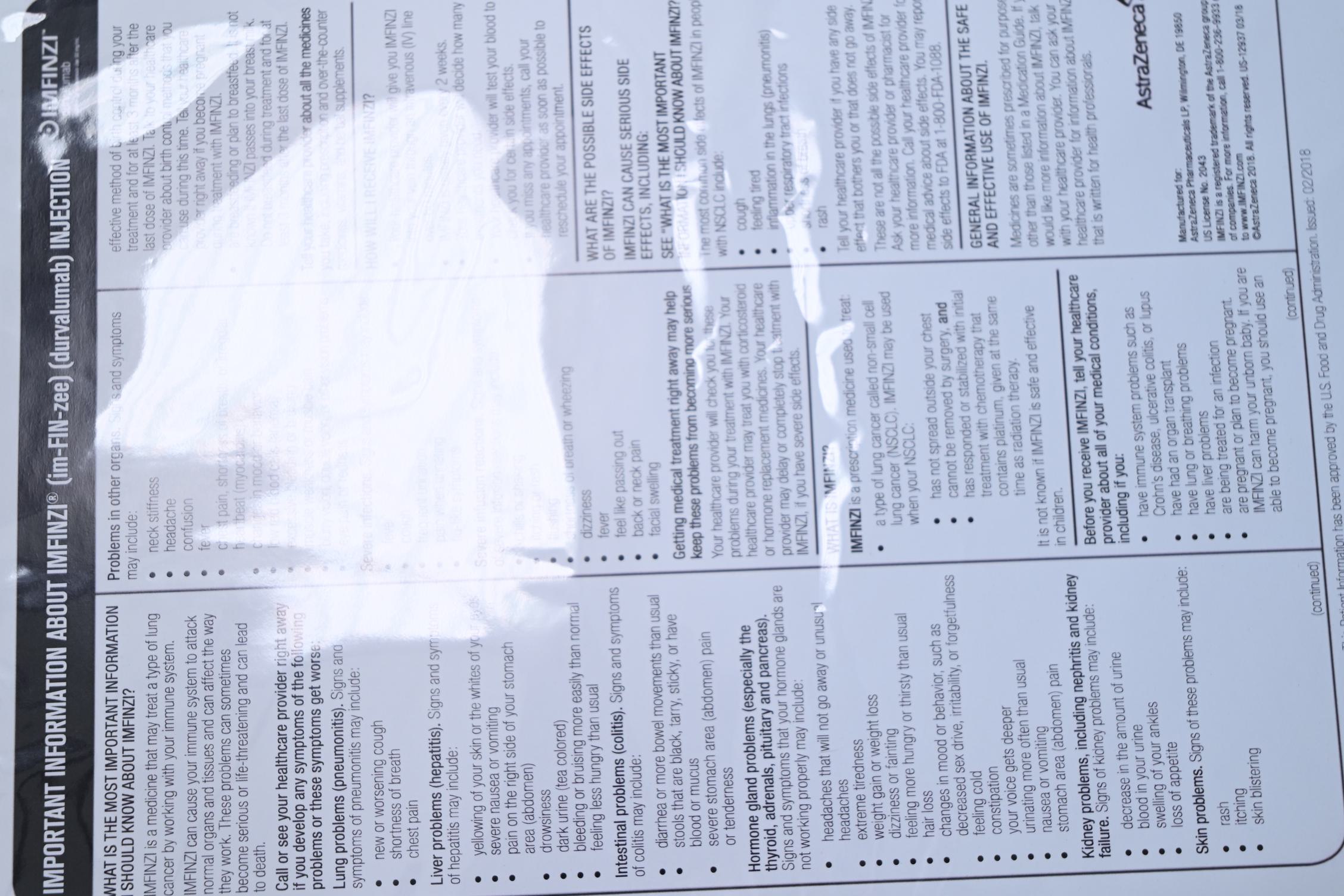} % 替换为实际文件路径
    \end{subfigure}
    \begin{subfigure}[b]{0.138\textwidth}
        \centering
        \includegraphics[angle=-90, width=\linewidth]{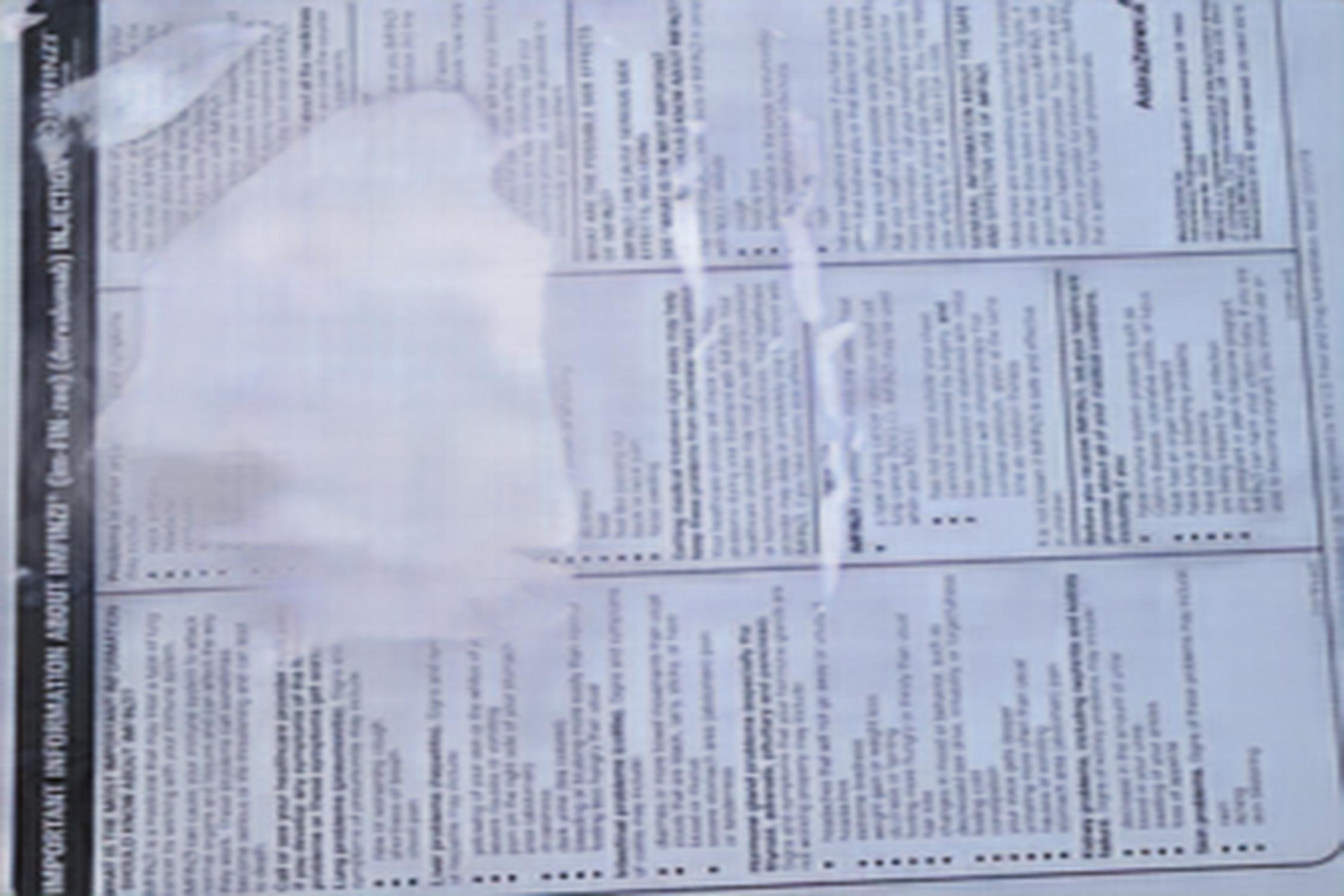} % 替换为实际文件路径
    \end{subfigure}
    \begin{subfigure}[b]{0.138\textwidth}
        \centering
        \includegraphics[angle=-90, width=\linewidth]{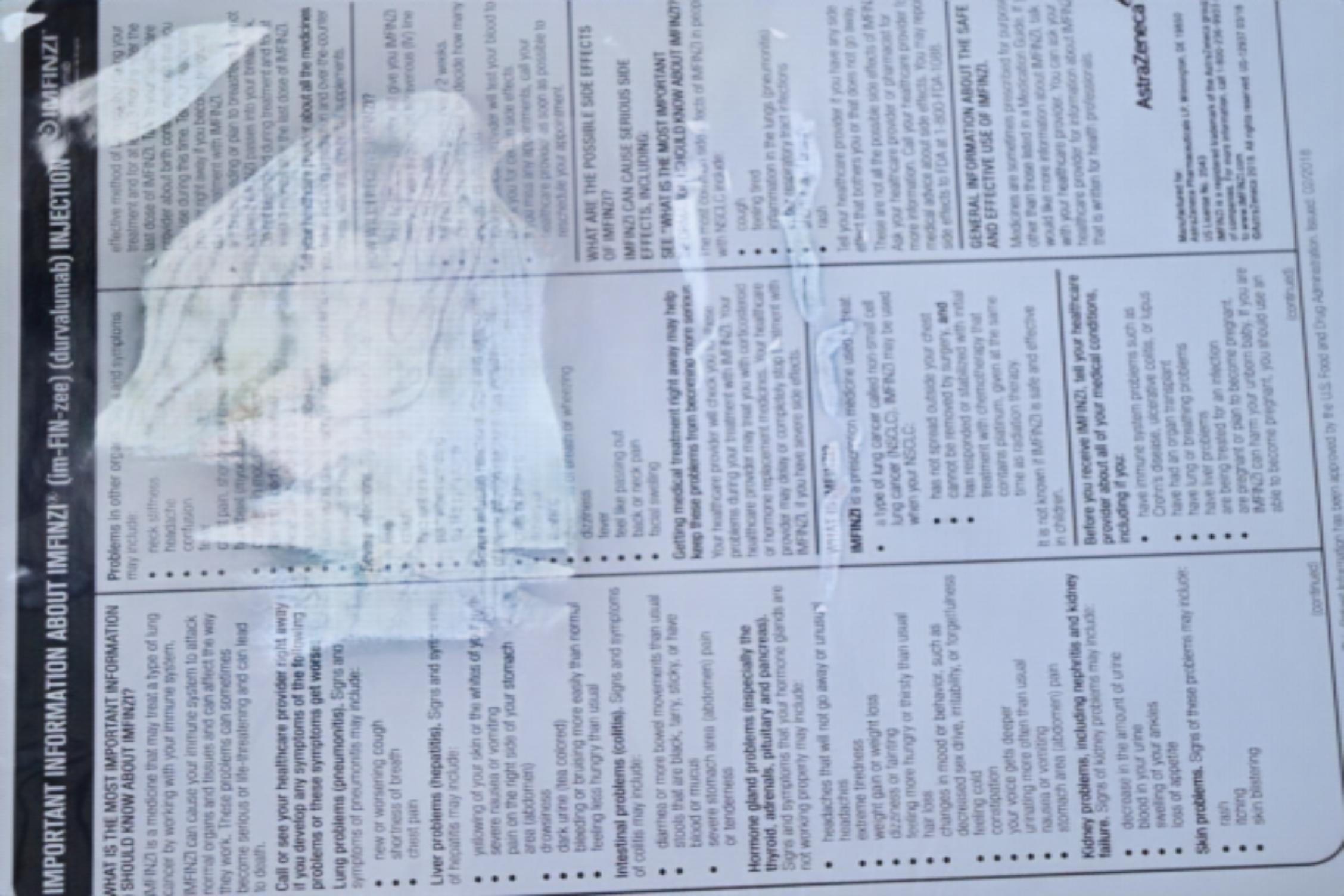} % 替换为实际文件路径
    \end{subfigure}
    \begin{subfigure}[b]{0.138\textwidth}
        \centering
        \includegraphics[angle=-90, width=\linewidth]{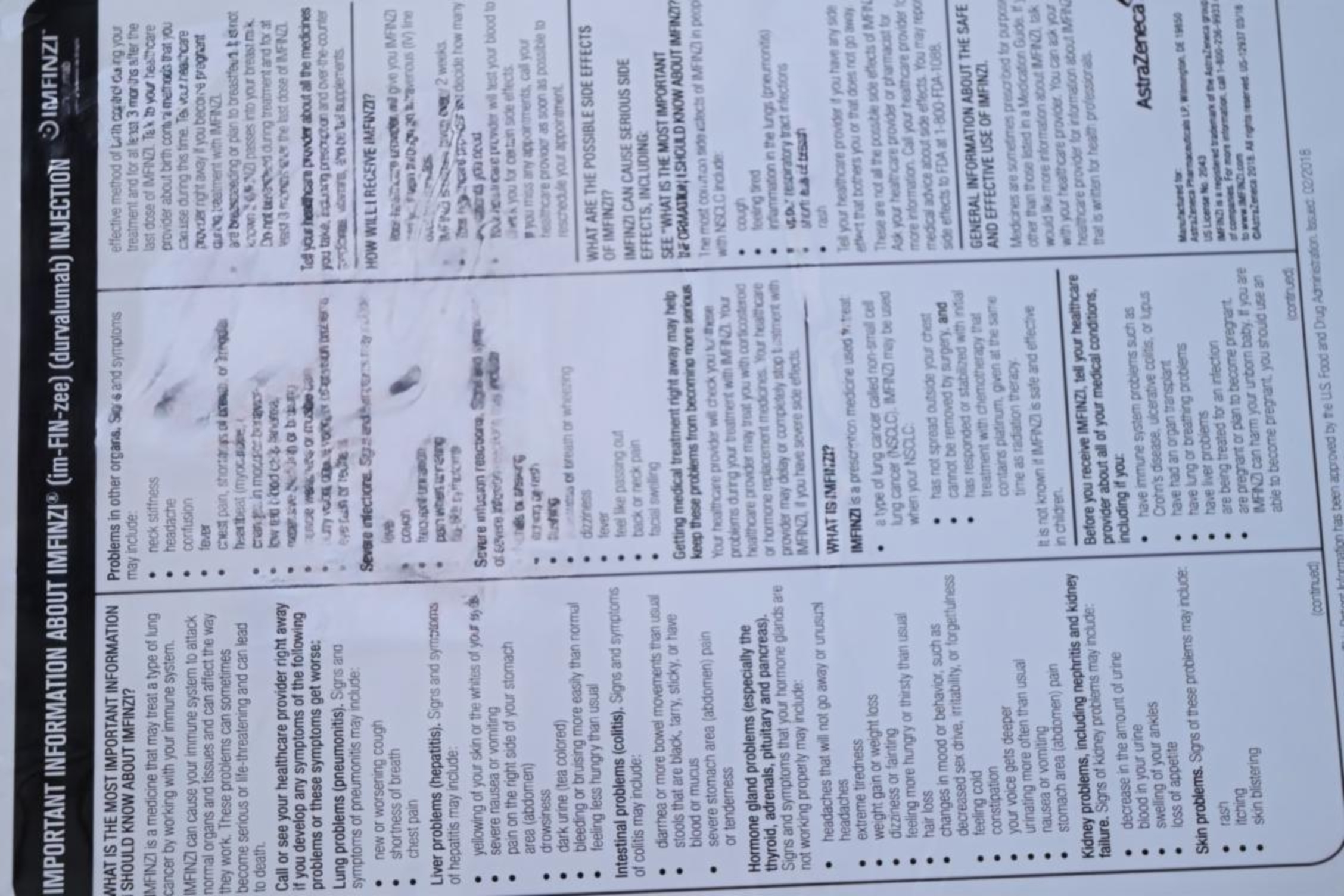} % 替换为实际文件路径
    \end{subfigure}
    \begin{subfigure}[b]{0.138\textwidth}
        \centering
        \includegraphics[angle=-90, width=\linewidth]{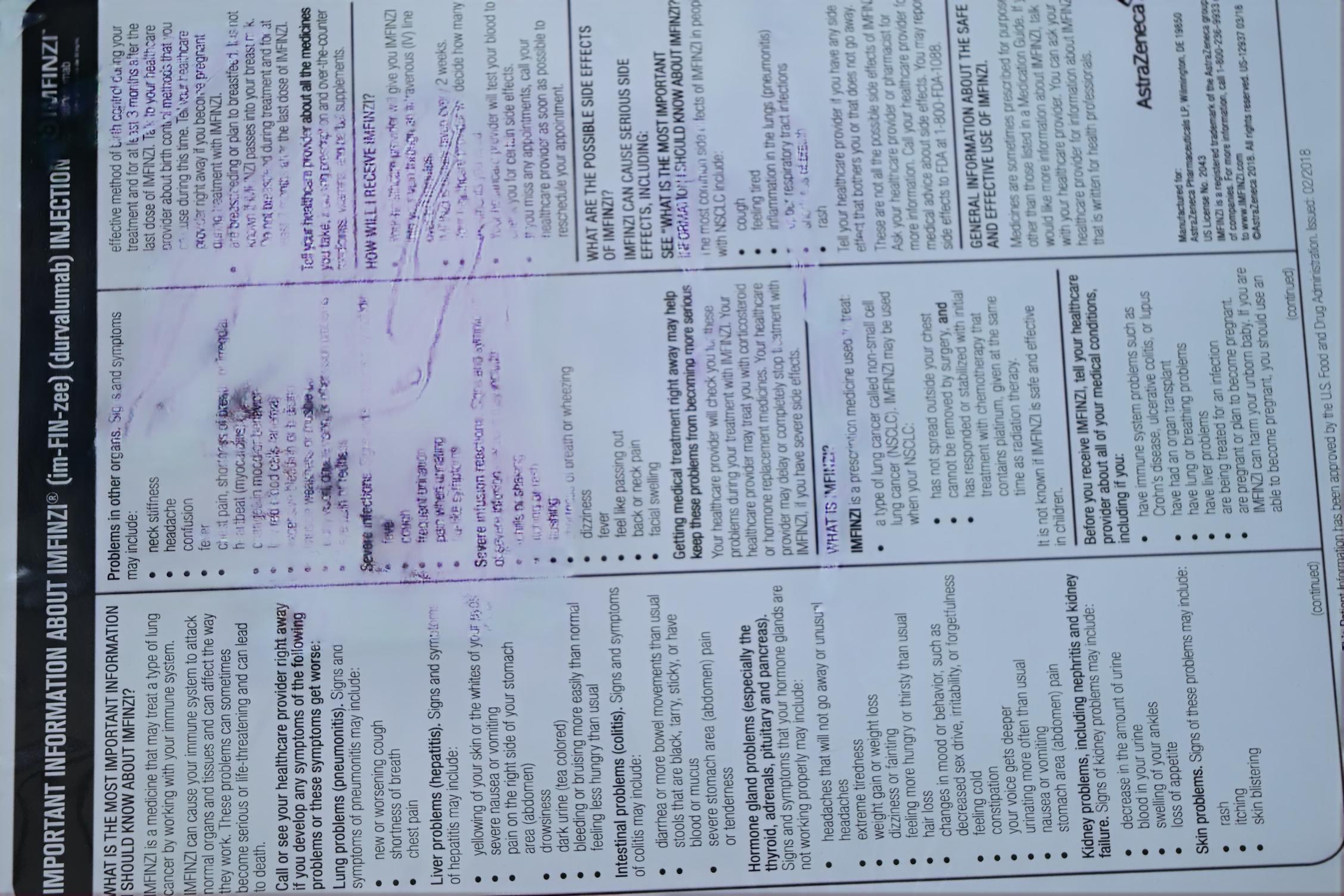} % 替换为实际文件路径
    \end{subfigure}
    \begin{subfigure}[b]{0.138\textwidth}
        \centering
        \includegraphics[angle=-90, width=\linewidth]{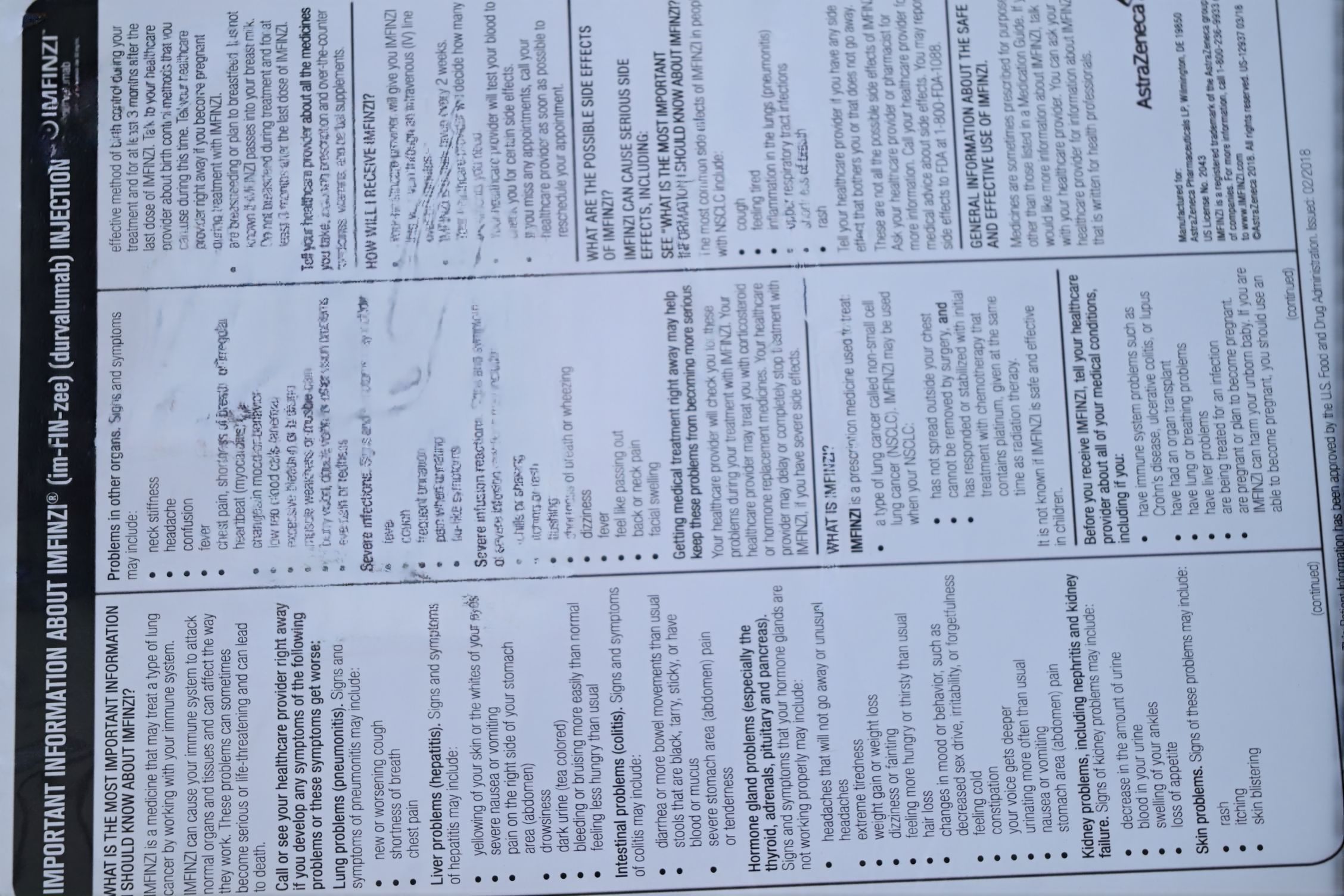} % 替换为实际文件路径
    \end{subfigure}
    \begin{subfigure}[b]{0.138\textwidth}
        \centering
        \includegraphics[angle=-90, width=\linewidth]{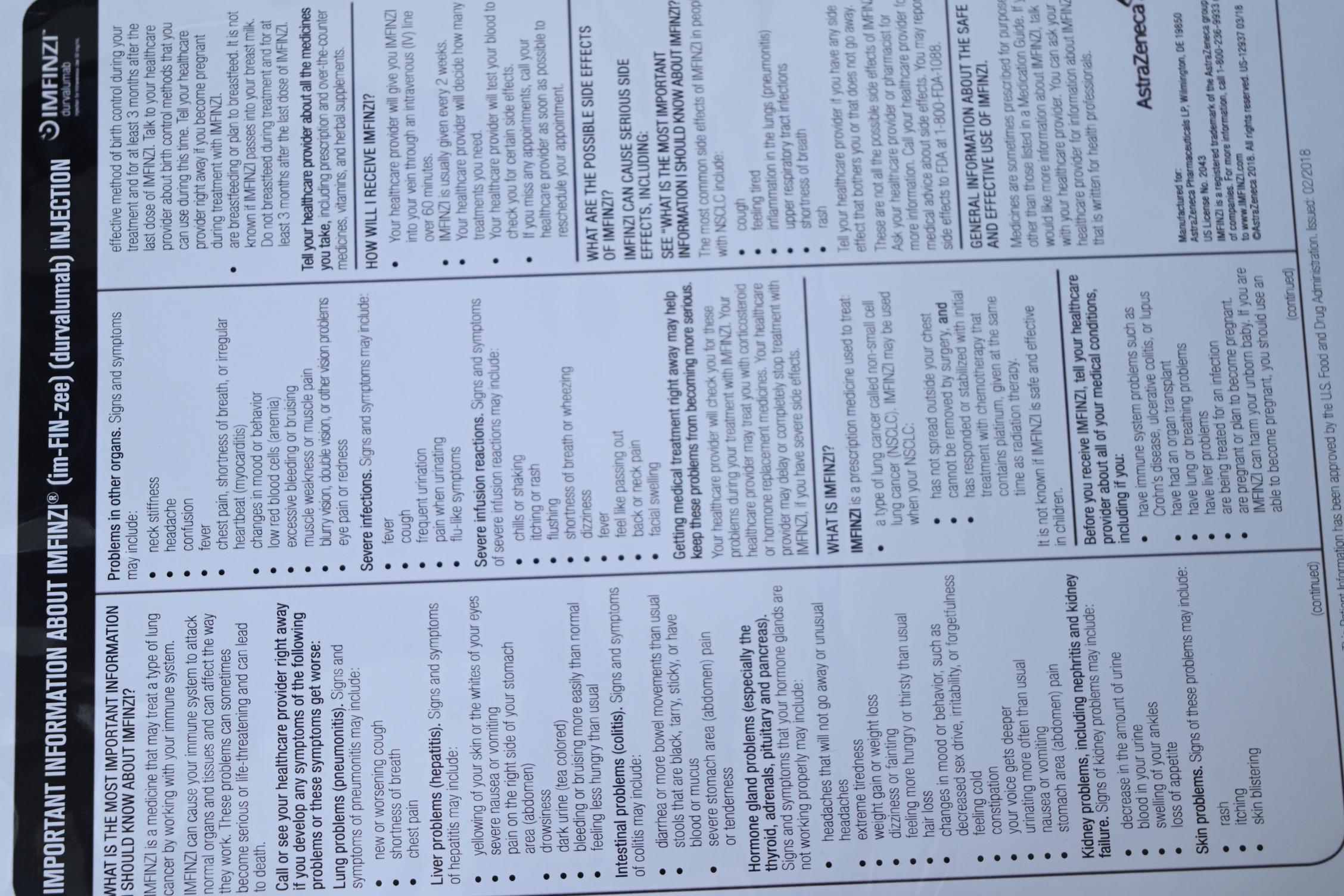} % 替换为实际文件路径
    \end{subfigure} \\
%-------------------------------------------------------------------------
    % 2nd   
    \begin{subfigure}[b]{0.138\textwidth}
        \centering
        \includegraphics[angle=0, width=\linewidth]{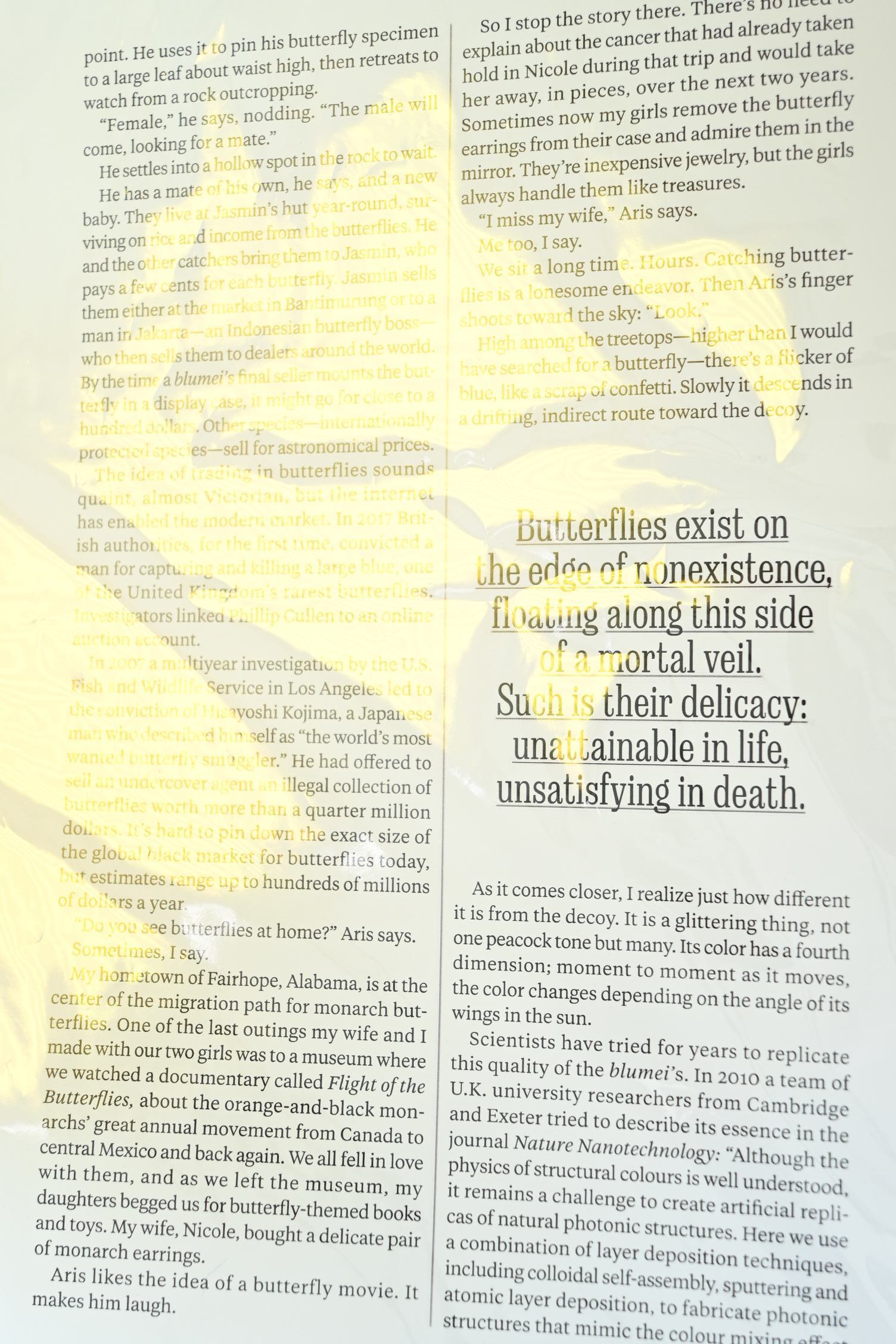}
        % \caption*{Input}
    \end{subfigure}
    \begin{subfigure}[b]{0.138\textwidth}
        \centering
        \includegraphics[angle=0, width=\linewidth]{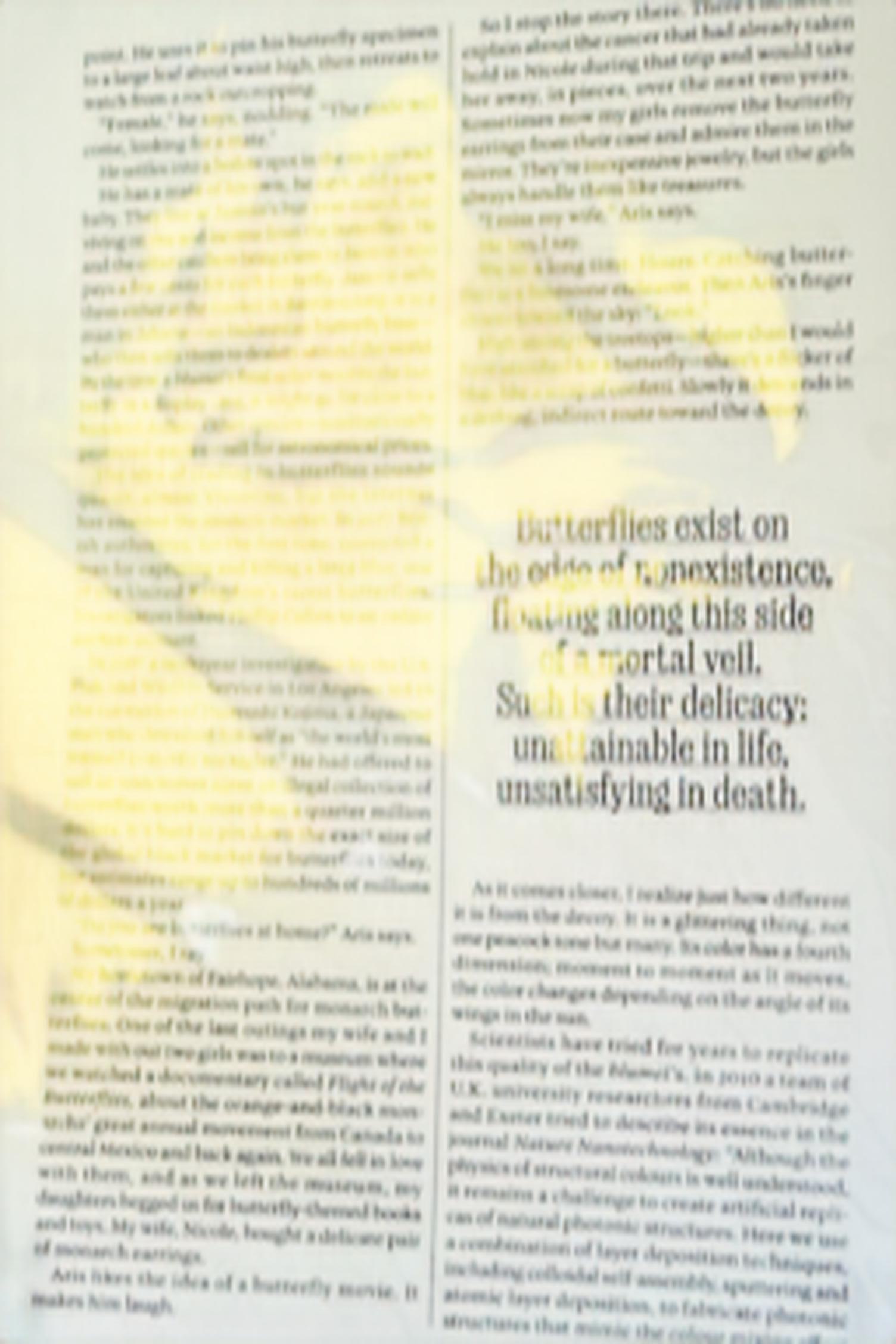}
        % \caption{JSHDR}
    \end{subfigure}
    \begin{subfigure}[b]{0.138\textwidth}
        \centering
        \includegraphics[angle=0, width=\linewidth]{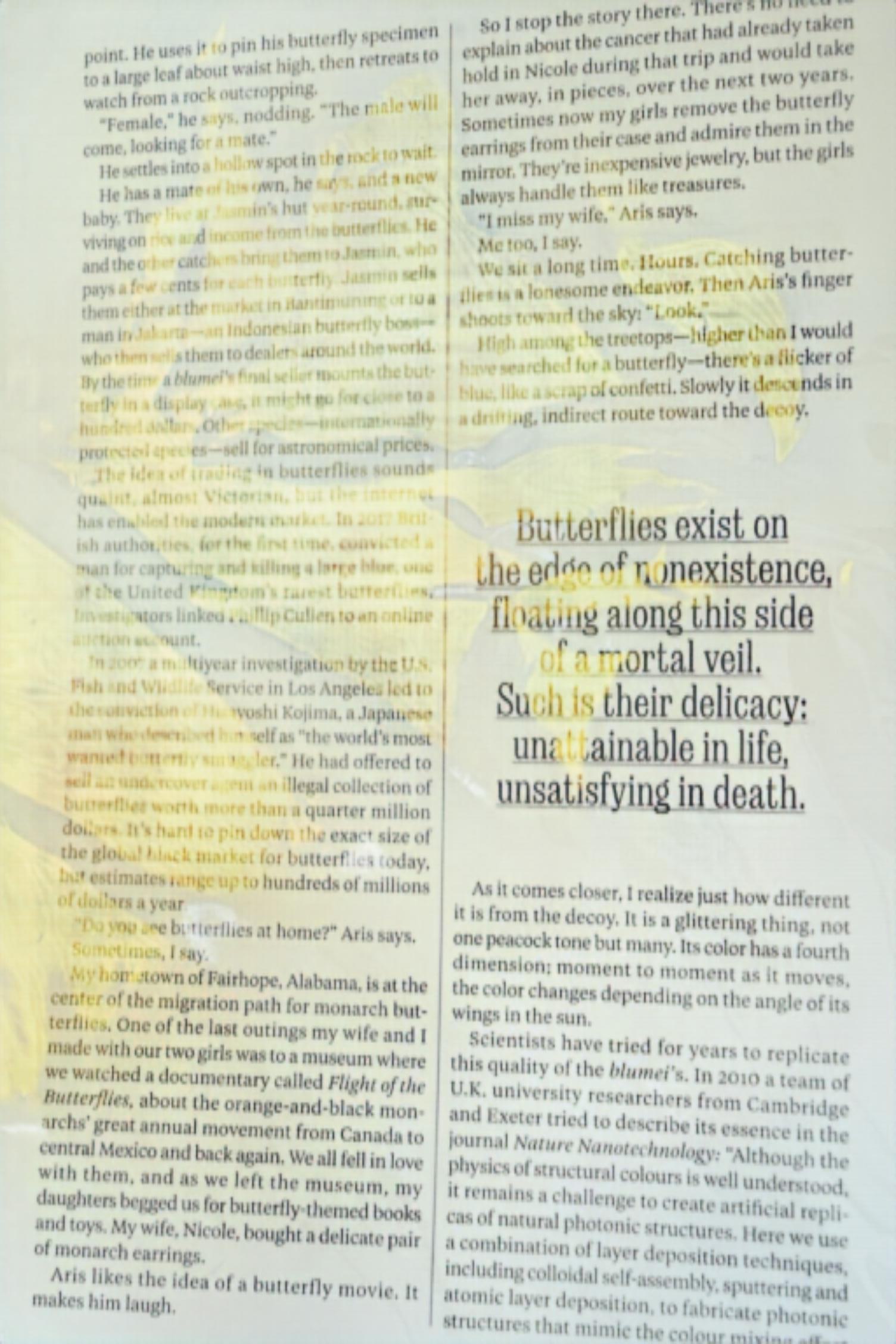}
        % \caption{TSHRNet}
    \end{subfigure}
    \begin{subfigure}[b]{0.138\textwidth}
        \centering
        \includegraphics[angle=0, width=\linewidth]{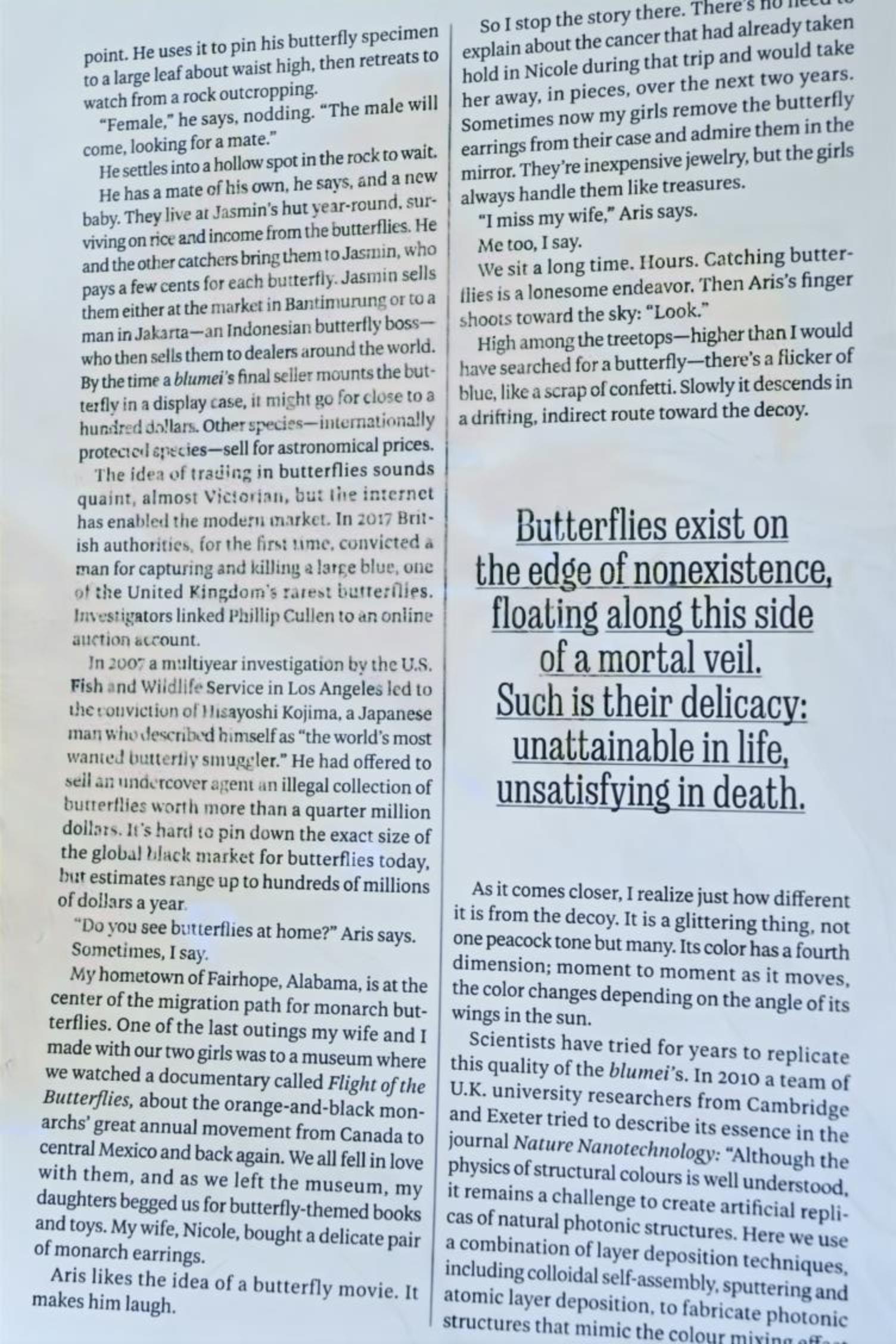}
        % \caption{DHAN-SHR}
    \end{subfigure}
    \begin{subfigure}[b]{0.138\textwidth}
        \centering
        \includegraphics[angle=0, width=\linewidth]{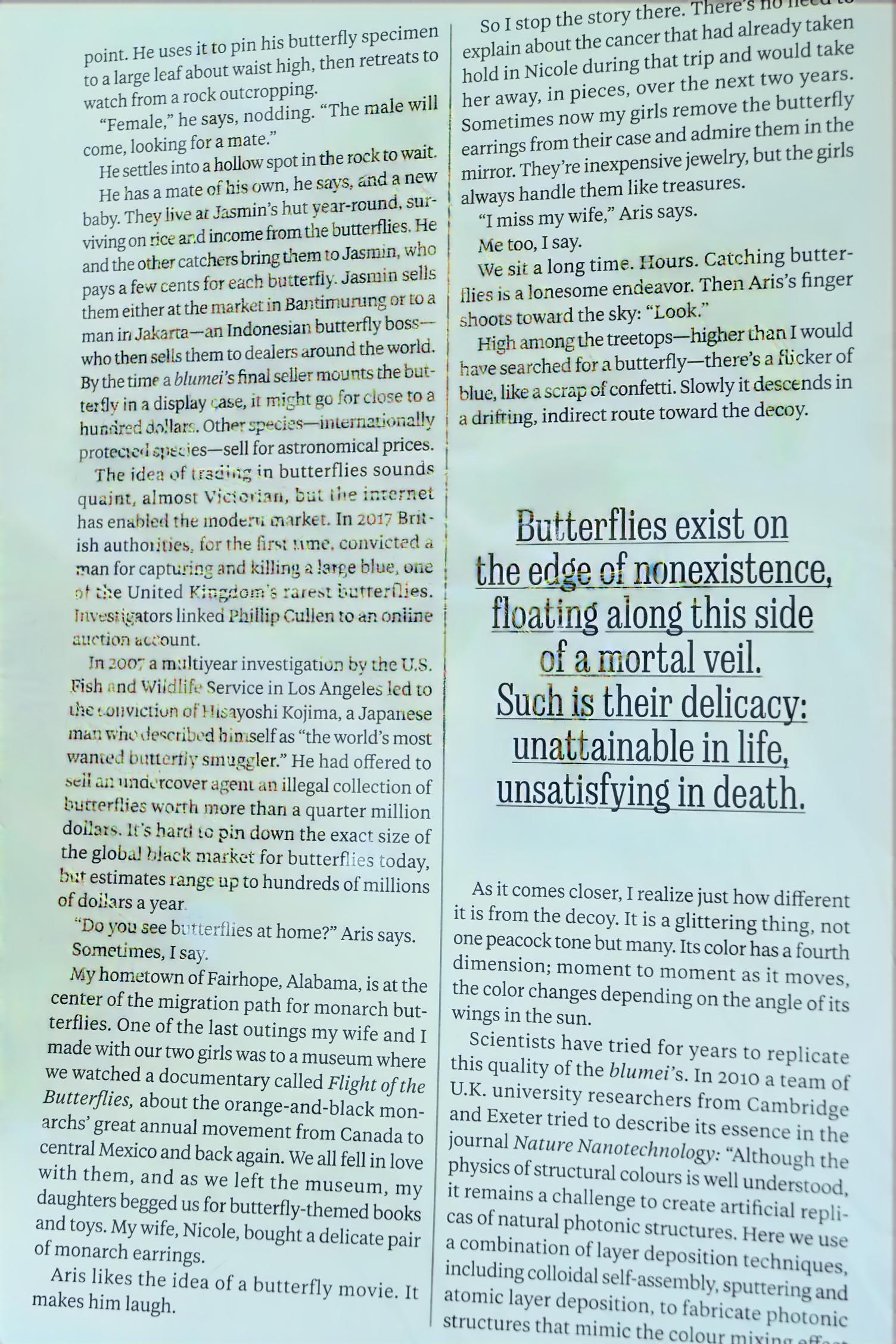}
        % \caption{DocShadowNet}
    \end{subfigure}
    \begin{subfigure}[b]{0.138\textwidth}
        \centering
        \includegraphics[angle=0, width=\linewidth]{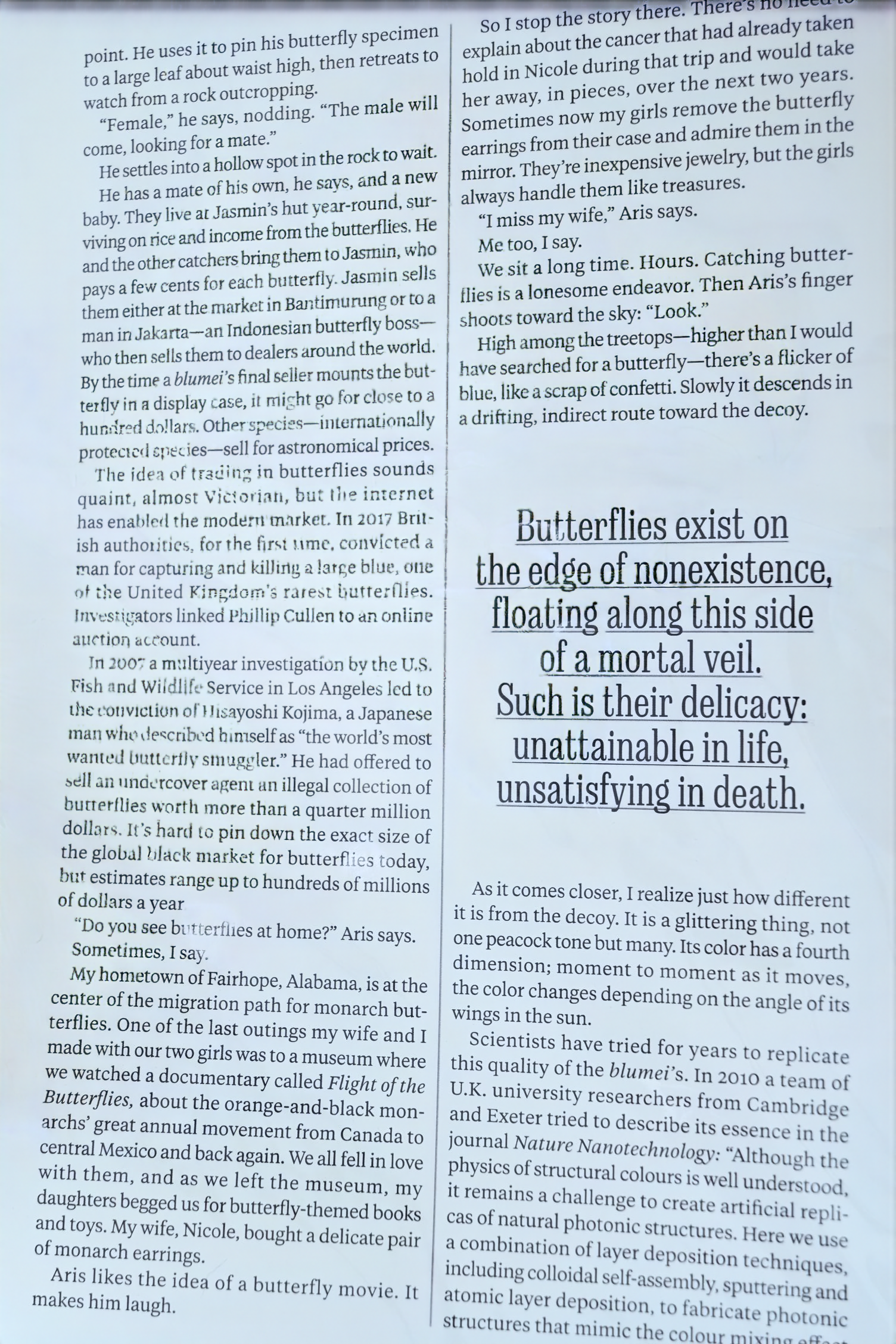}
        % \caption*{Ours}
    \end{subfigure}
        \begin{subfigure}[b]{0.138\textwidth}
        \centering
        \includegraphics[angle=0, width=\linewidth]{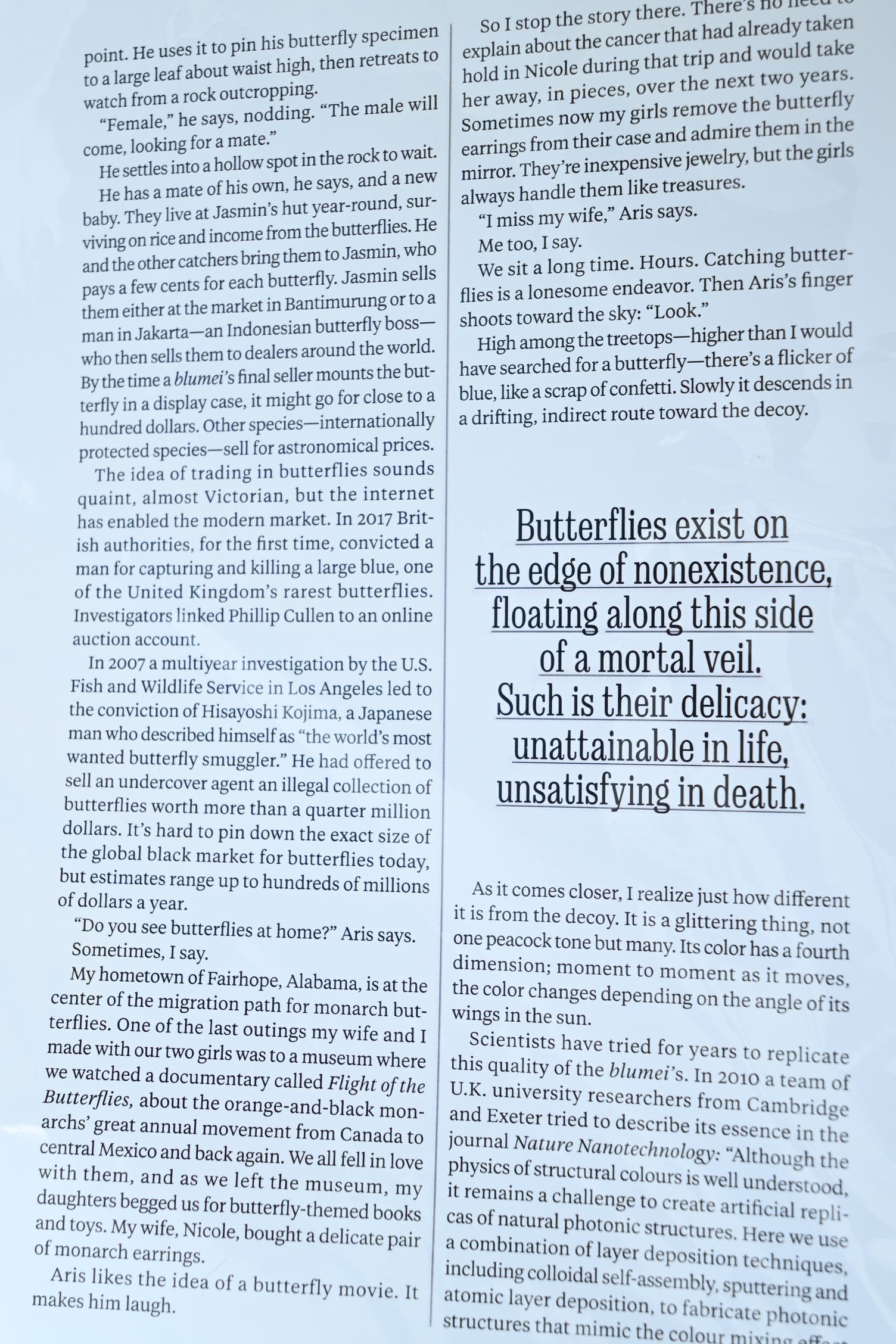}
        % \caption*{Reference}
    \end{subfigure}
    % the third row   
    \begin{subfigure}[b]{0.138\textwidth}
        \centering
        \includegraphics[angle=90, width=\linewidth]{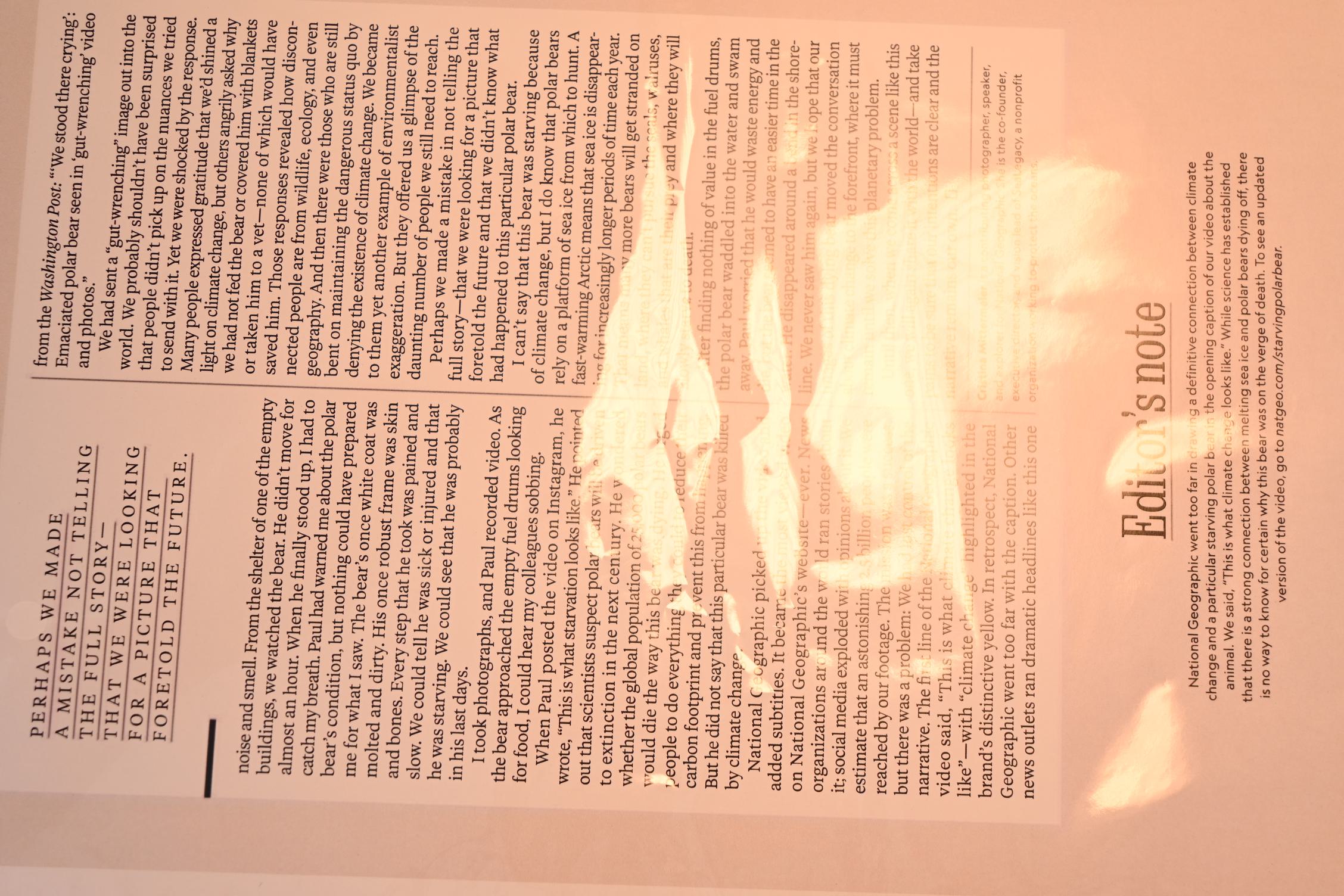}
        % \caption*{Input}
    \end{subfigure}
    \begin{subfigure}[b]{0.138\textwidth}
        \centering
        \includegraphics[angle=90, width=\linewidth]{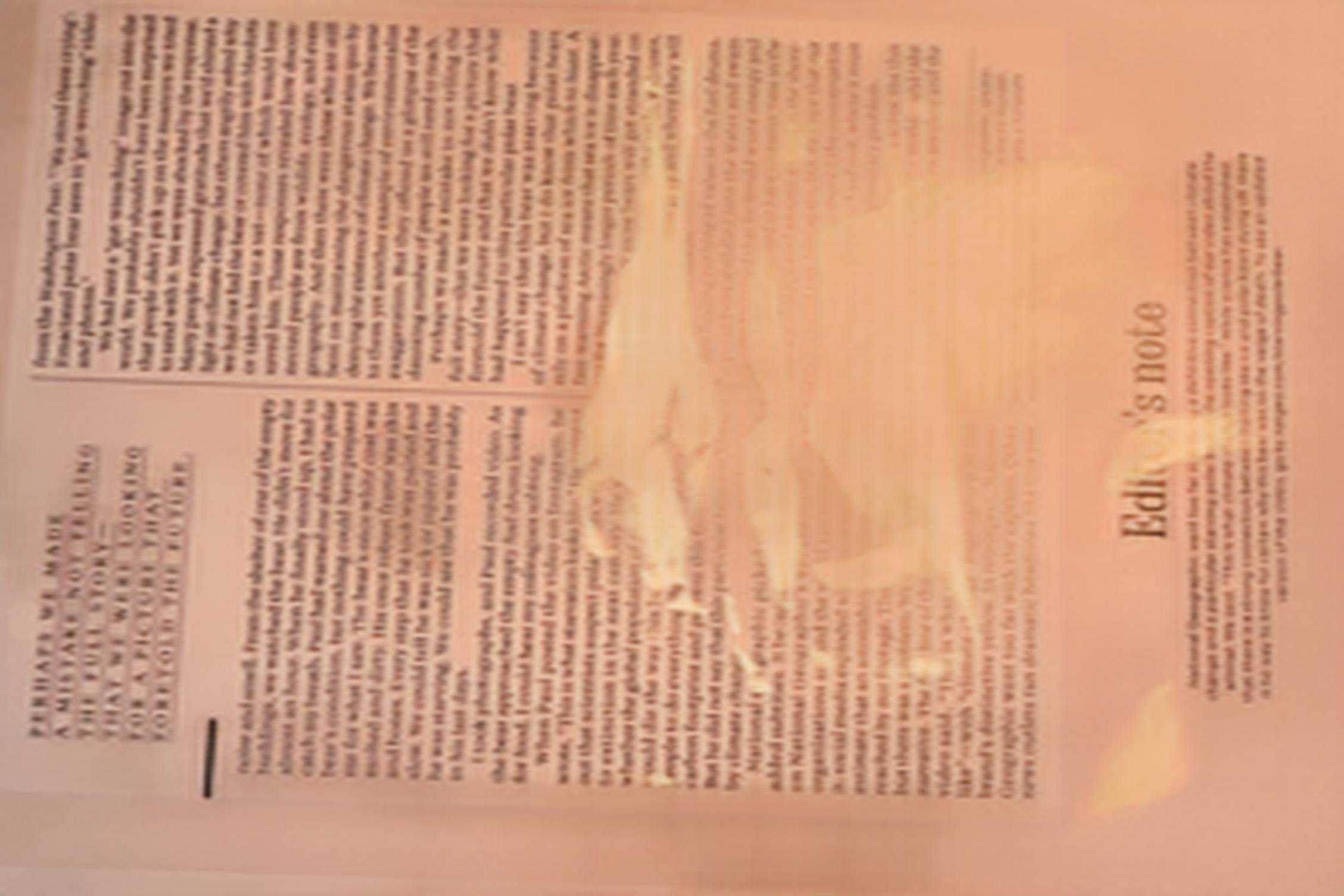}
        % \caption{JSHDR}
    \end{subfigure}
    \begin{subfigure}[b]{0.138\textwidth}
        \centering
        \includegraphics[angle=90, width=\linewidth]{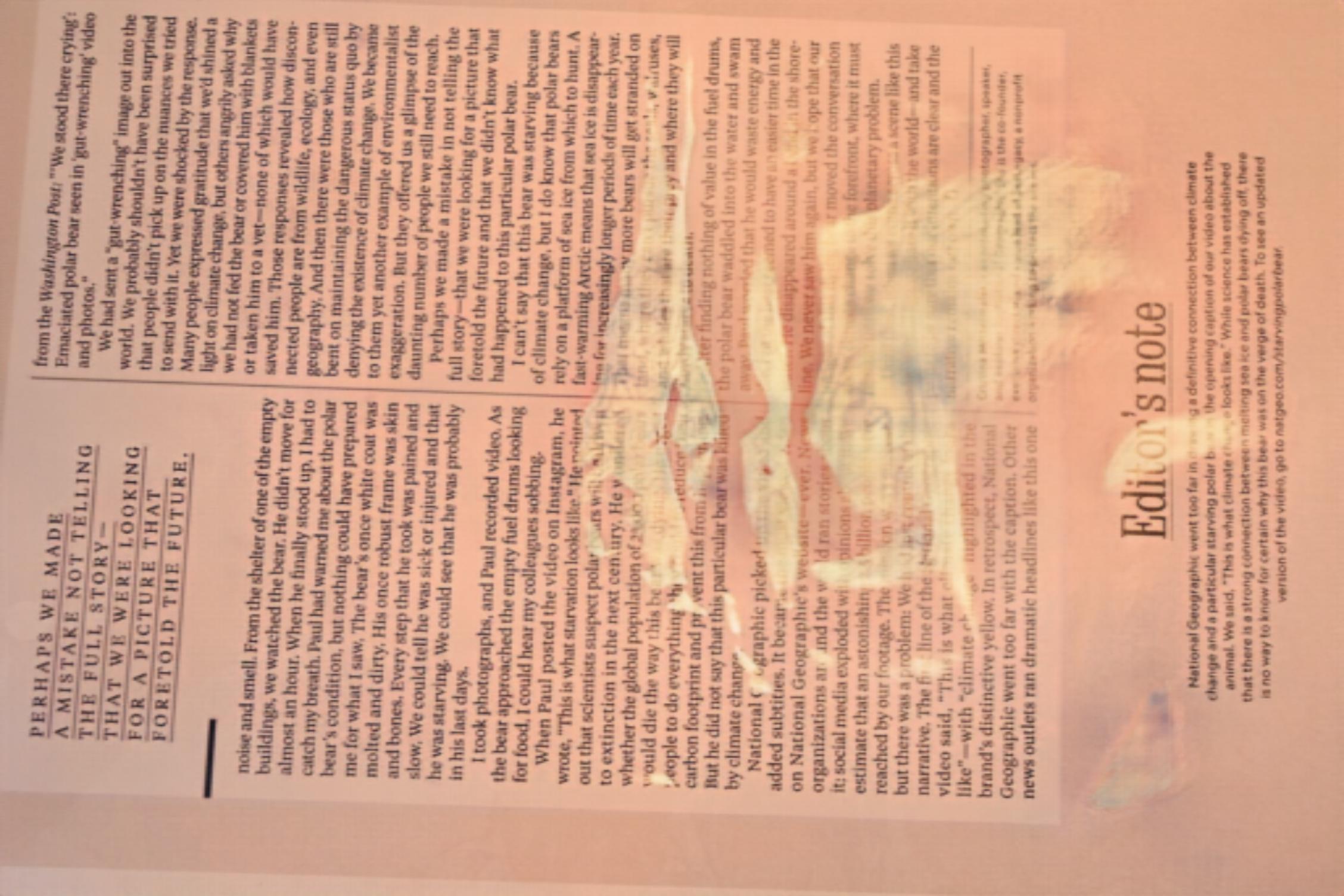}
        % \caption{TSHRNet}
    \end{subfigure}
    \begin{subfigure}[b]{0.138\textwidth}
        \centering
        \includegraphics[angle=90, width=\linewidth]{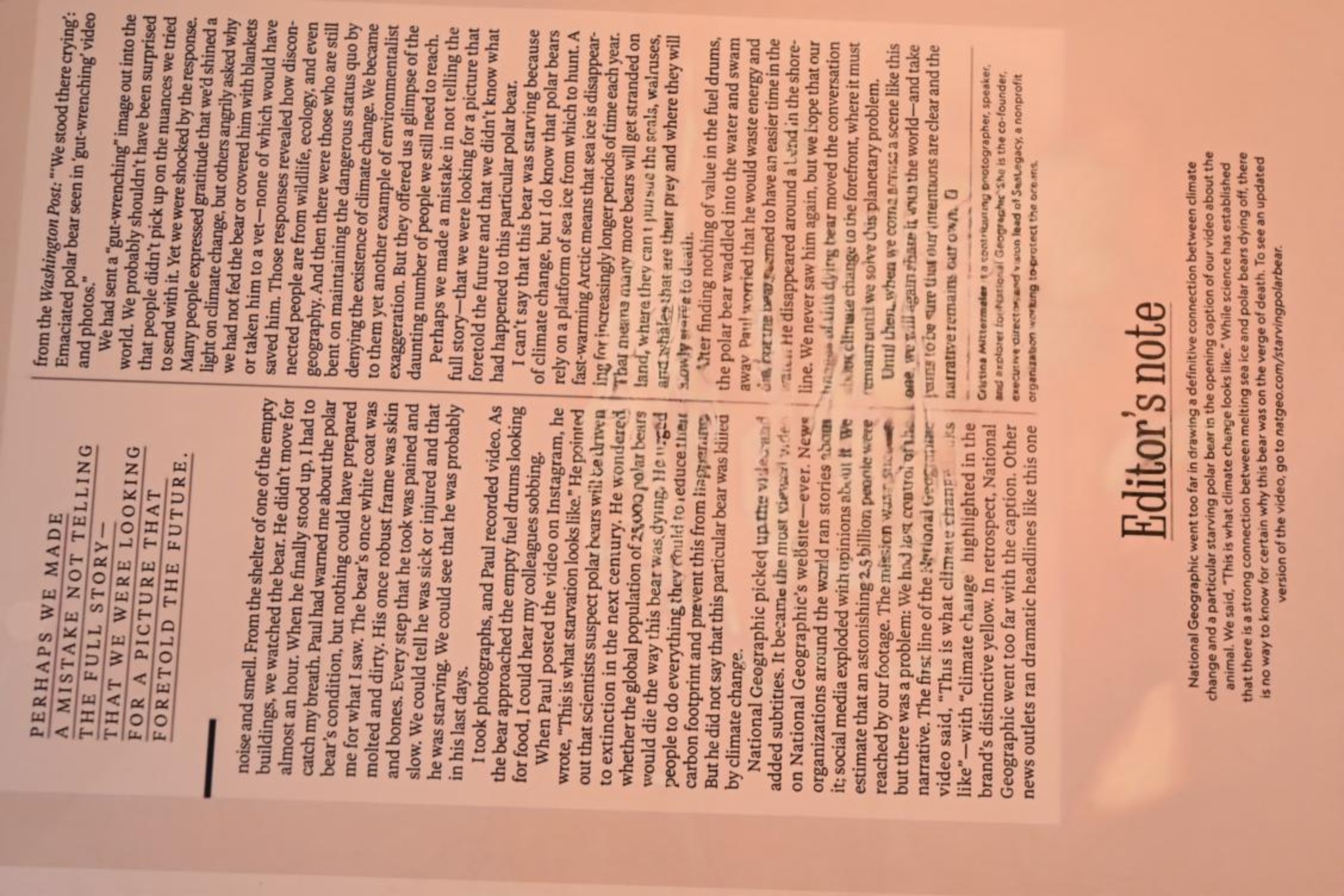}
        % \caption{DHAN-SHR}
    \end{subfigure}
    \begin{subfigure}[b]{0.138\textwidth}
        \centering
        \includegraphics[angle=90, width=\linewidth]{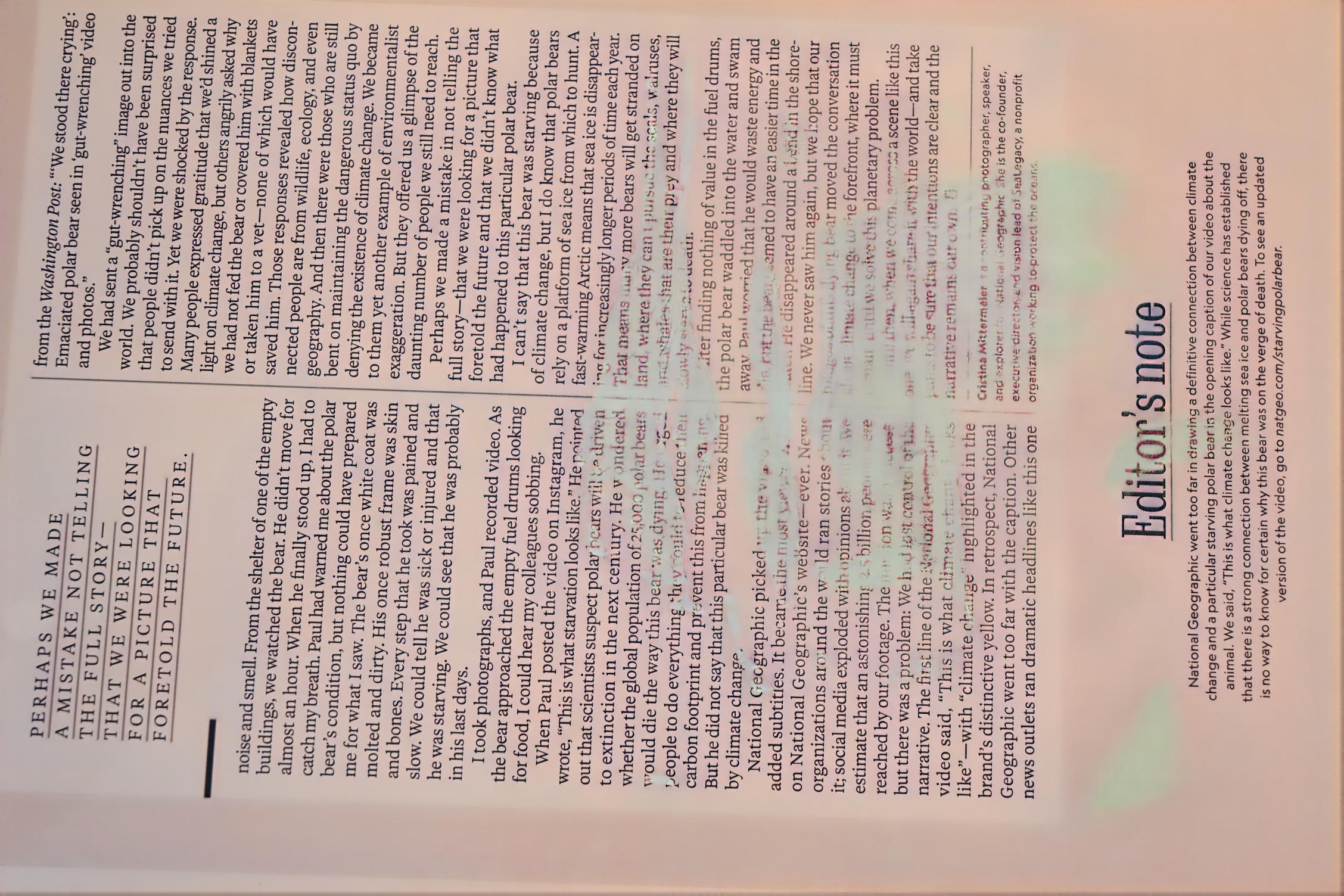}
        % \caption{DocShadowNet}
    \end{subfigure}
    \begin{subfigure}[b]{0.138\textwidth}
        \centering
        \includegraphics[angle=90, width=\linewidth]{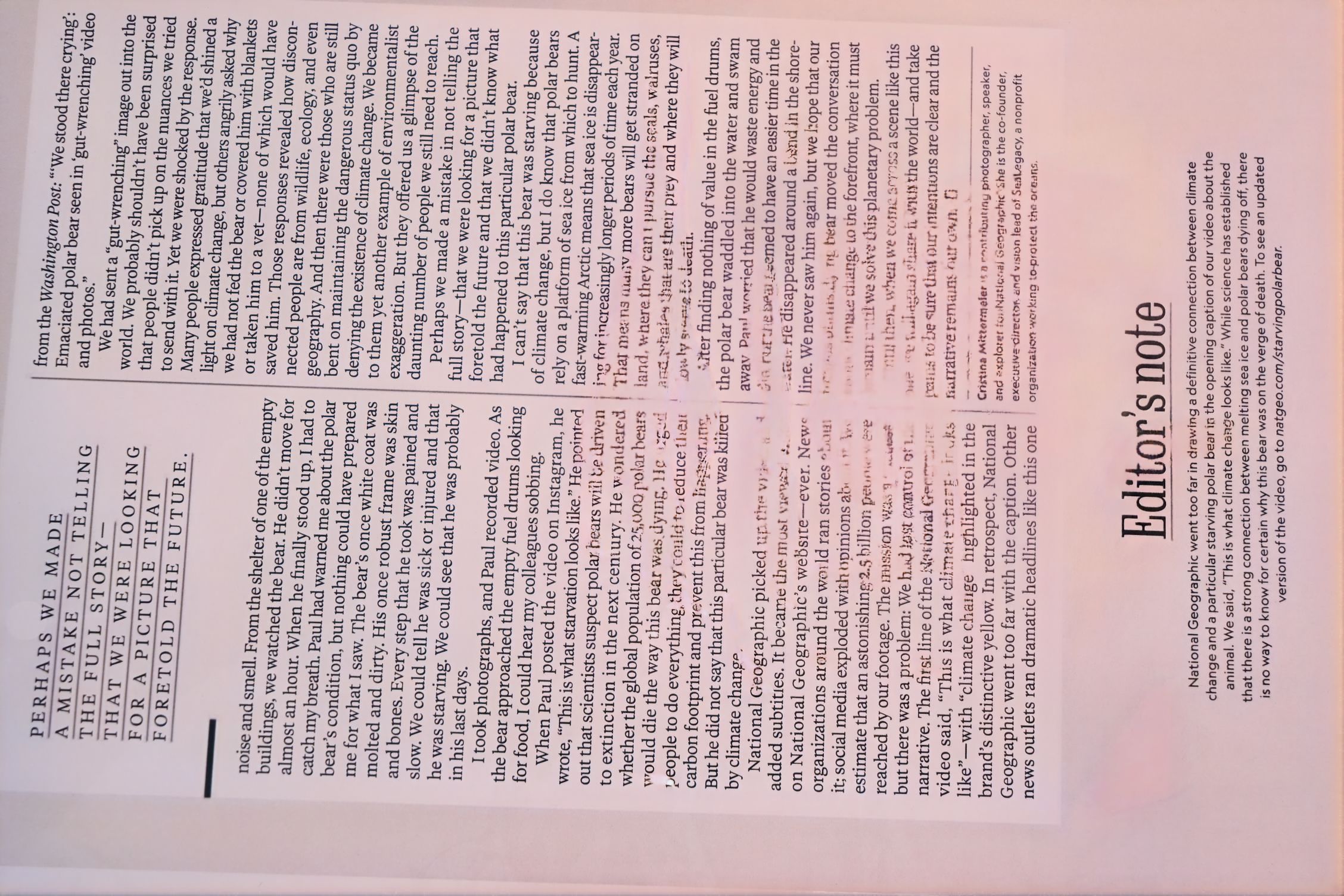}
        % \caption*{Ours}
    \end{subfigure}
        \begin{subfigure}[b]{0.138\textwidth}
        \centering
        \includegraphics[angle=90, width=\linewidth]{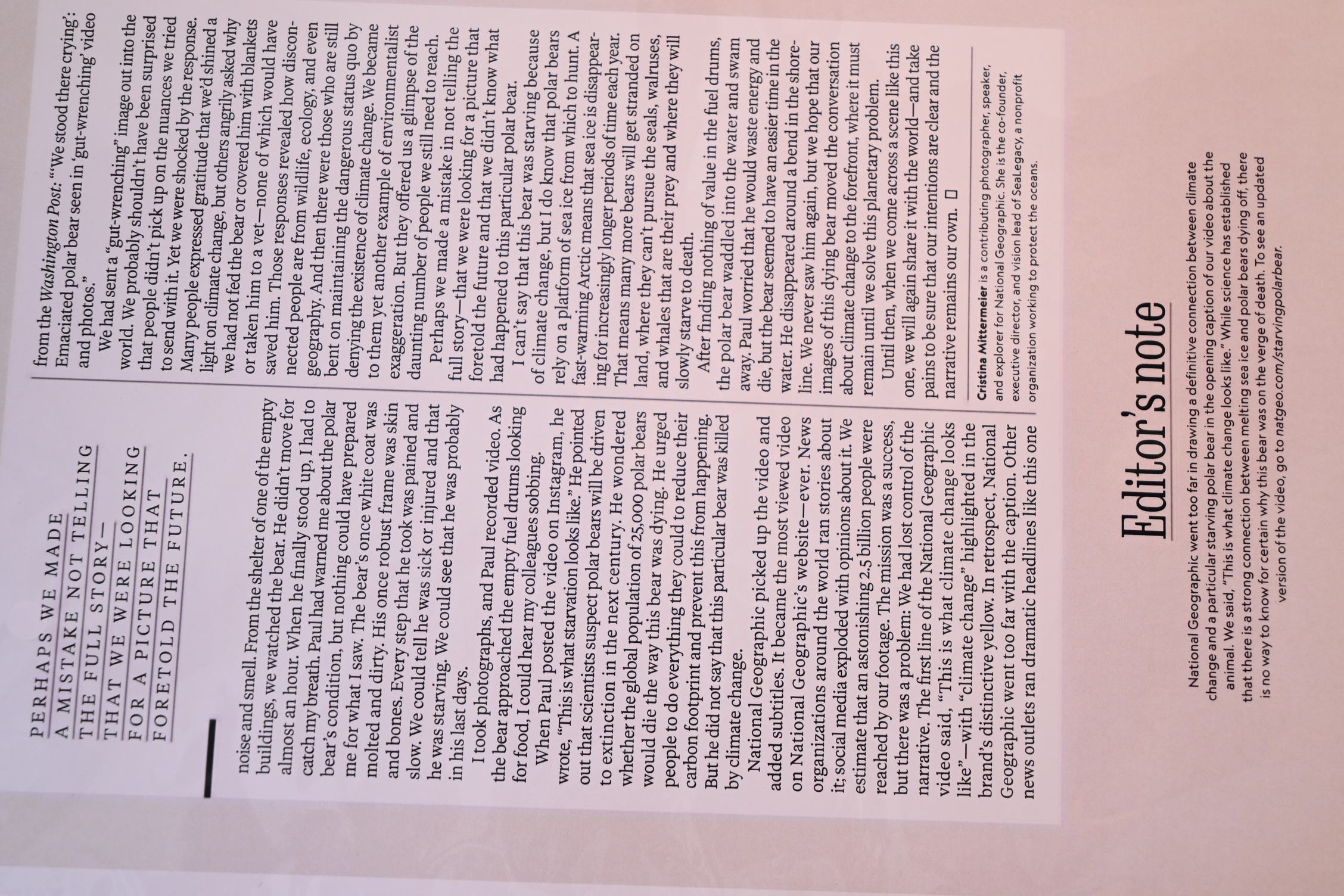}
        % \caption*{Reference}
    \end{subfigure}
%-------------------------------------------------------------------------
    % 3rd   
    \begin{subfigure}[b]{0.138\textwidth}
        \centering
        \includegraphics[angle=0, width=\linewidth]{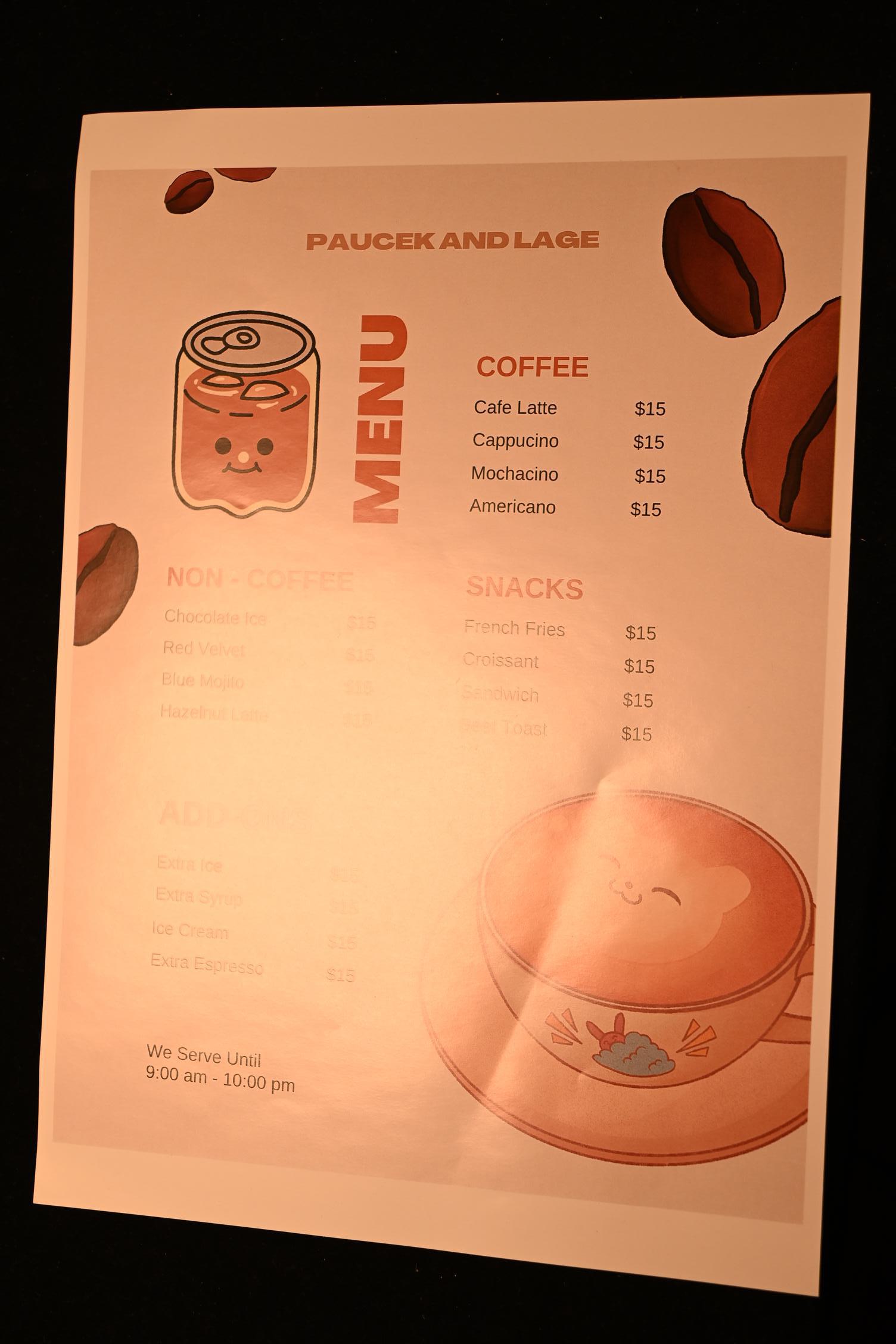}
        % \caption*{Input}
    \end{subfigure}
    \begin{subfigure}[b]{0.138\textwidth}
        \centering
        \includegraphics[angle=0, width=\linewidth]{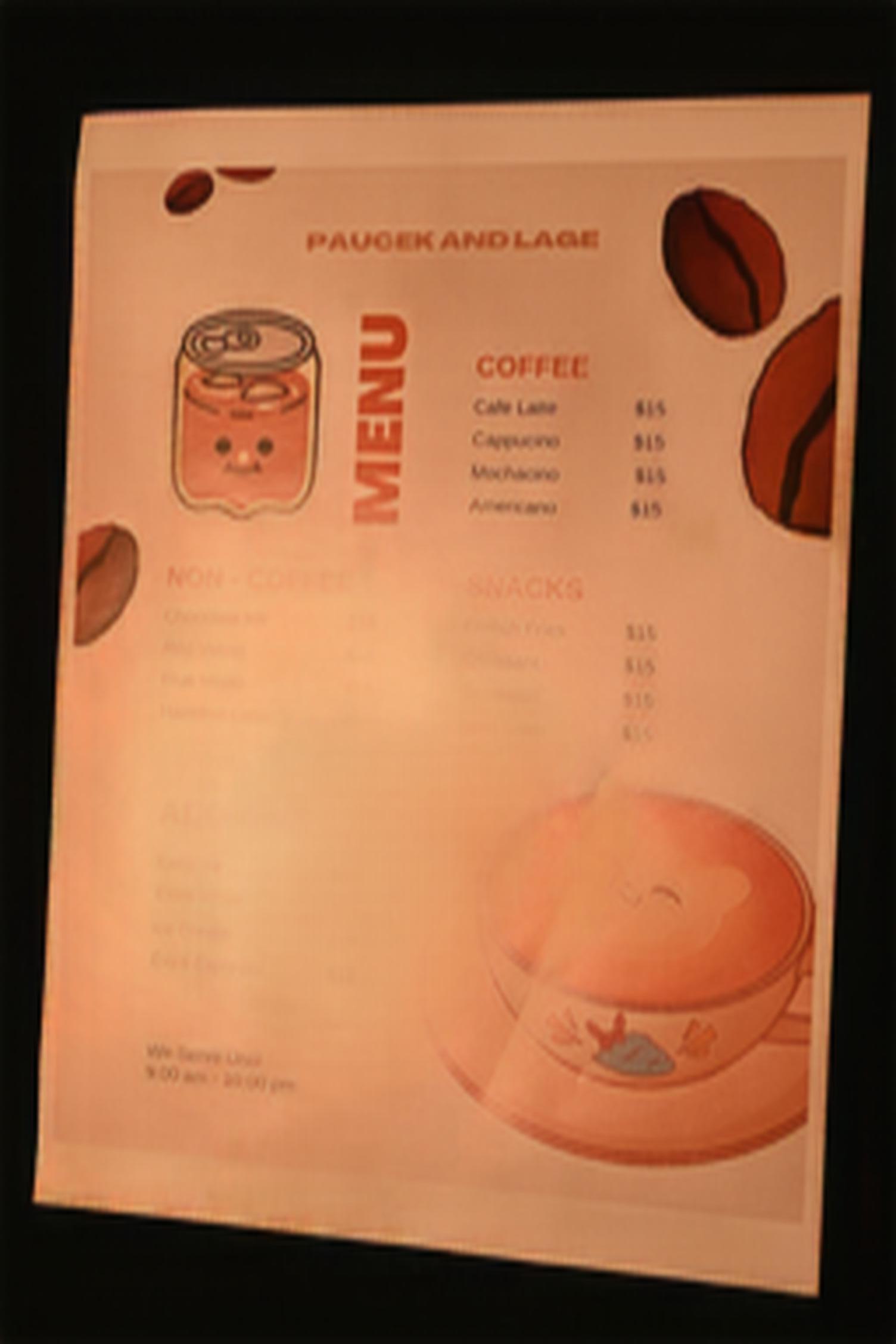}
        % \caption{JSHDR}
    \end{subfigure}
    \begin{subfigure}[b]{0.138\textwidth}
        \centering
        \includegraphics[angle=0, width=\linewidth]{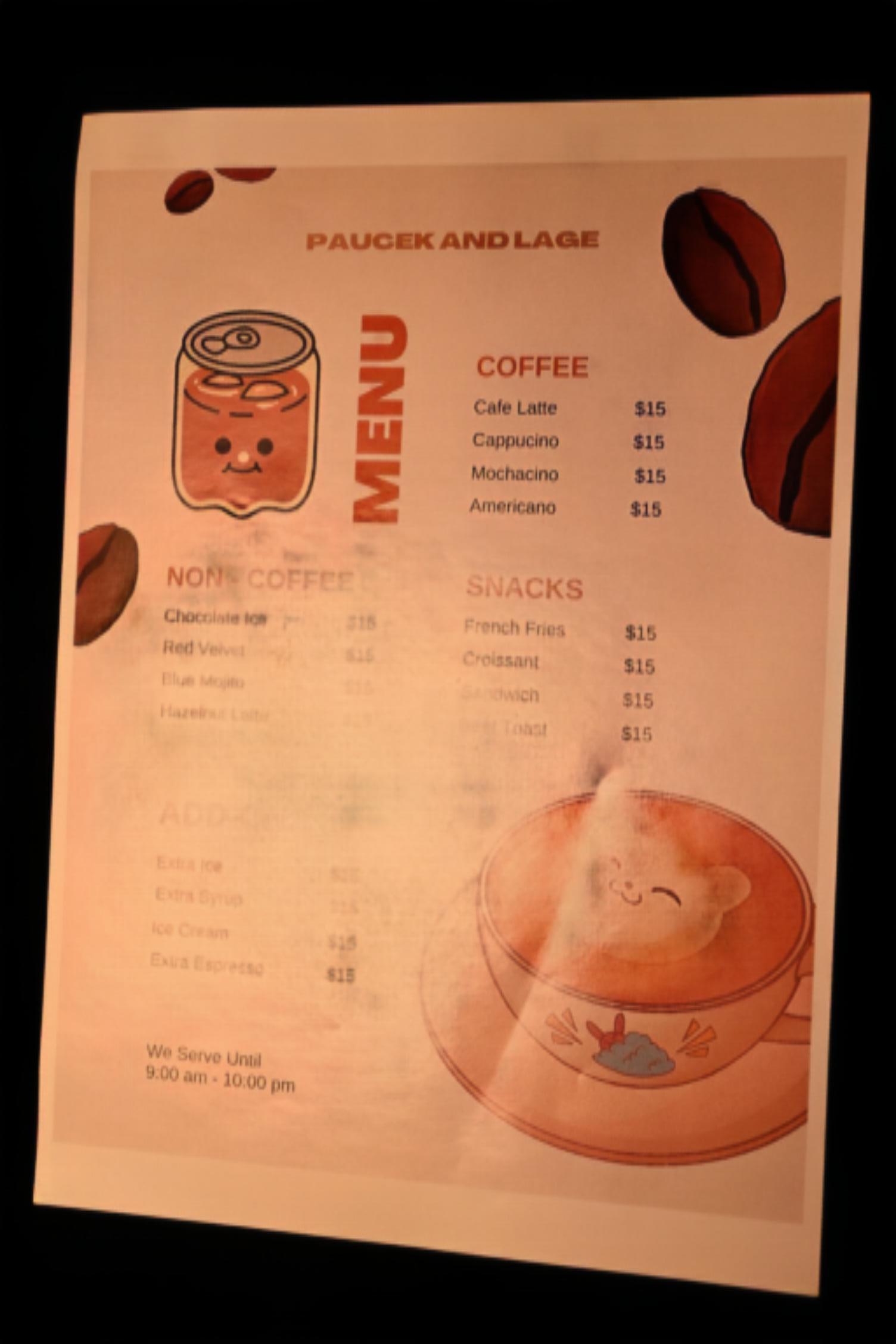}
        % \caption{TSHRNet}
    \end{subfigure}
    \begin{subfigure}[b]{0.138\textwidth}
        \centering
        \includegraphics[angle=0, width=\linewidth]{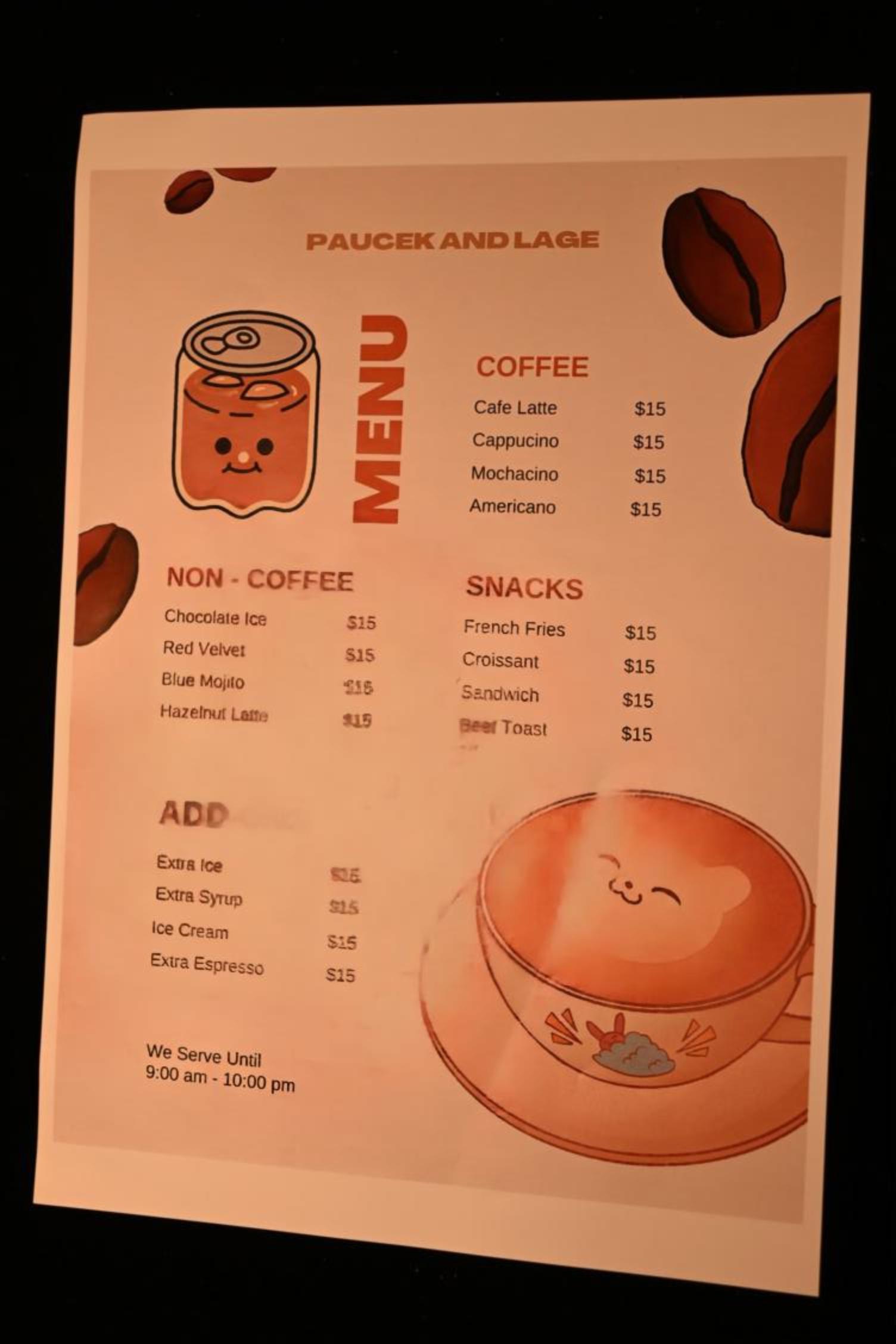}
        % \caption{DHAN-SHR}
    \end{subfigure}
    \begin{subfigure}[b]{0.138\textwidth}
        \centering
        \includegraphics[angle=0, width=\linewidth]{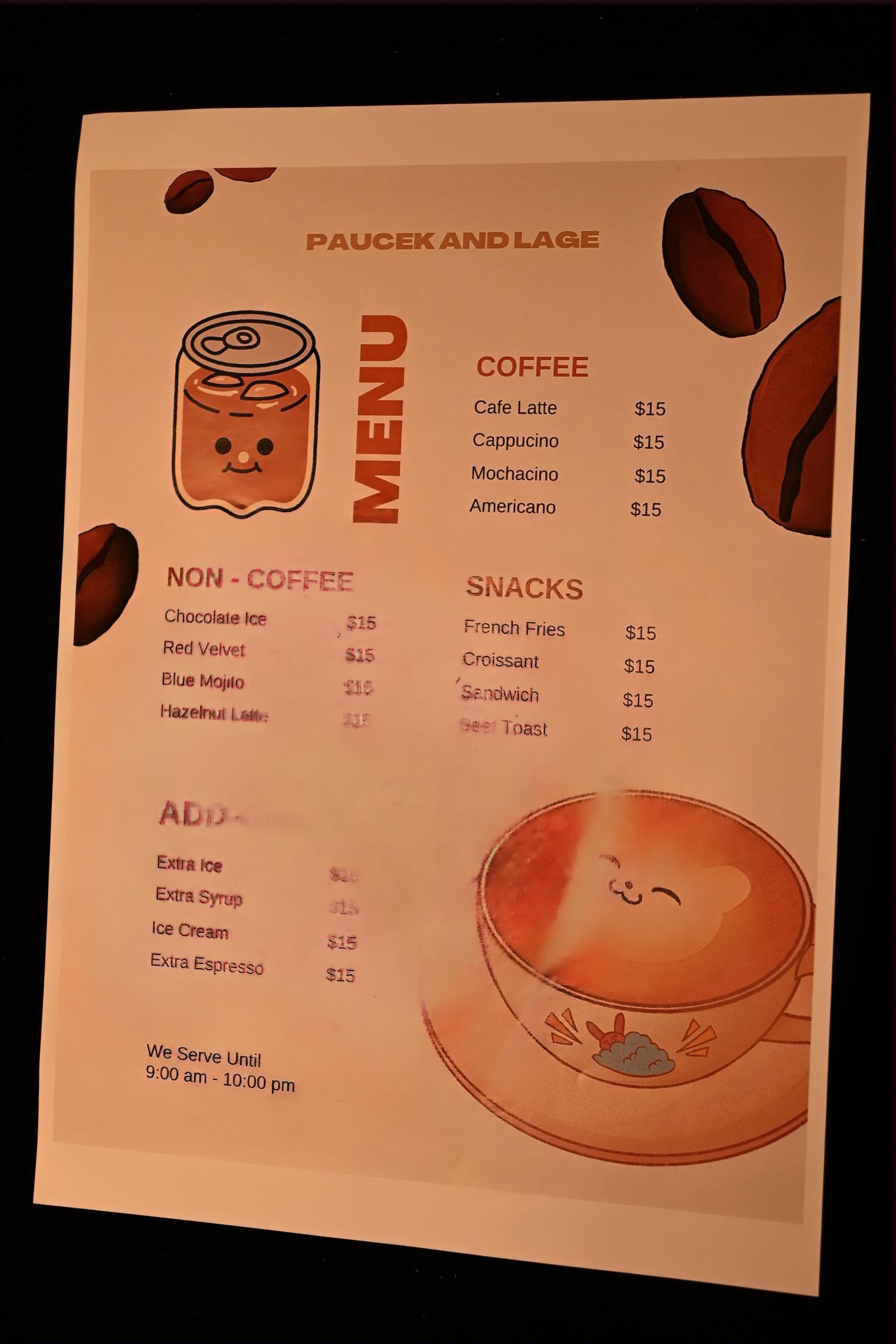}
        % \caption{DocShadowNet}
    \end{subfigure}
    \begin{subfigure}[b]{0.138\textwidth}
        \centering
        \includegraphics[angle=90, width=\linewidth]{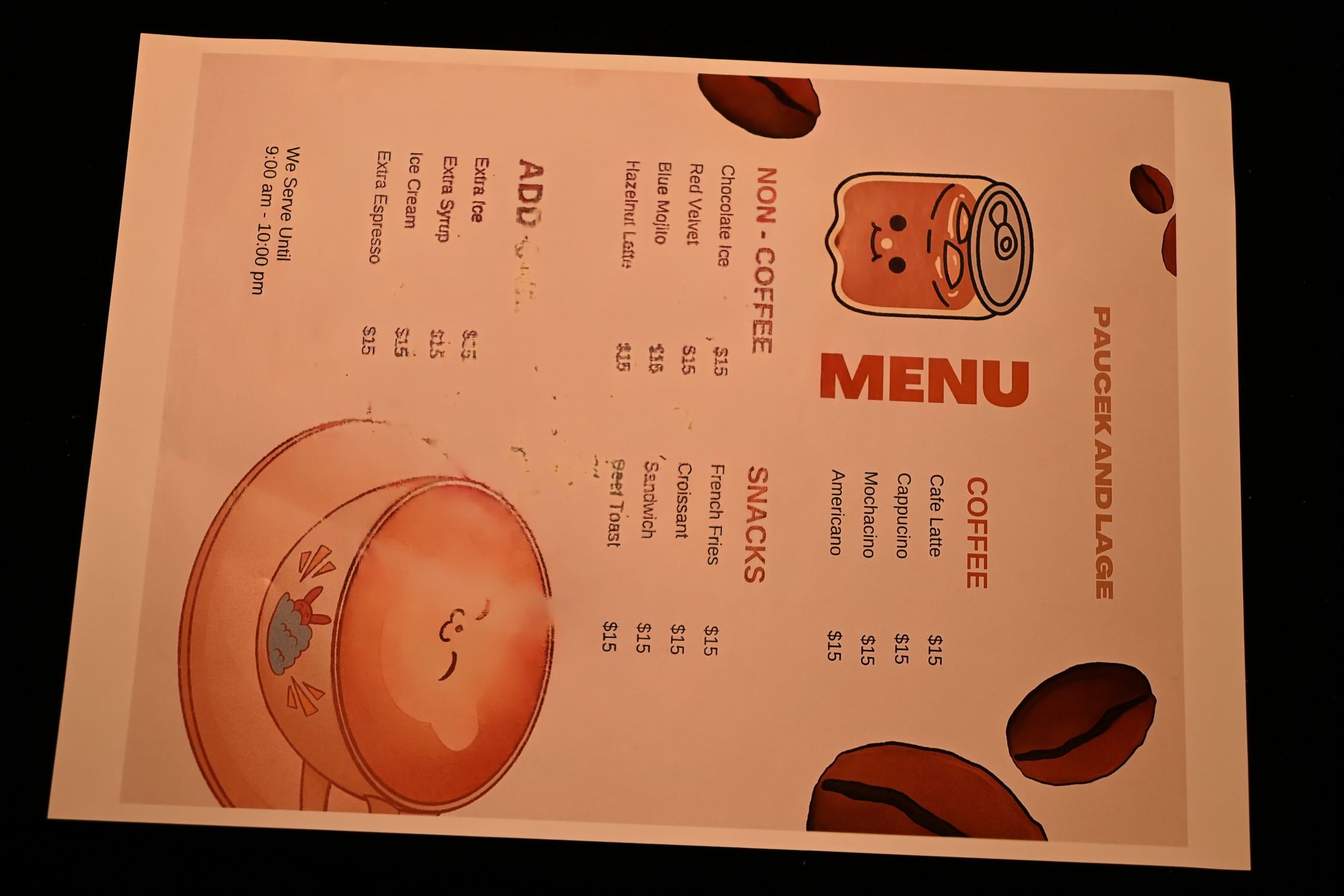}
        % \caption*{Ours}
    \end{subfigure}
        \begin{subfigure}[b]{0.138\textwidth}
        \centering
        \includegraphics[angle=0, width=\linewidth]{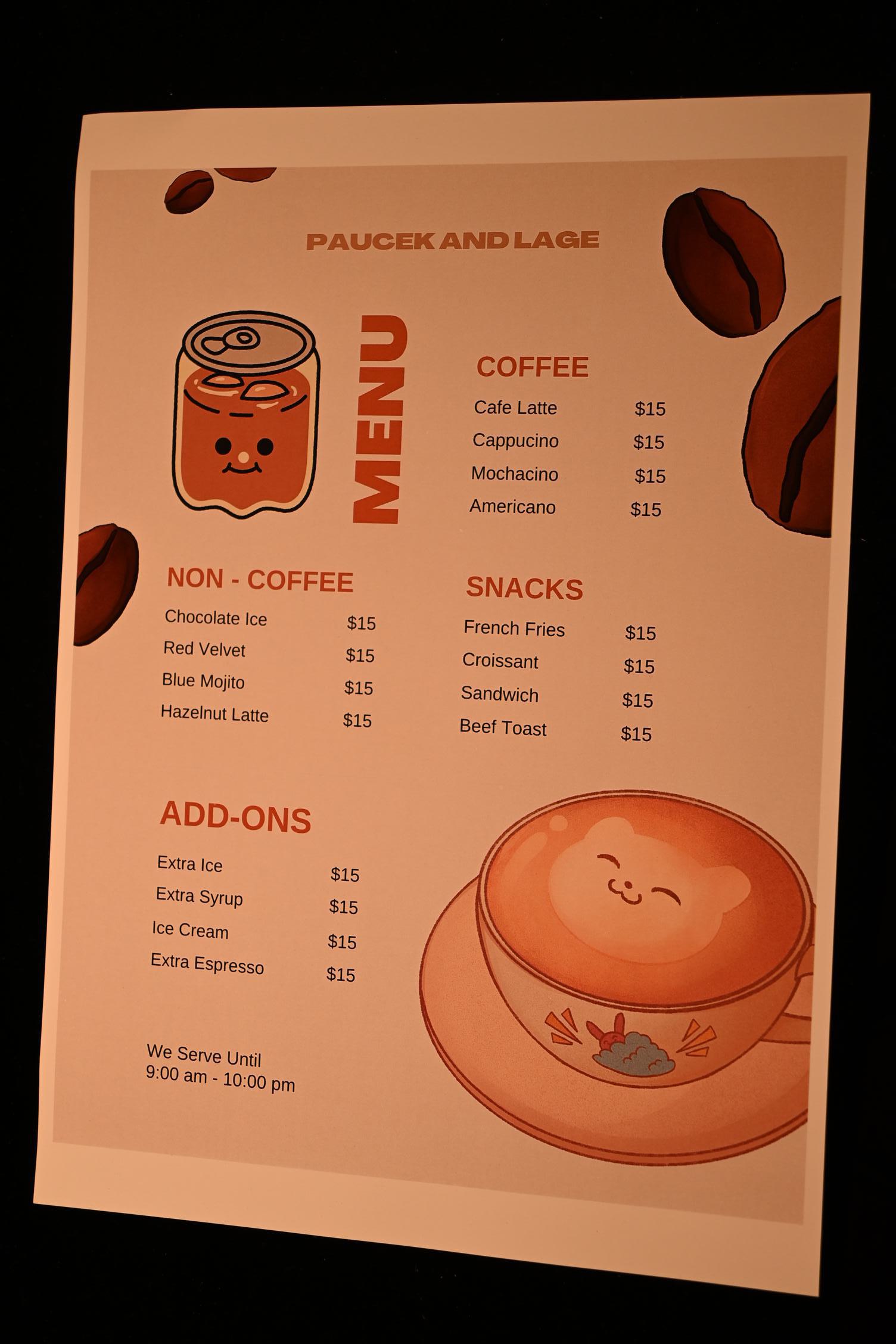}
        % \caption*{Reference}
    \end{subfigure} 
%-------------------------------------------------------------------------
 
%-------------------------------------------------------------------------
    
    %the fourth row  
    \begin{subfigure}[b]{0.138\textwidth}
        \centering
        \includegraphics[angle=-90, width=\linewidth]{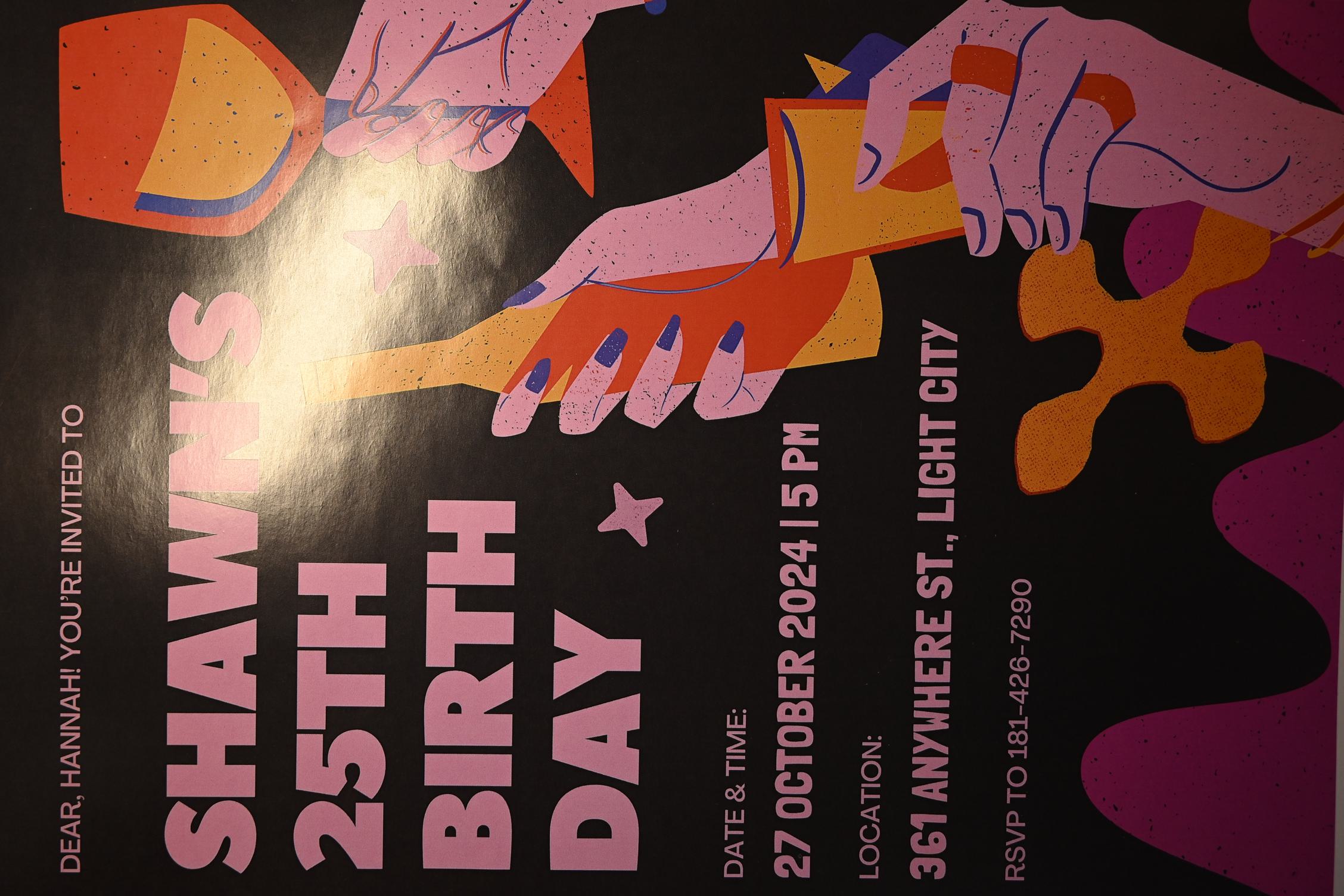}
        \caption*{Input}
    \end{subfigure}
    \begin{subfigure}[b]{0.138\textwidth}
        \centering
        \includegraphics[angle=-90, width=\linewidth]{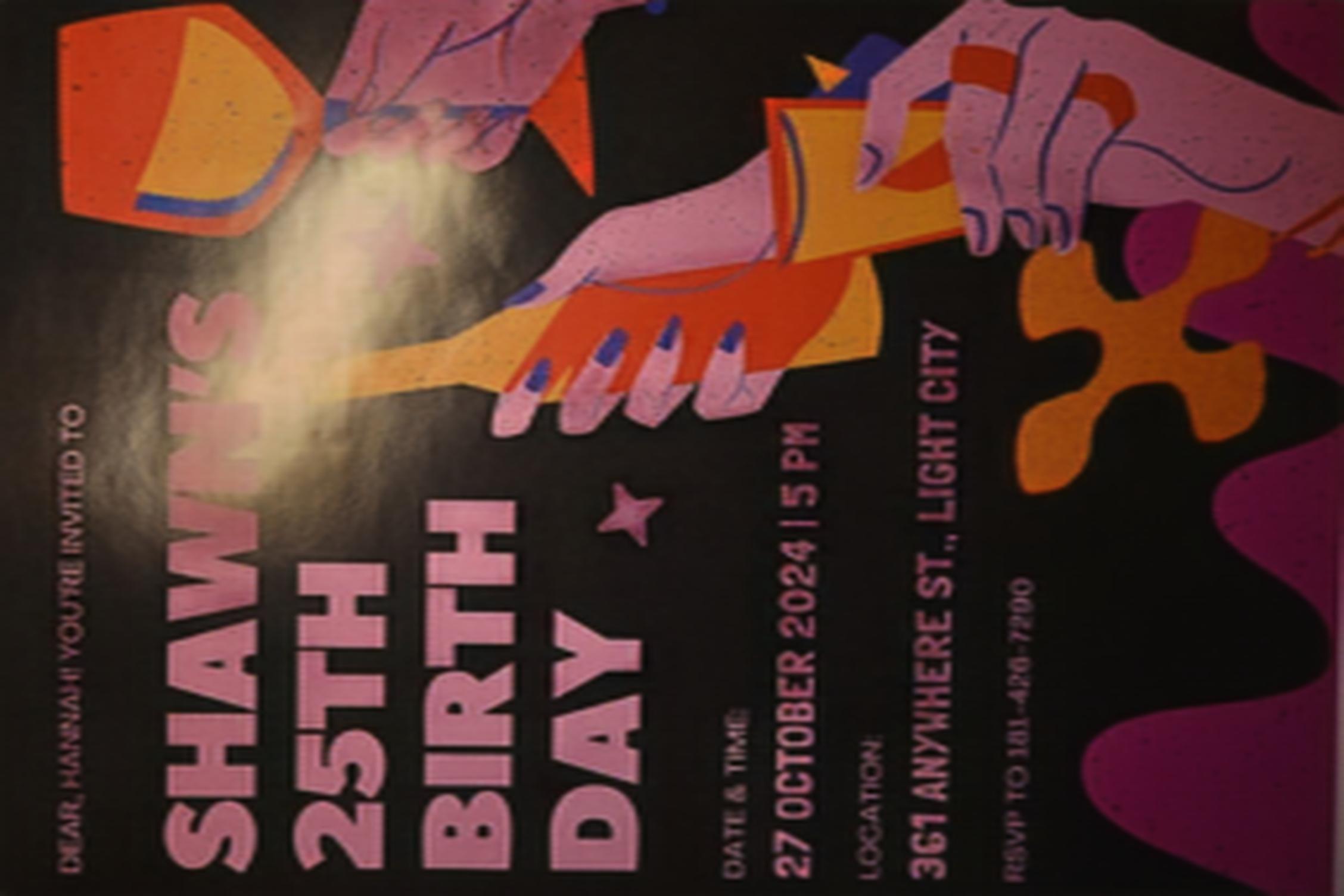}
        \caption{JSHDR}
    \end{subfigure}
    \begin{subfigure}[b]{0.138\textwidth}
        \centering
        \includegraphics[angle=-90, width=\linewidth]{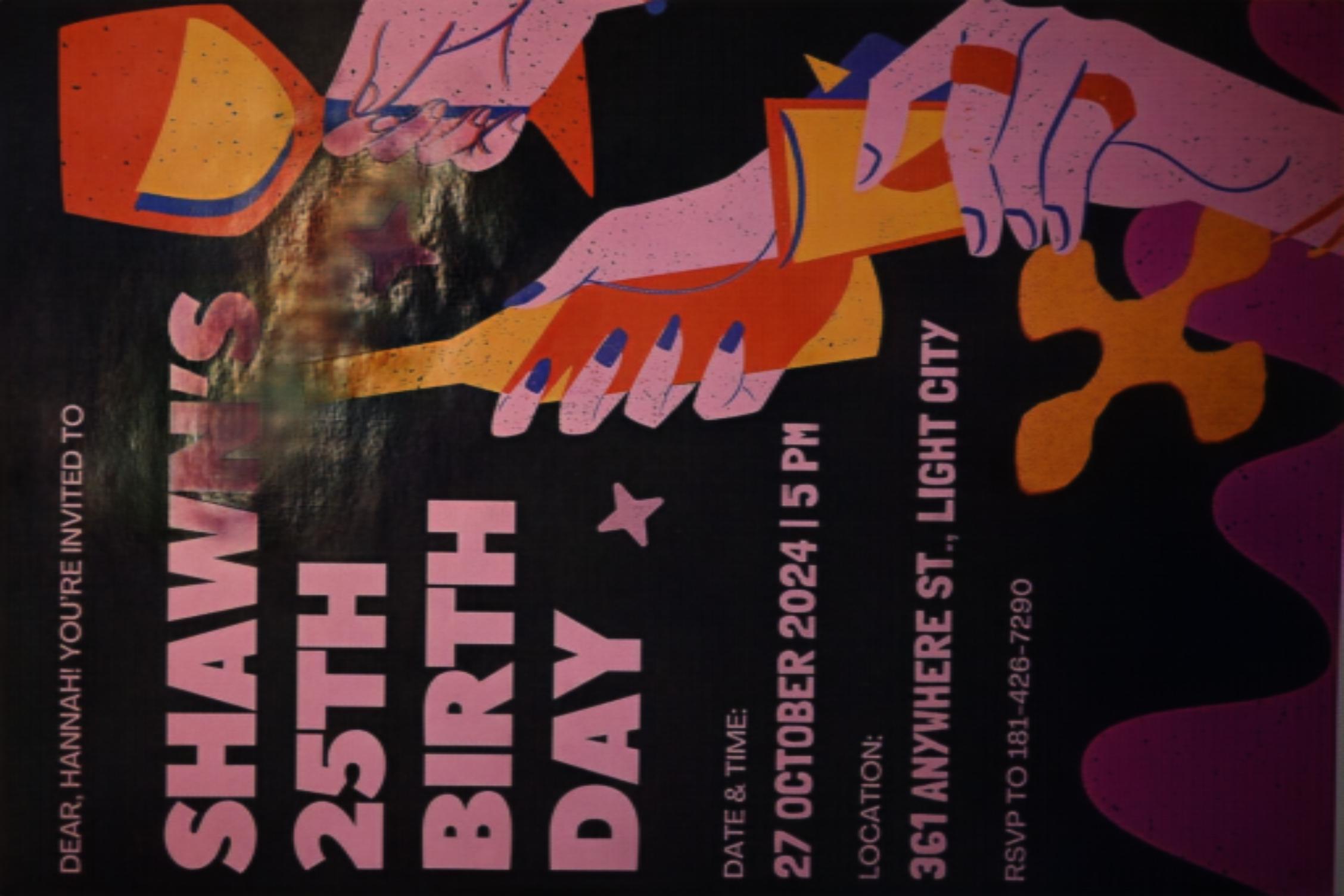}
        \caption{TSHRNet}
    \end{subfigure}
    \begin{subfigure}[b]{0.138\textwidth}
        \centering
        \includegraphics[angle=-90, width=\linewidth]{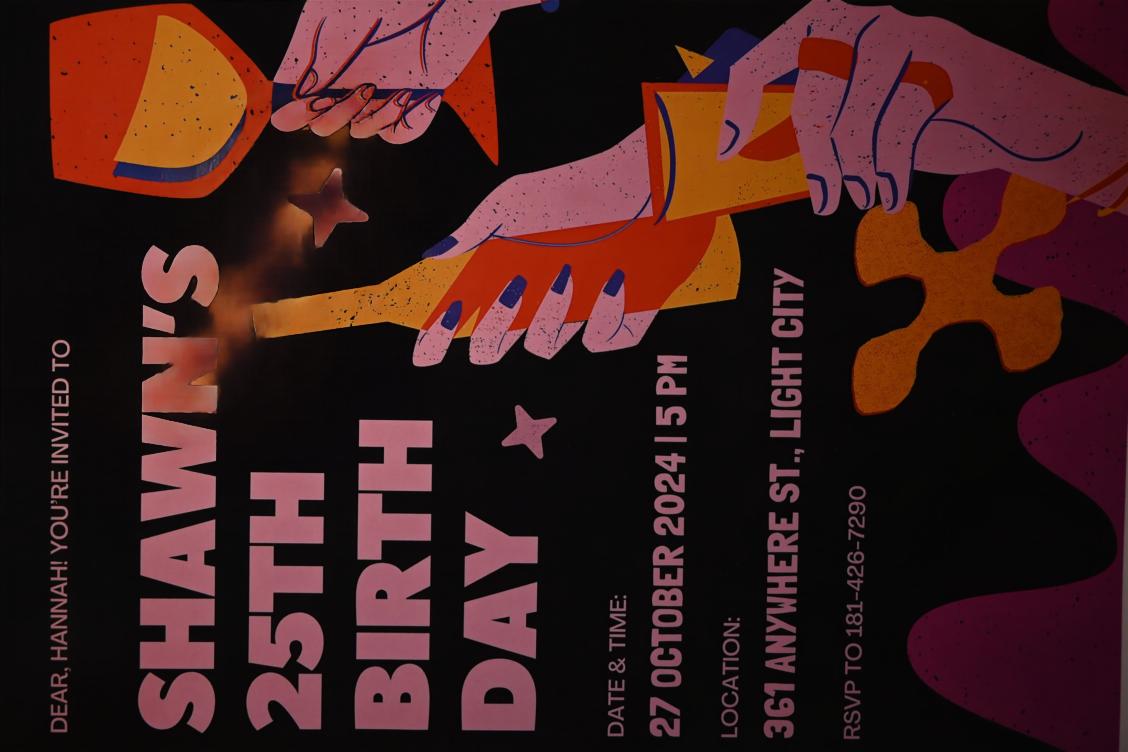}
        \caption{DHAN-SHR}
    \end{subfigure}
    \begin{subfigure}[b]{0.138\textwidth}
        \centering
        \includegraphics[angle=-90, width=\linewidth]{image/figure_9/DocShadowNet/013508.jpg}
         \caption{DocShadowNet}
    \end{subfigure}
    \begin{subfigure}[b]{0.138\textwidth}
        \centering
        \includegraphics[angle=-90, width=\linewidth]{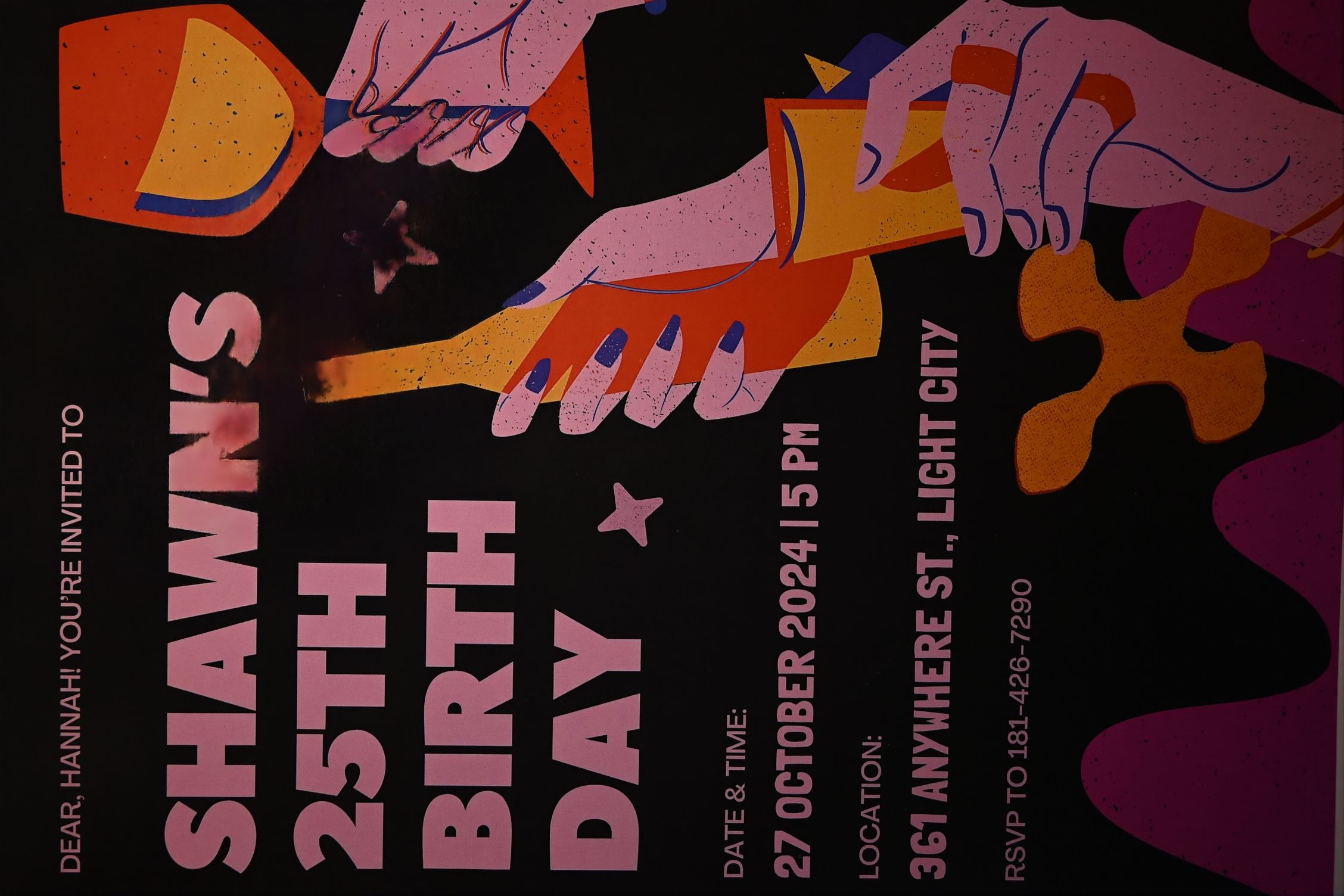}
        \caption*{Ours}
    \end{subfigure}
        \begin{subfigure}[b]{0.138\textwidth}
        \centering
        \includegraphics[angle=-90, width=\linewidth]{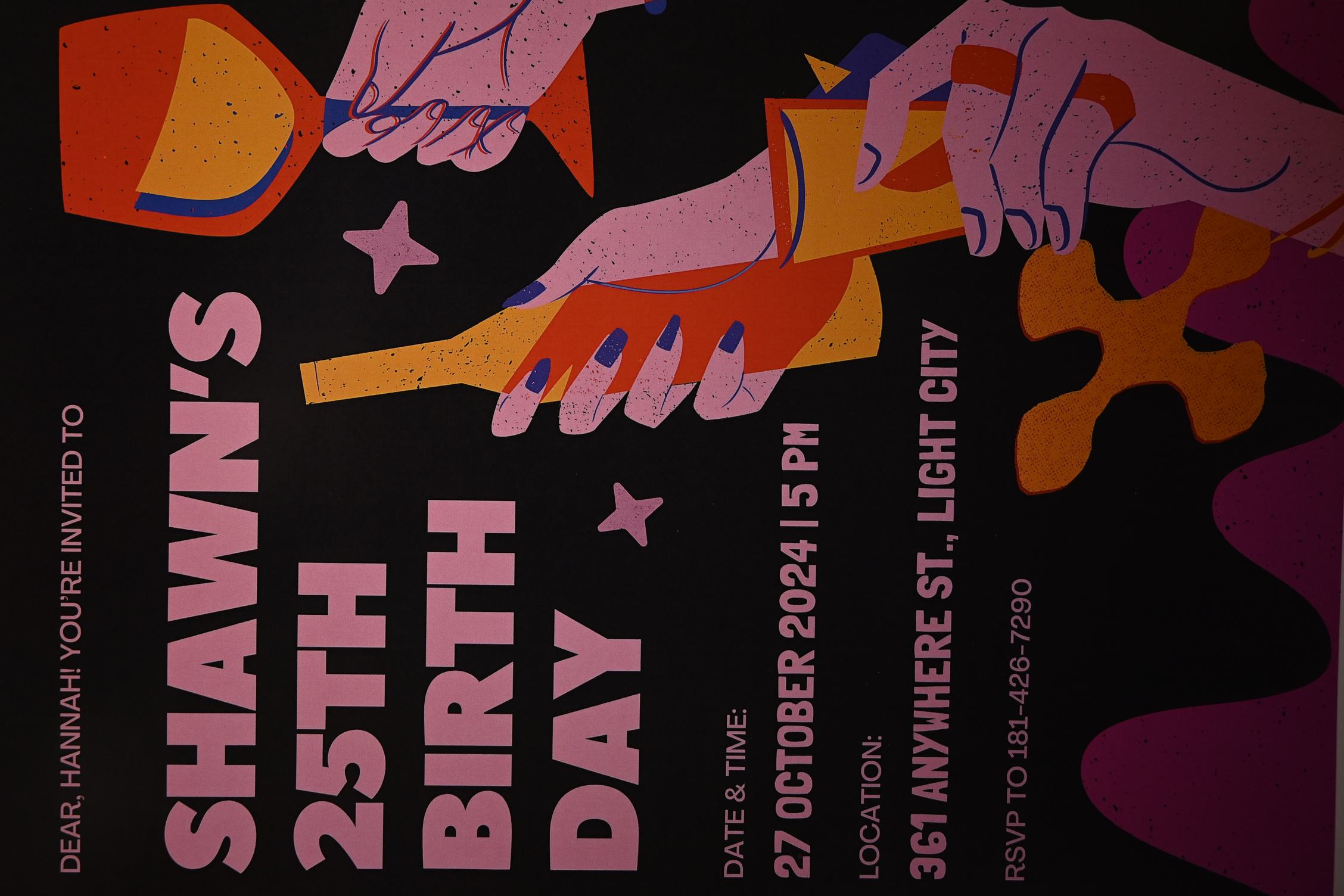}
        \caption*{GT}
    \end{subfigure} \\[-12pt]

    \caption{More qualitative comparisons of document highlight removal on our collected DocHR14K dataset. From left to right: the input highlight image, the estimated results of (a) JSHDR, (b) TSHRNet, (c) DHAN-SHR, (d) DocShadowNet, ours, and the ground truth image, respectively. \textbf{Zoom in} for best fit.}
    \label{fig:Qualitatively Comparison of DocHR14K}
\end{figure*}

\begin{figure*}[tp]  
    % the first row
    \begin{subfigure}[b]{0.138\textwidth}
        \centering
        \includegraphics[angle=0, width=\linewidth]{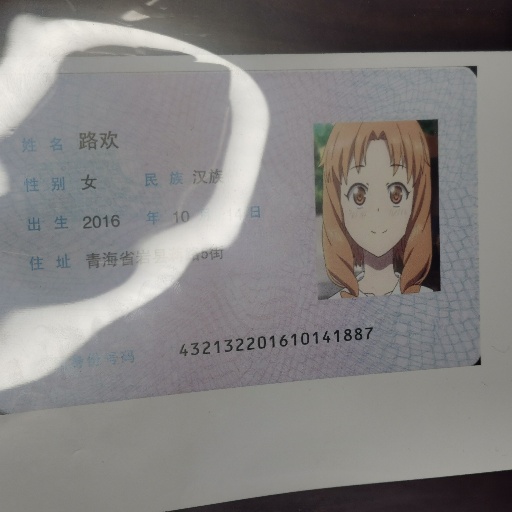} % 替换为实际文件路径
    \end{subfigure}
    \begin{subfigure}[b]{0.138\textwidth}
        \centering
        \includegraphics[angle=0, width=\linewidth]{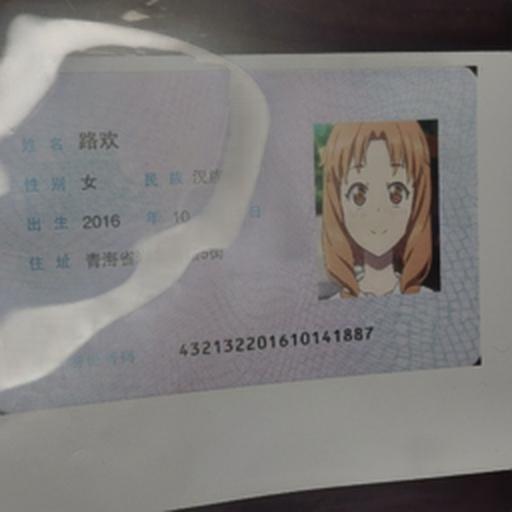} % 替换为实际文件路径
    \end{subfigure}
    \begin{subfigure}[b]{0.138\textwidth}
        \centering
        \includegraphics[angle=0, width=\linewidth]{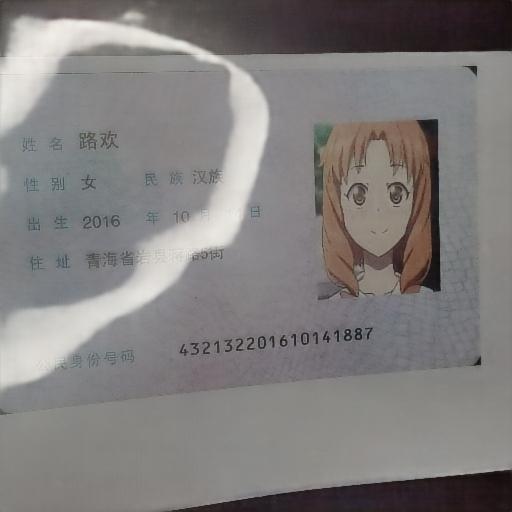} % 替换为实际文件路径
    \end{subfigure}
    \begin{subfigure}[b]{0.138\textwidth}
        \centering
        \includegraphics[angle=0, width=\linewidth]{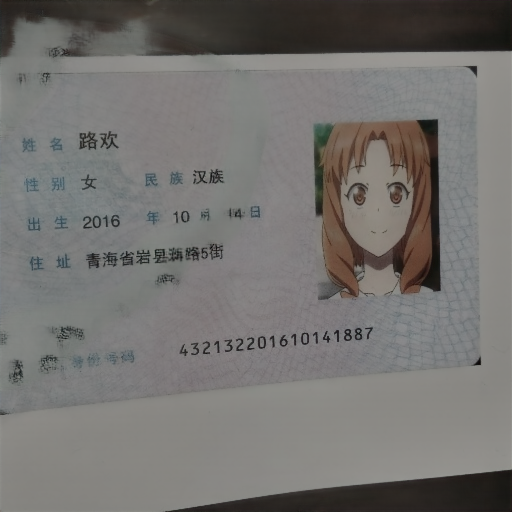} % 替换为实际文件路径
    \end{subfigure}
    \begin{subfigure}[b]{0.138\textwidth}
        \centering
        \includegraphics[angle=0, width=\linewidth]{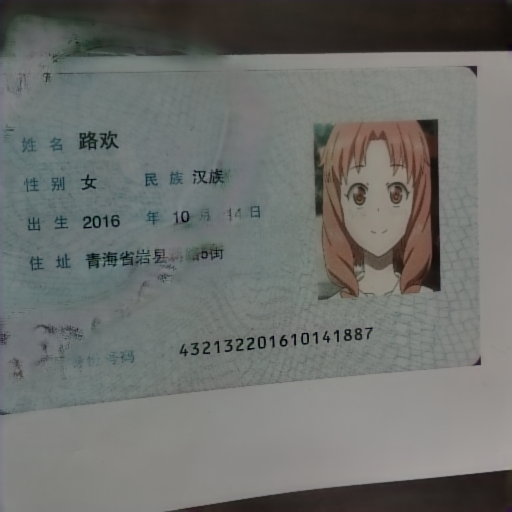} % 替换为实际文件路径
    \end{subfigure}
    \begin{subfigure}[b]{0.138\textwidth}
        \centering
        \includegraphics[angle=0, width=\linewidth]{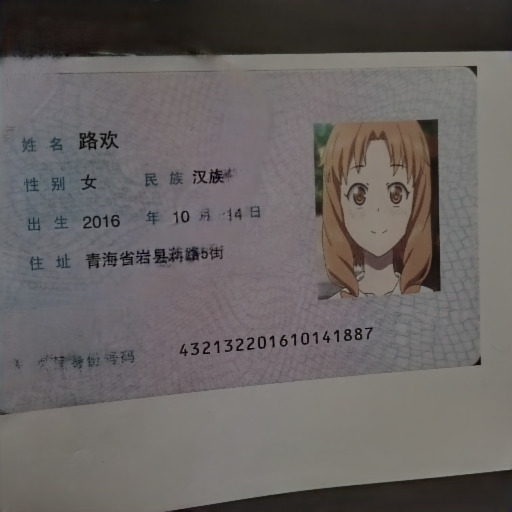} % 替换为实际文件路径
    \end{subfigure}
    \begin{subfigure}[b]{0.138\textwidth}
        \centering
        \includegraphics[angle=0, width=\linewidth]{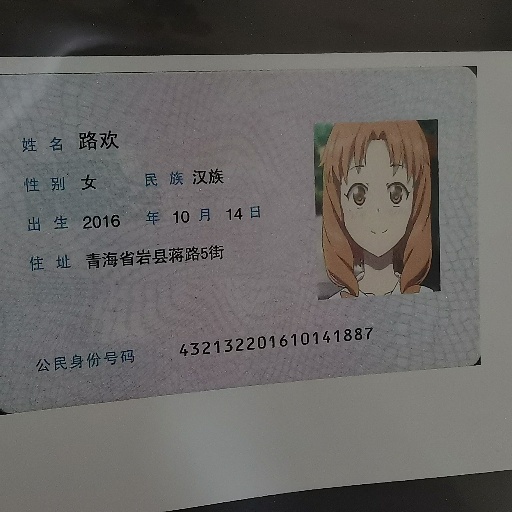} % 替换为实际文件路径
    \end{subfigure} \\
%-------------------------------------------------------------------------
    % the second row   
    \begin{subfigure}[b]{0.138\textwidth}
        \centering
        \includegraphics[angle=0, width=\linewidth]{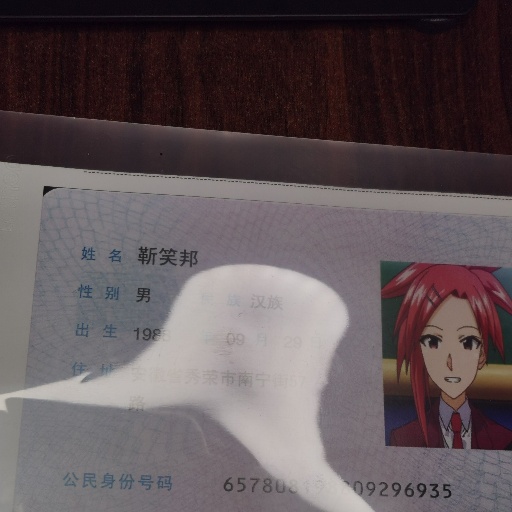}
        % \caption*{Input}
    \end{subfigure}
    \begin{subfigure}[b]{0.138\textwidth}
        \centering
        \includegraphics[angle=0, width=\linewidth]{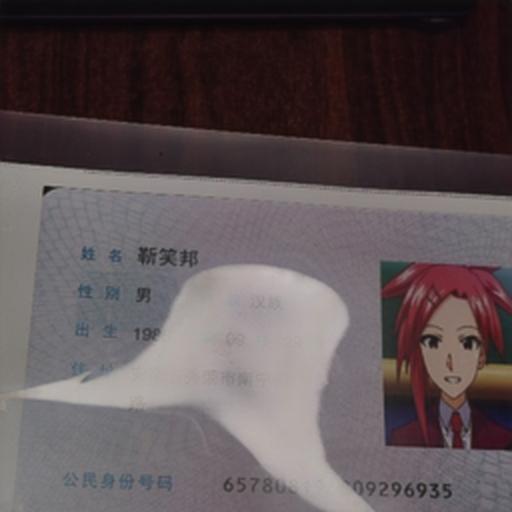}
        % \caption*{JSHDR}
    \end{subfigure}
    \begin{subfigure}[b]{0.138\textwidth}
        \centering
        \includegraphics[angle=0, width=\linewidth]{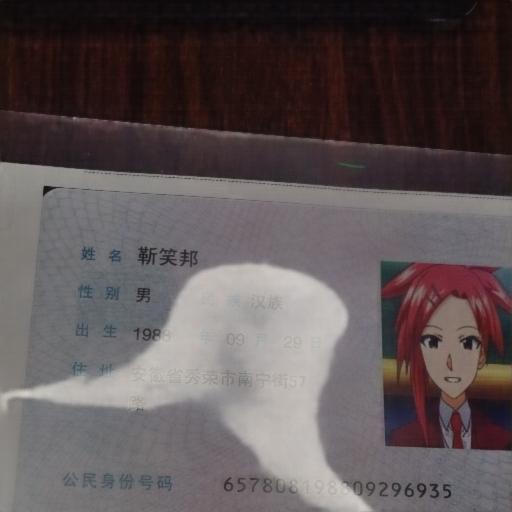}
        % \caption*{TSHRNet}
    \end{subfigure}
    \begin{subfigure}[b]{0.138\textwidth}
        \centering
        \includegraphics[angle=0, width=\linewidth]{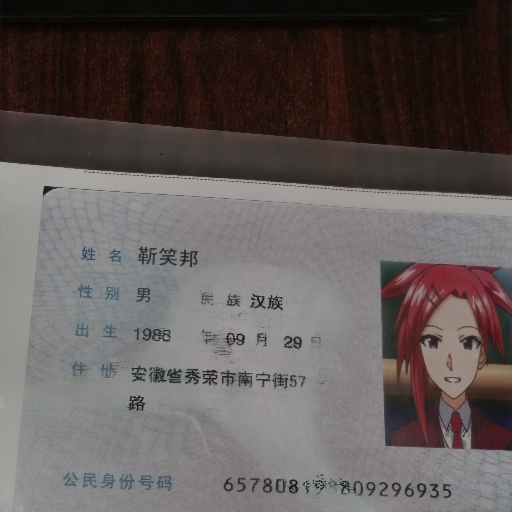}
        % \caption*{DHAN-SHR}
    \end{subfigure}
    \begin{subfigure}[b]{0.138\textwidth}
        \centering
        \includegraphics[angle=0, width=\linewidth]{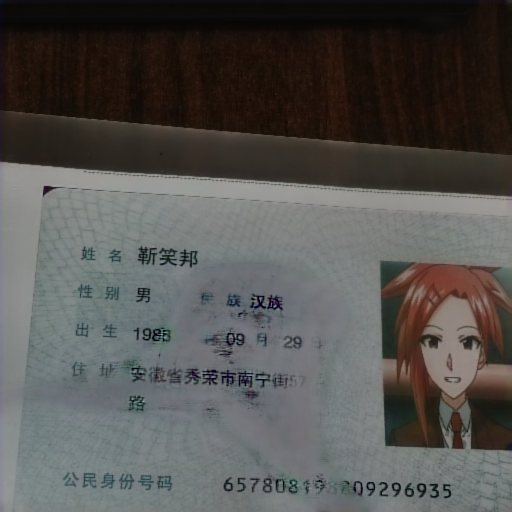}
        % \caption*{DocShadowNet}
    \end{subfigure}
    \begin{subfigure}[b]{0.138\textwidth}
        \centering
        \includegraphics[angle=0, width=\linewidth]{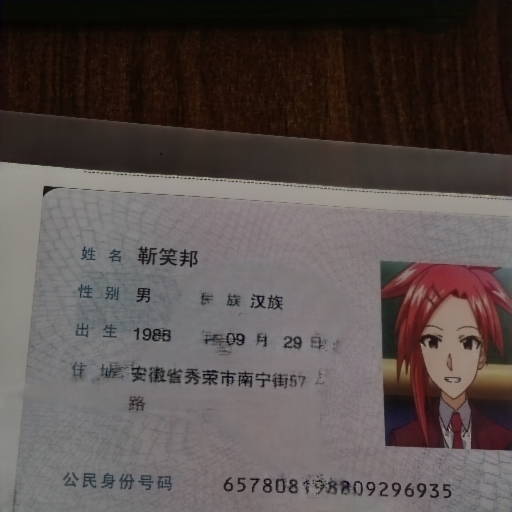}
        % \caption*{Ours}
    \end{subfigure}
        \begin{subfigure}[b]{0.138\textwidth}
        \centering
        \includegraphics[angle=0, width=\linewidth]{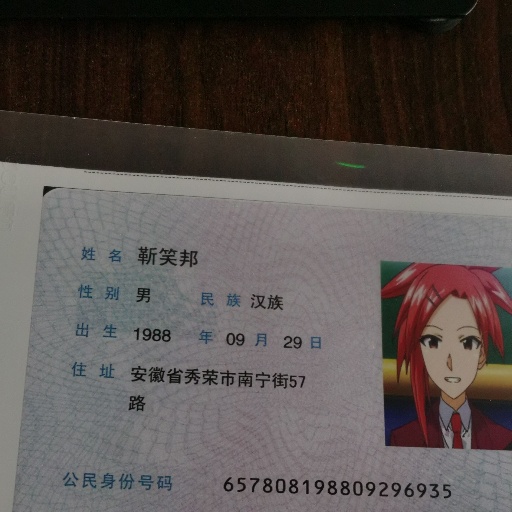}
        % \caption*{Reference}
    \end{subfigure} 
%-------------------------------------------------------------------------
%-------------------------------------------------------------------------
    % the second row   
    \begin{subfigure}[b]{0.138\textwidth}
        \centering
        \includegraphics[angle=0, width=\linewidth]{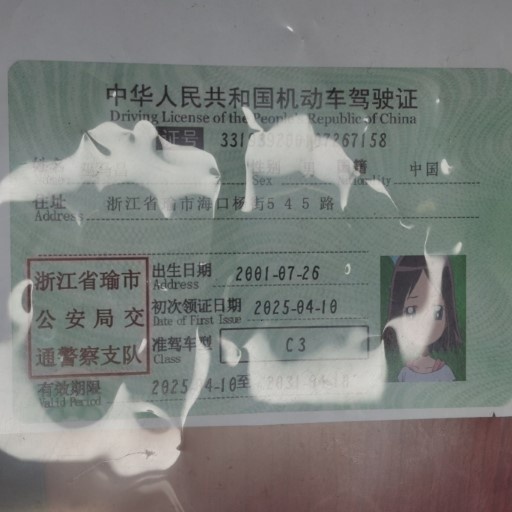}
        % \caption*{Input}
    \end{subfigure}
    \begin{subfigure}[b]{0.138\textwidth}
        \centering
        \includegraphics[angle=0, width=\linewidth]{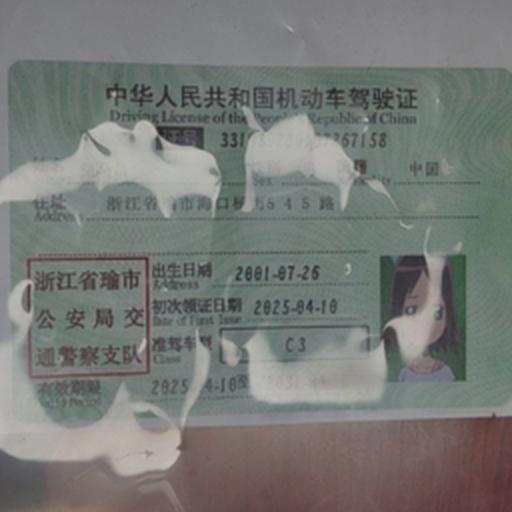}
        % \caption{JSHDR}
    \end{subfigure}
    \begin{subfigure}[b]{0.138\textwidth}
        \centering
        \includegraphics[angle=0, width=\linewidth]{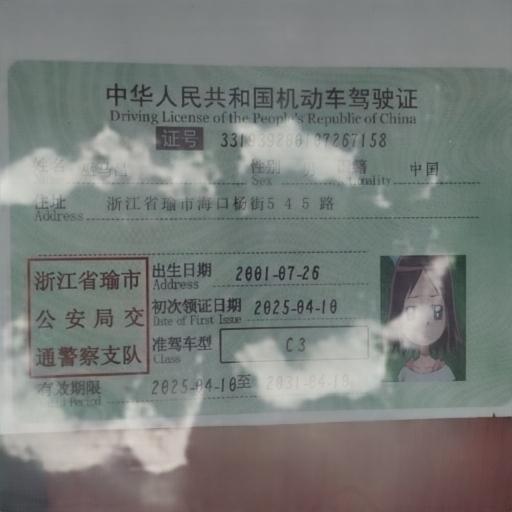}
        % \caption{TSHRNet}
    \end{subfigure}
    \begin{subfigure}[b]{0.138\textwidth}
        \centering
        \includegraphics[angle=0, width=\linewidth]{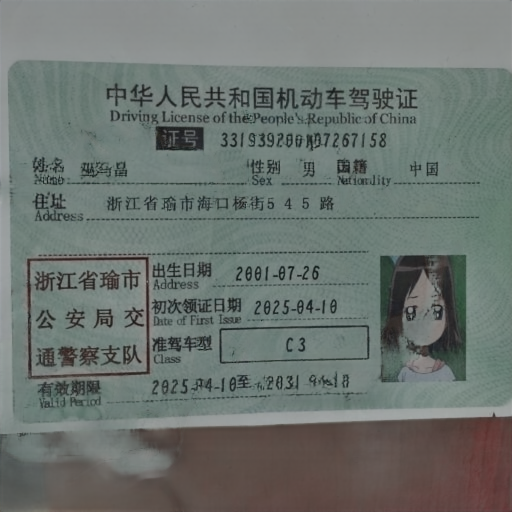}
        % \caption{DHAN-SHR}
    \end{subfigure}
    \begin{subfigure}[b]{0.138\textwidth}
        \centering
        \includegraphics[angle=0, width=\linewidth]{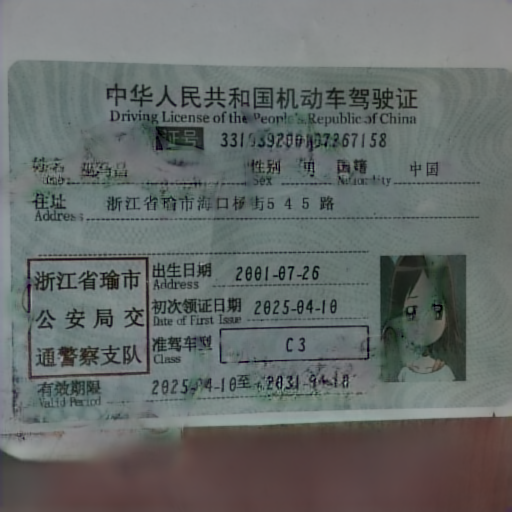}
        % \caption{DocShadowNet}
    \end{subfigure}
    \begin{subfigure}[b]{0.138\textwidth}
        \centering
        \includegraphics[angle=0, width=\linewidth]{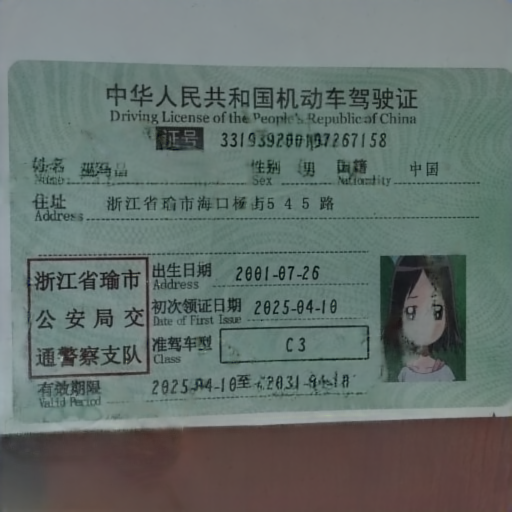}
        % \caption*{Ours}
    \end{subfigure}
        \begin{subfigure}[b]{0.138\textwidth}
        \centering
        \includegraphics[angle=0, width=\linewidth]{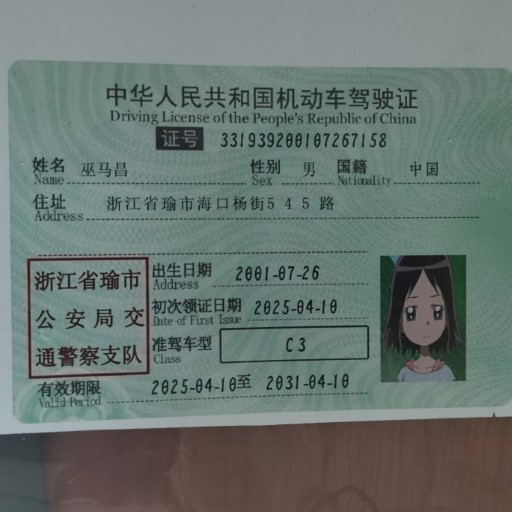}
        % \caption*{Reference}
    \end{subfigure} 
%-------------------------------------------------------------------------
%-------------------------------------------------------------------------
    % the second row   
    \begin{subfigure}[b]{0.138\textwidth}
        \centering
        \includegraphics[angle=0, width=\linewidth]{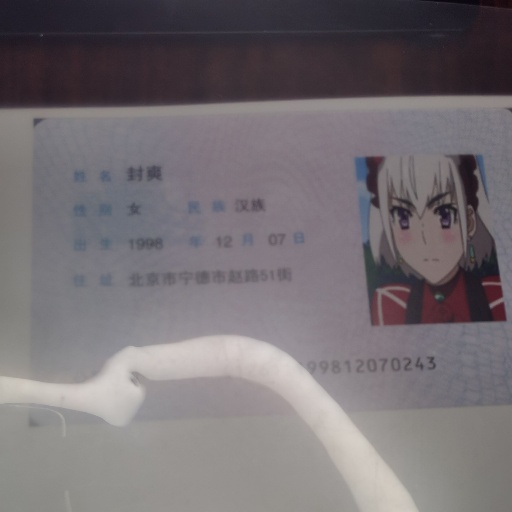}
        % \caption*{Input}
    \end{subfigure}
    \begin{subfigure}[b]{0.138\textwidth}
        \centering
        \includegraphics[angle=0, width=\linewidth]{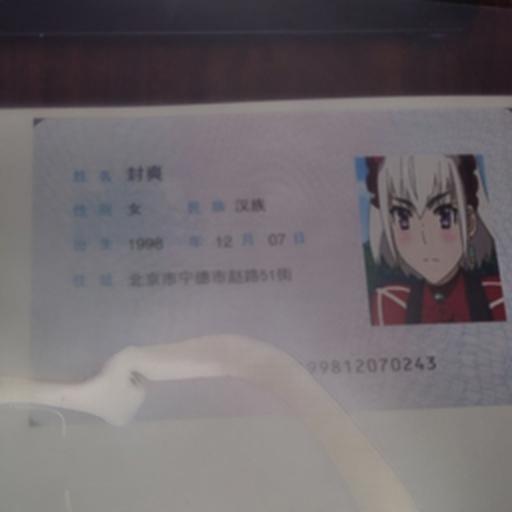}
        % \caption{JSHDR}
    \end{subfigure}
    \begin{subfigure}[b]{0.138\textwidth}
        \centering
        \includegraphics[angle=0, width=\linewidth]{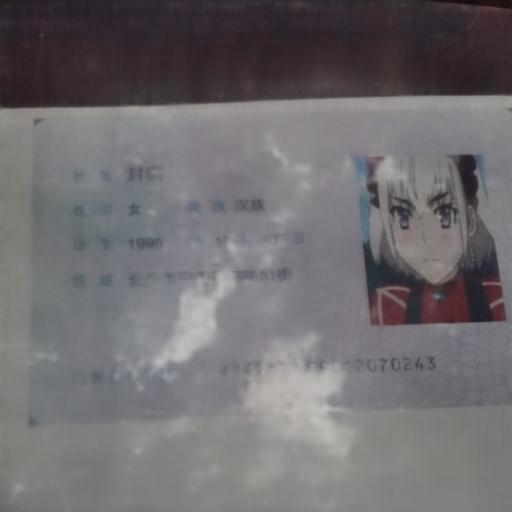}
        % \caption{TSHRNet}
    \end{subfigure}
    \begin{subfigure}[b]{0.138\textwidth}
        \centering
        \includegraphics[angle=0, width=\linewidth]{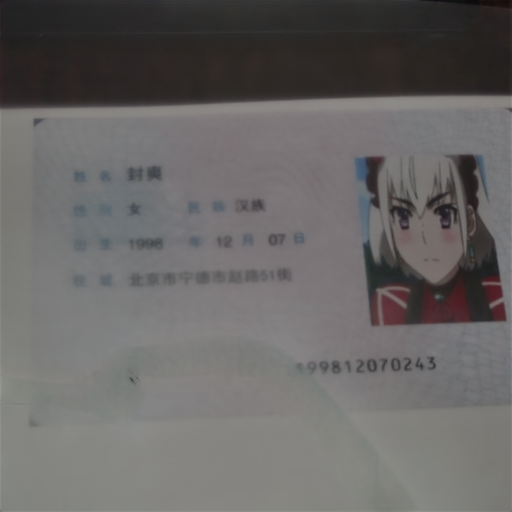}
        % \caption{DHAN-SHR}
    \end{subfigure}
    \begin{subfigure}[b]{0.138\textwidth}
        \centering
        \includegraphics[angle=0, width=\linewidth]{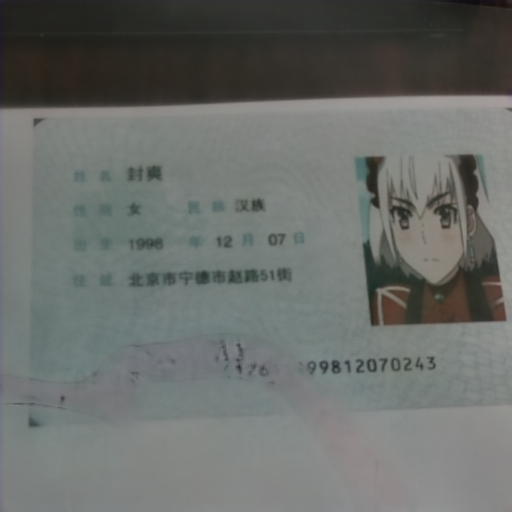}
        % \caption{DocShadowNet}
    \end{subfigure}
    \begin{subfigure}[b]{0.138\textwidth}
        \centering
        \includegraphics[angle=0, width=\linewidth]{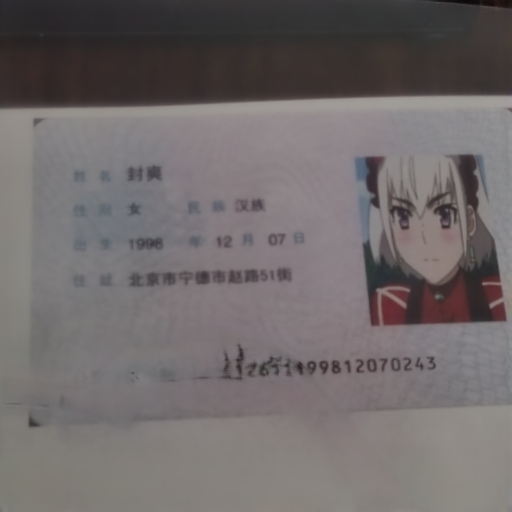}
        % \caption*{Ours}
    \end{subfigure}
        \begin{subfigure}[b]{0.138\textwidth}
        \centering
        \includegraphics[angle=0, width=\linewidth]{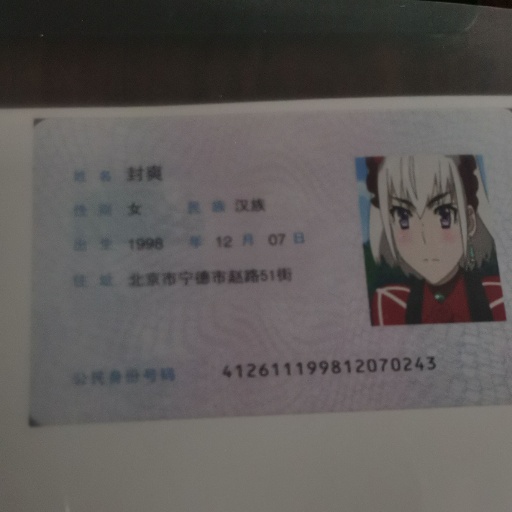}
        % \caption*{Reference}
    \end{subfigure} 
%-------------------------------------------------------------------------
    %the third row  
    \begin{subfigure}[b]{0.138\textwidth}
        \centering
        \includegraphics[angle=0, width=\linewidth]{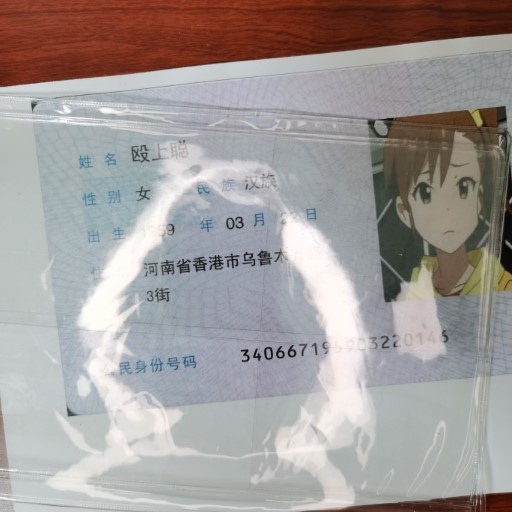}
        \caption*{Input}
    \end{subfigure}
    \begin{subfigure}[b]{0.138\textwidth}
        \centering
        \includegraphics[angle=0, width=\linewidth]{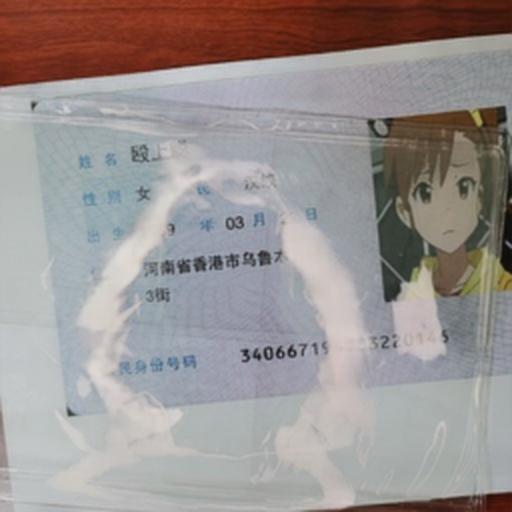}
        \caption{JSHDR}
    \end{subfigure}
    \begin{subfigure}[b]{0.138\textwidth}
        \centering
        \includegraphics[angle=0, width=\linewidth]{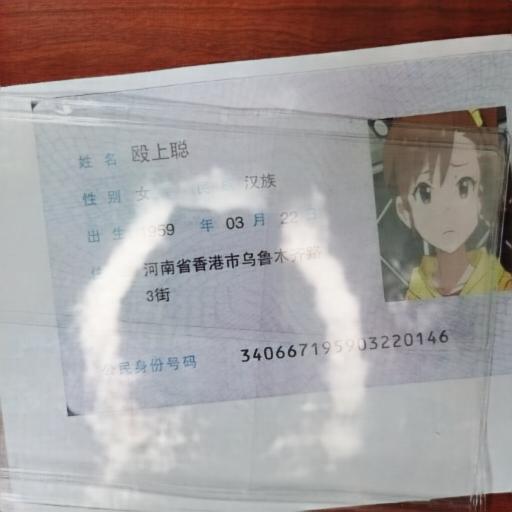}
        \caption{TSHRNet}
    \end{subfigure}
    \begin{subfigure}[b]{0.138\textwidth}
        \centering
        \includegraphics[angle=0, width=\linewidth]{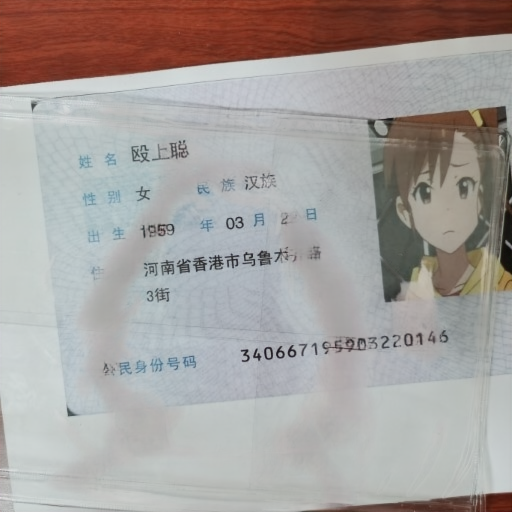}
        \caption{DHAN-SHR}
    \end{subfigure}
    \begin{subfigure}[b]{0.138\textwidth}
        \centering
        \includegraphics[angle=0, width=\linewidth]{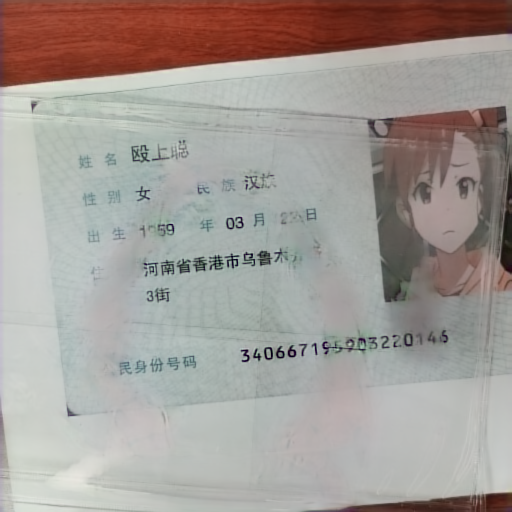}
         \caption{DocShadowNet}
    \end{subfigure}
    \begin{subfigure}[b]{0.138\textwidth}
        \centering
        \includegraphics[angle=0, width=\linewidth]{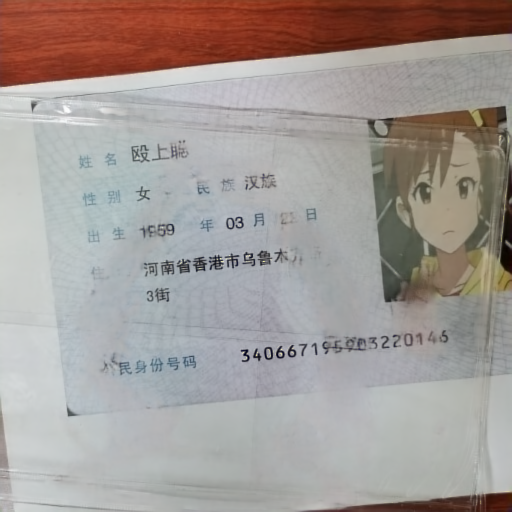}
        \caption*{Ours}
    \end{subfigure}
        \begin{subfigure}[b]{0.138\textwidth}
        \centering
        \includegraphics[angle=0, width=\linewidth]{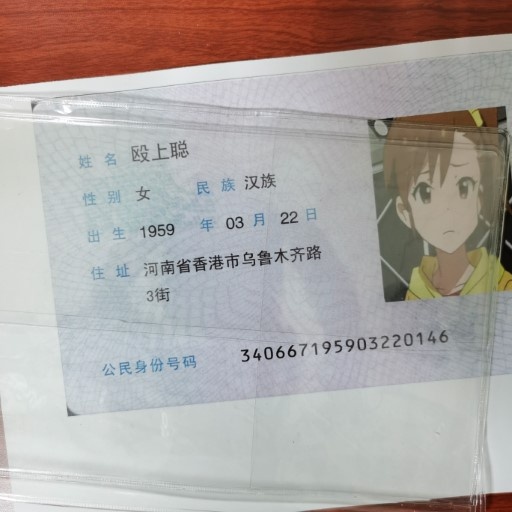}
        \caption*{GT}
    \end{subfigure}

    \caption{More qualitative comparisons of document highlight removal on public RD dataset. From left to right: the input highlight image, the estimated results of (a) JSHDR, (b) TSHRNet, (c) DHAN-SHR, (d) DocShadowNet, ours, and the ground truth image, respectively. \textbf{Zoom in} for best fit.}
    \label{fig:Qualitatively Comparison of RD}
\end{figure*}

\begin{figure*}[tp]
    \centering
    % 每列放一个方法，共7列，每列纵向放5张图像
    \begin{subfigure}[b]{0.12\linewidth}
        \centering
        \includegraphics[width=\linewidth]{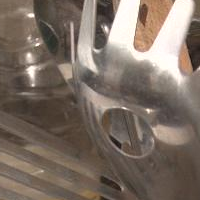}
        \includegraphics[width=\linewidth]{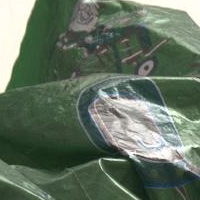}
        \includegraphics[width=\linewidth]{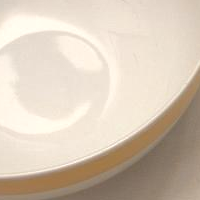}
        \includegraphics[width=\linewidth]{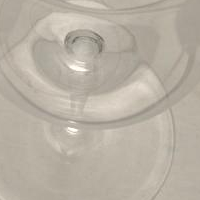}
        \includegraphics[width=\linewidth]{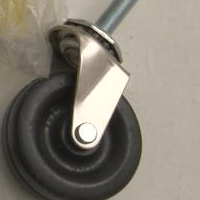}
        \caption*{Input}
    \end{subfigure}
    % \begin{subfigure}[b]{0.12\linewidth}
    %     \centering
    %     \includegraphics[width=\linewidth]{appendix/SHIQ_dataset/Yang/000390.jpg}\
    %     \includegraphics[width=\linewidth]{appendix/SHIQ_dataset/Yang/000142.jpg}\
    %     \includegraphics[width=\linewidth]{appendix/SHIQ_dataset/Yang/000811.jpg}\
    %     \includegraphics[width=\linewidth]{appendix/SHIQ_dataset/Yang/000506.jpg}
    %     \includegraphics[width=\linewidth]{appendix/SHIQ_dataset/Yang/000572.jpg}
    %     \caption*{Yang}
    % \end{subfigure}
    \begin{subfigure}[b]{0.12\linewidth}
        \centering
        \includegraphics[width=\linewidth]{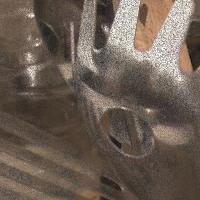}       
        \includegraphics[width=\linewidth]{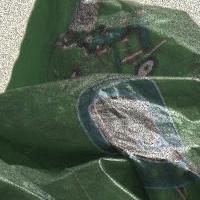}
        \includegraphics[width=\linewidth]{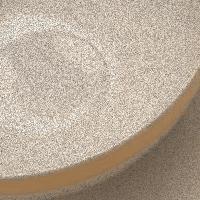} 
        \includegraphics[width=\linewidth]{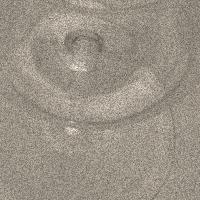}
        \includegraphics[width=\linewidth]{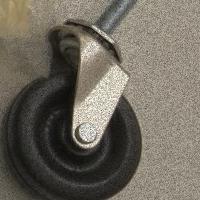}
        \caption*{Akashi}
    \end{subfigure}
    \begin{subfigure}[b]{0.12\linewidth}
        \centering
        \includegraphics[width=\linewidth]{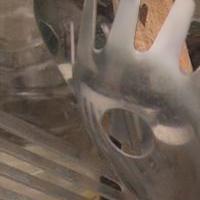}       
        \includegraphics[width=\linewidth]{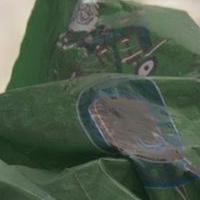}
        \includegraphics[width=\linewidth]{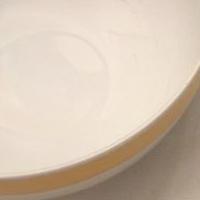} 
        \includegraphics[width=\linewidth]{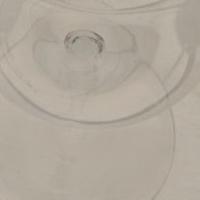}
        \includegraphics[width=\linewidth]{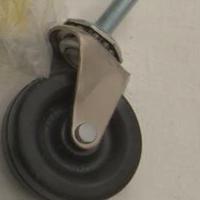}
        \caption*{JSHDR}
    \end{subfigure}
    \begin{subfigure}[b]{0.12\linewidth}
        \centering
        \includegraphics[width=\linewidth]{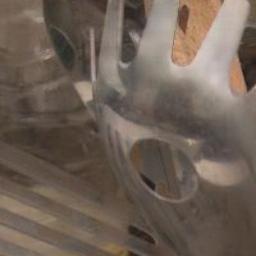}
        \includegraphics[width=\linewidth]{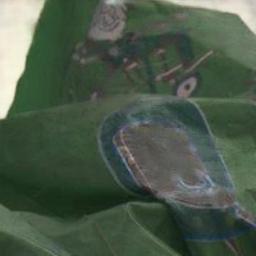}
        \includegraphics[width=\linewidth]{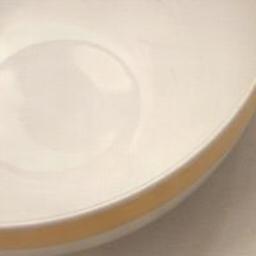}
        \includegraphics[width=\linewidth]{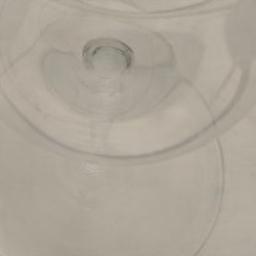}
        \includegraphics[width=\linewidth]{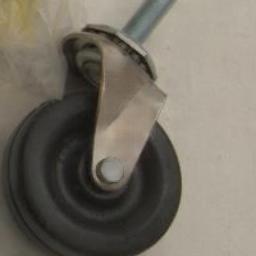}
        \caption*{TSHRNet}
    \end{subfigure}
    \begin{subfigure}[b]{0.12\linewidth}
        \centering
        \includegraphics[width=\linewidth]{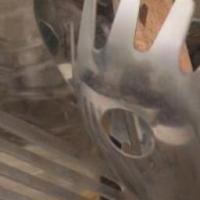}
        \includegraphics[width=\linewidth]{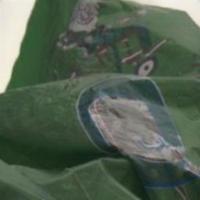}
        \includegraphics[width=\linewidth]{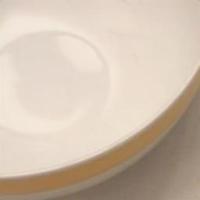}
        \includegraphics[width=\linewidth]{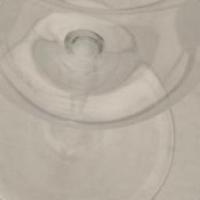}
        \includegraphics[width=\linewidth]{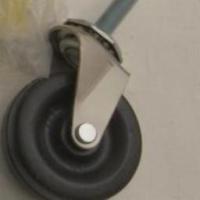}
        \caption*{DHAN-SHR}
    \end{subfigure}
    \begin{subfigure}[b]{0.12\linewidth}
        \centering
        \includegraphics[width=\linewidth]{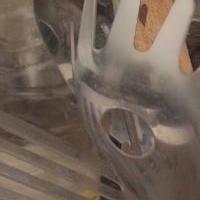}
        \includegraphics[width=\linewidth]{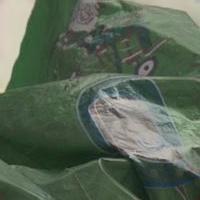}
        \includegraphics[width=\linewidth]{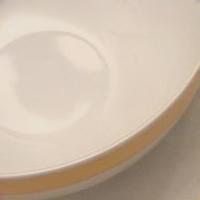}
        \includegraphics[width=\linewidth]{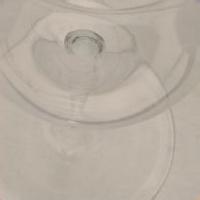}
        \includegraphics[width=\linewidth]{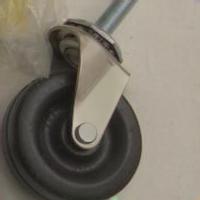}
        \caption*{DocShadowNet}
    \end{subfigure}
    \begin{subfigure}[b]{0.12\linewidth}
        \centering
        \includegraphics[width=\linewidth]{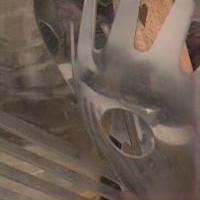}
        \includegraphics[width=\linewidth]{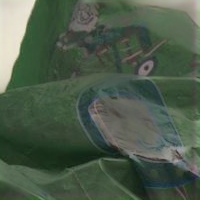}
        \includegraphics[width=\linewidth]{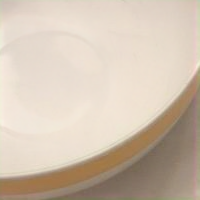}
        \includegraphics[width=\linewidth]{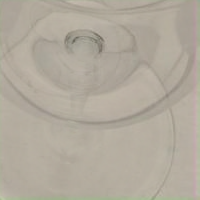}
        \includegraphics[width=\linewidth]{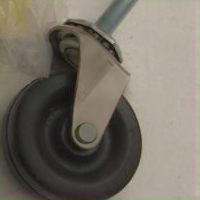}
        \caption*{Ours}
    \end{subfigure}
    \begin{subfigure}[b]{0.12\linewidth}
        \centering
        \includegraphics[width=\linewidth]{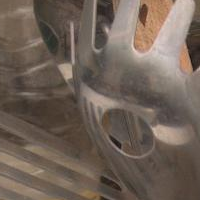}
        \includegraphics[width=\linewidth]{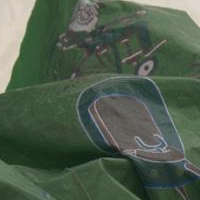}
        \includegraphics[width=\linewidth]{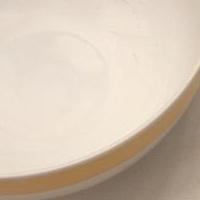}
        \includegraphics[width=\linewidth]{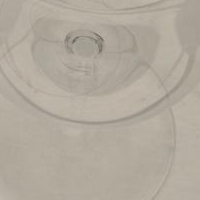}
        \includegraphics[width=\linewidth]{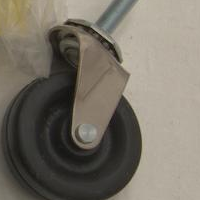}
        \caption*{GT}
    \end{subfigure}

    \caption{Qualitative comparisons of document highlight removal on public SHIQ dataset. From left to right: the input highlight image, the estimated results of Akashi, JSHDR,  TSHRNet, DHAN-SHR, DocShadowNet, ours, and the ground truth image, respectively.}
    \label{fig:Qualitatively Comparison of SHIQ}
\end{figure*}

\Cref{fig:Qualitatively Comparison of DocHR14K}, \Cref{fig:Qualitatively Comparison of RD} and \Cref{fig:Qualitatively Comparison of SHIQ} provide more visual comparisons among SOTA methods and our L\textsuperscript{2}HRNet on DocHR14K, RD, and SHIQ dataset, respectively.  As illustrated, our method achieves more effective global highlight removal with fewer highlight residuals and sharper text details compared to existing SOTA approaches. These results demonstrate the effectiveness of our proposed highlight location prior and the diffusion-based enhancement model, which operates on the low- and high-frequency bands decomposed by the Laplacian Pyramid.

%-------------------------------------------------------------------------

\section{Limitations and Future Work}
\label{sec: Limitations and Future Work}
While our method excels at removing highlights in document images, it has two main limitations: (1) under extreme overexposure, the original content can be lost forever and cannot be recovered; and (2) our diffusion‐based enhancement, even when paired with a DPM‐Solver, still incurs relatively slow inference. In future work, we plan to (i) integrate image inpainting to fill truly missing regions using semantic cues, and (ii) switch to faster latent‐space approaches—such as flow‐matching methods or latent diffusion models—to dramatically reduce runtime.
% %-------------------------------------------------------------------------

%-------------------------------------------------------------------------

%-------------------------------------------------------------------------

\clearpage
\end{document}